\documentclass[10pt,twocolumn,letterpaper]{article}
\usepackage{cvpr}              % To produce the CAMERA-READY version

\usepackage[dvipsnames]{xcolor}
\usepackage{multirow}
\usepackage{array}
\usepackage{vcell}
\usepackage{makecell}
\usepackage{afterpage}
\usepackage{tikz}
\usetikzlibrary{calc}

\definecolor{ms_note}{RGB}{0, 181, 190}
\definecolor{burgundy}{RGB}{150, 0, 0}
\definecolor{pink}{rgb}{1,0.0,1.0}

\definecolor{rgbd_model}{RGB}{180, 96, 6}
\definecolor{warp_controlnet_model}{RGB}{66, 133, 244}
\definecolor{feature_sharing}{RGB}{56, 118, 29}

\definecolor{cvprblue}{rgb}{0.21,0.49,0.74}
\usepackage[pagebackref,breaklinks,colorlinks,citecolor=cvprblue]{hyperref}

\def\paperID{3769} % *** Enter the Paper ID here
\def\confName{CVPR}
\def\confYear{2025}

\title{Morpheus: Text-Driven 3D Gaussian Splat Shape and Color Stylization}
\author{\\
Niantic Labs\\
}
\author{Zawar Qureshi*\\
Niantic Labs\\
}

\newcommand{\gap}{\hspace{10pt}}
\author{Jamie Wynn*$^{1}$ \gap Zawar Qureshi*$^{1}$ \gap Jakub Powierza$^{1}$ \gap Jamie Watson$^{1,2}$ \gap Mohamed Sayed$^{1}$  \\ 
 $^{1}$Niantic \hspace{30pt} \hspace{30pt}$^{2}$UCL\\
 \url{https://nianticlabs.github.io/morpheus/}}
 
\begin{document}

\twocolumn[{%
\renewcommand\twocolumn[1][]{#1}%
\maketitle
\begin{center}
    \centering
    \captionsetup{type=figure}
    \includegraphics[width=\textwidth]{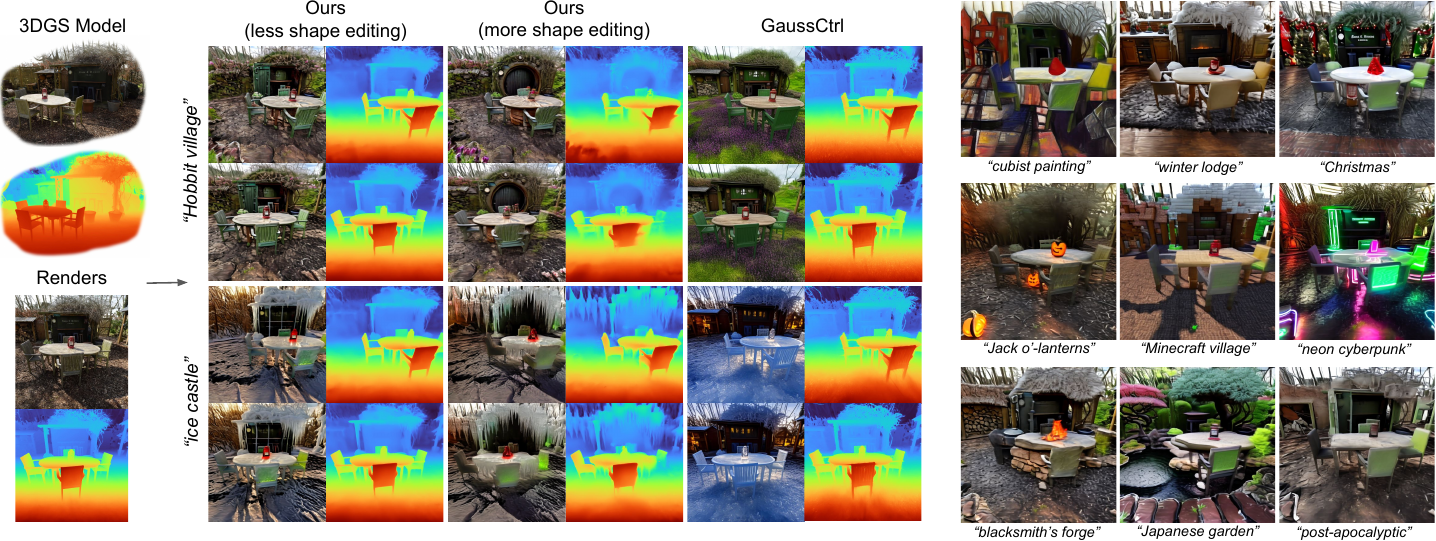}
    \captionof{figure}{We introduce a new method for novel-view stylization using text prompts. The output of our method is a stylized 3D Gaussian Splatting model, from which we show renders here. Our method allows stylization control of both appearance and shape. Using the same prompt, our method can produce different stylizations with the same overall texture, but variable shape alteration allowing for more striking shape and color stylization compared to GaussCtrl~\cite{wu2024gaussctrl}. We show multiple stylizations of the same scene. \href{https://nianticlabs.github.io/morpheus/}{Code online}.}
\end{center}%
}]

\begin{abstract}

Exploring real-world spaces using novel-view synthesis is fun, and reimagining those worlds in a different style adds another layer of excitement. Stylized worlds can also be used for downstream tasks where there is limited training data and a need to expand a model's training distribution. Most current novel-view synthesis stylization techniques lack the ability to convincingly change geometry. This is because any geometry change requires increased style strength which is often capped for stylization stability and consistency. In this work, we propose a new autoregressive 3D Gaussian Splatting stylization method. As part of this method, we contribute a new RGBD diffusion model that allows for strength control over appearance and shape stylization. To ensure consistency across stylized frames, we use a combination of novel depth-guided cross attention, feature injection, and a Warp ControlNet conditioned on composite frames for guiding the stylization of new frames. We validate our method via extensive qualitative results, quantitative experiments, and a user study. 

\end{abstract}
\def\thefootnote{*}\footnotetext{denotes equal contribution.}\def\thefootnote{\arabic{footnote}}
\section{Introduction}
As humans, we want to explore worlds and stories beyond what we experience in our daily lives for entertainment or education. Our history of art reflects this; we started with paintings on stone walls, moved on to pigment on canvas, mastered the art and technicalities of photography, and then invented moving pictures. When this was not enough, we pushed for better ways of experiencing these worlds by either adding a sense of depth as with stereoscopes or Sensorama~\cite{wade2002charles, gutierrez2023ballad} or by allowing freedom and control over viewpoint~\cite{carter2024picturing}. More recently, these experiences have been realized by simulating a world using computer generated graphics. Realism is achieved with painstaking work by artists to make textures and models, and then work by graphics researchers to allow real-time renders of these complex and detailed worlds for them to come alive. This also enables the creation of new kinds of worlds that are no longer necessarily `real' but still detailed and immersive enough to convince viewers and players that they are real.

However, this all comes at a cost. While building 3D representations of structures and objects by hand -- often in the form of meshes -- allows for full artistic control, it requires expensive artistic skill and time. Recent work aims to cut down the effort needed to bring real objects into the virtual world. This could involve reconstructing a 3D model of an object from one~\cite{qian2023magic123} or a collection of images~\cite{schoenberger2016mvs, yao2018mvsnet}. A more convenient approach is to render novel views of a scene using image-based rendering~\cite{hedman2016InsideOut, mildenhall2021nerf, kerbl20233d}. The only requirement for these methods is a dense collection of images with camera poses obtained via either SLAM~\cite{zhang2023goslam, zhu2022niceslam} or SfM~\cite{schoenberger2016sfm}. The most recent of these methods, 3D Gaussian Splatting~\cite{kerbl20233d} (3DGS), uses primitives that can be rendered in real-time using conventional rasterization techniques, allowing almost-seamless integration into existing rendering pipelines. While this gives us realism with little effort, we are limited by what we can capture in the real world. The next challenge is to change these captures to allow for the exploration of and immersion in stylized versions of those worlds.

There are many works on altering captures of the real world. The simplest problem is editing or stylizing 2D images, where many recent methods excel~\cite{isola2017image, li2016precomputed, rombach2022high}; there, stylization is often controlled using language prompts and example images. A more challenging setting is stylizing 3D representations. Since existing 2D generative and stylization models are powerful and mature, they are often used as a building block for 3D editing. There, stylization has to be consistent from one view to another, or viewer immersion is broken. These 2D models often lack explicit understanding of geometry, the ability to output a representation of modified geometry, and strength control over appearance and shape stylization. Because of this, modification is often superficial and limited to texture changes, especially since multi-view consistency is required and is easily broken with increased stylization strength -- especially if geometry is edited.

To tackle these challenges, we introduce a new method for stylizing 3D Gaussian Splats. Our method is informed by the geometry present in the input 3DGS model and allows strength control over appearance and shape stylization. Our method operates on renders of a 3DGS, producing frame-by-frame stylization for arbitrary camera trajectories. We show that stylized 3DGS models made from those stylized frames are qualitatively and quantitatively superior to existing methods. We highlight our contributions as:
\begin{enumerate}
    \item an autoregressive pipeline for stylizing 3D Gaussian Splatting models of scenes given text prompts,
    \item an RGBD stylization diffusion model conditioned on a text prompt and an RGBD image with separate controls over geometry and appearance,
    \item a ControlNet conditioned on warped frame composites for propagating previous frame stylization,
    \item and depth-informed feature sharing for consistent frame-to-frame stylization.
\end{enumerate}

\begin{figure*}[h!]
    \centering
    \vspace{-5pt}
    \newcommand{\qualimwidth}{0.23\textwidth}
    \renewcommand{\tabcolsep}{2pt}
    \small
    \includegraphics[width=.9\textwidth]{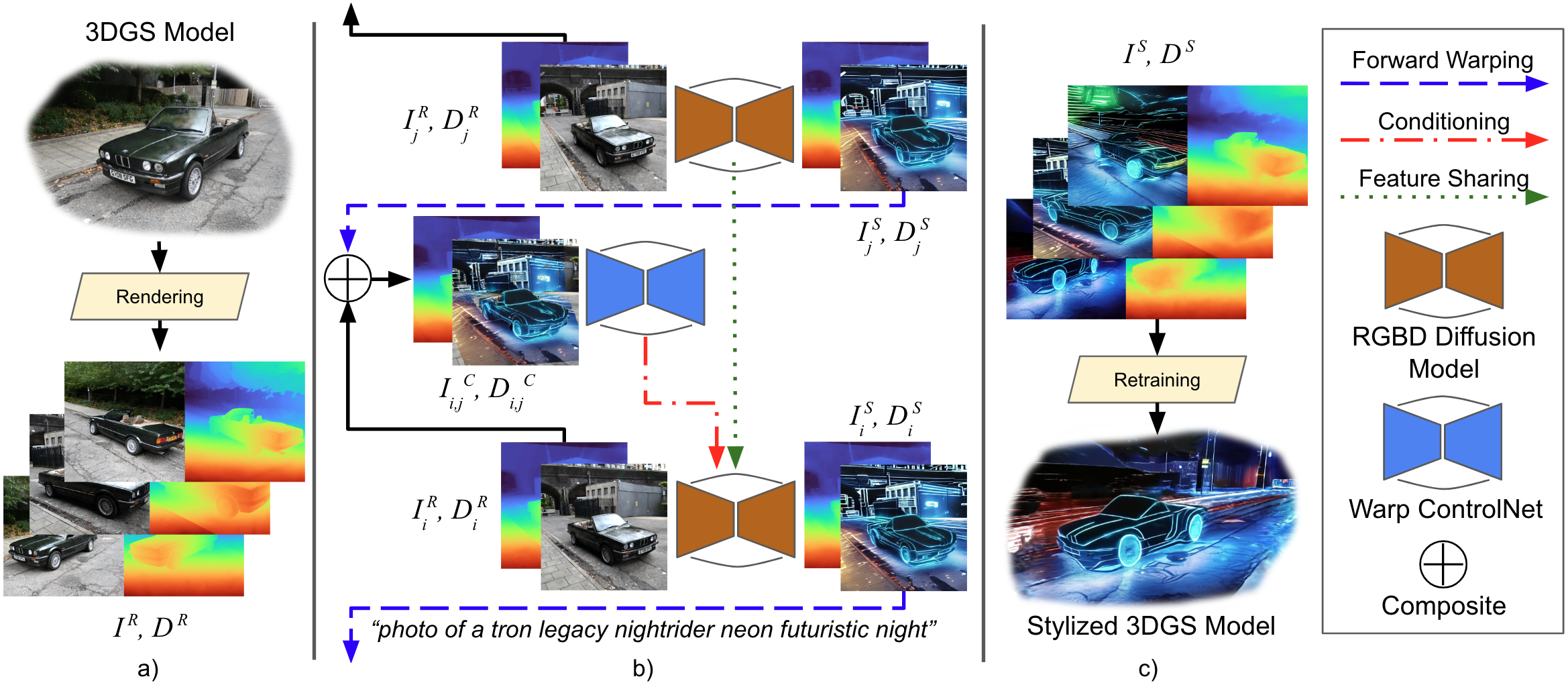}
    
    \vspace{-5pt}
    \caption{\textbf{Method Overview} a) Our pipeline takes as input a novel view synthesis model, in this case a 3D Gaussian Splatting (3DGS) model, and first renders a set of representative images and their depth maps ${\{I^R, D^R\}}$. b) Our pipeline stylizes rendered images autoregressively. We use a novel \textcolor{rgbd_model}{RGBD diffusion model} (Section~\ref{sec:rgbd_model}) conditioned on the input RGBD render ${\{I_i^S, D_i^S\}}$, a stylization prompt, and stylization noise parameters that modulate the strength of appearance and shape stylization. For every subsequent frame, we warp previously stylized frames ${\{I_j^S, D_j^S\}}$ to the current frame and form a composite ${\{I_{ij}^C, D_{ij}^C\}}$. We use a \textcolor{warp_controlnet_model}{Warp ControlNet} (Section~\ref{sec:warpingcontrolnet}) conditioned on the warped composite and a validity mask to guide the RGBD stylization of the current frame ${\{I_i^R, D_i^R\}}$ to produce ${\{I_i^S, D_i^S\}}$. During diffusion we use \textcolor{feature_sharing}{depth-informed feature sharing} (Section~\ref{sec:depth_informed_sharing}) to propagate deep stylization features. c) We then retrain a 3DGS model using newly stylized frames ${\{I^S, D^S\}}$.}
    \vspace{-5pt}
    \label{fig:overview_method}    
\end{figure*}

\section{Related Work}

\noindent\textbf{Novel-View Synthesis (NVS)} is a popular task in Computer Graphics where, given an image or a collection of posed images of a scene, an image is output from an arbitrary view. Early approaches construct lumigraphs~\cite{gortler2023lumigraph} -- volumes that capture the behavior of light as it passes through a scene -- that can be used to render novel views. Subsequent approaches aim to reconstruct textured geometry~\cite{chaurasia2013depth, penner2017soft, hedman2016InsideOut} to utilize traditional rendering pipelines, multiplane images for layered rendering~\cite{szeliski1999stereo, tucker2020single}, and learned networks for combining multiple image fragments~\cite{hedman2018DeepBlending, riegler2020free}. More recent methods render novel views volumetrically using implicit functions that learn a radiance field of a scene via gradient descent~\cite{mildenhall2021nerf}. An alternate approach~\cite{kerbl20233d} is to optimize a set of 3D Gaussian primitives that can be rendered down to images via splatting in real-time. Subsequent work uses regularization during optimization~\cite{deng2022depth, pumarola2021d, chen2021mvsnerf, chen2024mvsplat}, models raw capture~\cite{mildenhall2022nerf}, improves rendering time~\cite{chen2023mobilenerf, hedman2021baking, garbin2021fastnerf, muller2022instant, liu2024fast}, improves training speed and/or quality~\cite{reiser2021kilonerf, muller2022instant, barron2021mip, zhang2020nerf++}, estimates camera parameters~\cite{wang2021nerf, bian2023nope}, or incorporates semantic understanding~\cite{kerr2023lerf, qin2024langsplat}.

\noindent\textbf{2D Image Stylization} Early image stylization approaches first extract low level image features such as gradients, edges, and local segments~\cite{kyprianidis2012state} and then place brush strokes~\cite{haeberli1990paint, hertzmann1998painterly, lu2010interactive}, build mosaics~\cite{kim2002jigsaw}, apply artistic dithering~\cite{ostromoukhov1999multi}, or place cubist blocks~\cite{collomosse2003cubist}. While these methods are capable of limited styles, follow-up work allows the use of templates or reference images~\cite{zhao2011portrait, pouli2011progressive}. Given the simple explicit library of edits, local edits throughout the image are globally consistent, and such brush strokes or texture abstractions can be propagated through video~\cite{wang2004video, hertzmann2000painterly}.

Learned stylization has significantly expanded the library of available styles by conditioning generation on example images~\cite{isola2017image, li2016precomputed}. More recent diffusion models~\cite{ho2020denoising, saharia2022photorealistic} have improved fidelity and resolution, allowed for stylization~\cite{wang2023stylediffusion}, enabled text-based image editing~\cite{brooks2023instructpix2pix}, and provided generation conditioned on depth, keypoints, or edges when combined with ControlNets~\cite{zhang2023adding}. Latent Diffusion models generate images by progressively denoising a noisy 2D latent map using a U-Net and then decoding the latent into the full image. A ControlNet mirrors the architecture of a diffusion model's U-Net and is trained to influence the denoising process of the diffusion model. It first encodes a control condition and then shares intermediate features across to the diffusion model at every layer during the denoising process.     

\noindent\textbf{Consistent Stylization} 2D stylization can vary dramatically depending on the input image, and so stylization consistency from one frame to another is a challenge. While multiple frames can be generated or edited simultaneously as in video diffusion models~\cite{ho2022video, blattmann2023stable} or 2D-based 3D scene generation~\cite{gao2024cat3d}, these models are often expensive in terms of training time, inference time, and memory usage. There are also difficulties in acquiring suitable 3D training data. The problem is especially prevalent for NVS stylization, where captions may also be required for the data. Some methods including Instruct-NeRF2NeRF and Instruct-GS2GS~\cite{haque2023instruct, igs2gs}, SNeRF~\cite{ nguyen2022snerf}, and VicaNeRF~\cite{dong2024vica} gradually stylize a collection of images by alternating between modifying individual frames and NVS optimization. These methods require lengthy offline processing and produce results with either limited shape alteration or blurry textures. Score Distillation Sampling~\cite{poole2022dreamfusion} has been used to generate 3D scenes from 2D models~\cite{ruiz2023dreambooth, tang2023dreamgaussian, poole2022dreamfusion}, but it often produces hazy results, even when used for stylization as in our early experiments. Other NVS stylization methods achieve multi-view consistency by sharing latent information in intermediate stages through cross-attention as in GaussCtrl~\cite{cao2023masactrl, wu2024gaussctrl}, direct feature injection as in DGE~\cite{chen2024dge, hertz2022prompt}, flow- or depth-based warping~\cite{bao2023latentwarp, fan2024videoshop, feng2024wave}, or warp-friendly noise representations~\cite{chang2024warped}. These methods may suffer by sharing erroneous information from multiple views with complex geometry. In contrast, we use depth-guided feature sharing that respects the scene's geometry and leads to more consistent stylizations. G3DST~\cite{meric2024g3dst} train generalizable modules for stylizing NeRFs but exhibits limited stylization beyond texture changes. 3D scene-level stylization is possible via 3D noise representations as in ConsistDreamer~\cite{chen2024consistdreamer, khalid2023latenteditor}, Gaussian-embedded features ~\cite{liu2024stylegaussian}, or Gaussian primitive tracing~\cite{liu2024stylegaussian}. Notably, ConsistDreamer's stylizations are surface level with local texture modification. In contrast, our method makes dramatic and controllable geometry changes to the source 3DGS scene. 

\section{Method}

\begin{figure*}[h!]
    \centering
    \vspace{-5pt}
    \renewcommand{\tabcolsep}{2pt}
    \small
    \includegraphics[width=\linewidth]{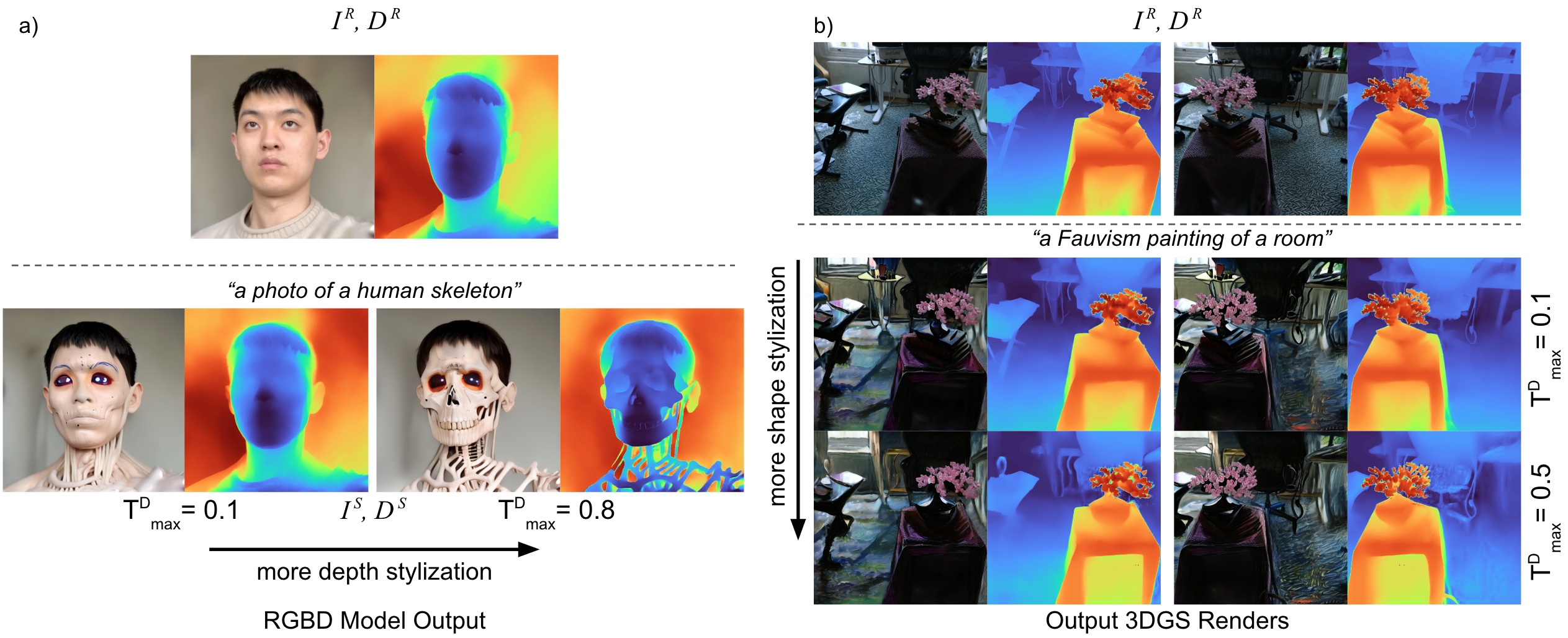}
    
    \vspace{-5pt}
    \caption{For the same prompt, we vary stylization strength for geometry. a) We show the output of our RGBD model for the same stylization prompt but with varying depth stylization strengths. Note how the depths change when we ask for higher depth stylization but the overall color gamut does not. b) We show the effect of shape stylization in output 3DGS models from our method.}
    \vspace{-5pt}
    \label{fig:splat_stylization_strength}    
\end{figure*}

Our method takes as input a depth-regularized 3D Gaussian Splatting (3DGS) model of a scene from which we render a set of RGB and depth images using poses from a representative camera trajectory. It also takes in a stylization prompt and two values indicating the strength of both appearance and geometry stylization. The intermediate output of our model is a set of consistently stylized RGBD frames. We use these stylized frames to train our output stylized 3DGS model. In Section~\ref{sec:rgbd_model} we describe our 2D \textcolor{rgbd_model}{RGBD diffusion model} that allows for appearance and shape stylization strength control using separate denoising of color and depth. When stylizing every new frame in the sequence, we composite the original render and projections of selected previously stylized frames as input to a RGBD-informed \textcolor{warp_controlnet_model}{Warp ControlNet} (Section~\ref{sec:warpingcontrolnet}) to guide stylization of new frames. We use a mixture of \textcolor{feature_sharing}{depth-informed feature-sharing} strategies to share feature information across from stylized frames to the current frame (Section~\ref{sec:depth_informed_sharing}) to encourage consistent stylization. Our method is outlined in Figure~\ref{fig:overview_method}.

% \begin{itemize}
%     % \item $I_0$: image 0
%     % \item $I_i$: image i
%     % \item $I_i^S$: image i stylized
%     % \item $I_i^C$: image i composite
%     % \item $D_i$: depth 
%     % \item $D_i^S$: depth stylized 
%     % \item $D_i^C$: image i composite
%     \item $\mathbf{x}\in \mathbb{R}^{H\times W\times4}$: RGBD tensor in diffusion
%     \item $\epsilon$ noise 
%     \item $t$ time step
%     \item $Q, K, V$ attention tensors
%     \item $\text{Attn}$ attention
%     \item $P_i$: projection matrix  
% \end{itemize}

\subsection{\textcolor{rgbd_model}{Geometry and Appearance Stylization}}\label{sec:rgbd_model}
Our RGBD diffusion model takes as input an image render $I_i^R$, the rendered depth map $D_i^R$, a prompt, and style strength parameters for both depth and color, and outputs stylized color $I_i^S$ and depth $D_i^S$. Following previous diffusion stylization methods, during inference we progressively apply noise to our inputs before denoising with prompt conditioning. Specifically, we use DDIM inversion~\cite{mokady2023null} on image and depth latents. In particular, starting at diffusion time $t=0$, we first add Gaussian noise to get to time $t=T_\text{noise}$, and then use DDIM inversion to get to noise level $T_\text{max}$. This latent is then denoised with the network conditioned on the target prompt. 
%This use of noise for the first few steps prior to DDIM inversion allows us to guard against overfitting on the high-frequency components of the rendered RGBD, since forward process tends to destroy high-frequency components first.

We wish to control the stylization strength of color and depth separately, while still editing these channels in a manner that keeps them consistent with one another. To do this, we denoise both of them simultaneously and modify the noise schedules for each respective channel by introducing two separate maximum timesteps for adding noise, $T^D_\text{max}$ and $T^I_\text{max}$ for depth and color respectively. During denoising, we do not permit the network to change the depth until $t < T^D_\text{max}$, nor to change the RGB channels until $t < T^I_\text{max}$. Inspired by diffusion inpainting methods~\cite{rombach2021highresolution}, we pass in masks $M^D_t$ and $M^I_t$, consisting of ones if the current noising/denoising timestep satisfies the condition $t \leq T^{I,D}_\text{max}$ and $0$ otherwise. This allows us to inform the network of whether its changes to the RGB and D channels will be accepted or not on a given denoising step, just as the mask passed into inpainting models informs the network of whether its changes to a given pixel will be accepted. Since we predict scale-invariant depth, we scale the stylized depth map, $D_i^S$, from the RGBD model using the rendered depth $D_i^R$. We show examples of variable-depth stylization in Fig.~\ref{fig:splat_stylization_strength}.

\subsection{\textcolor{warp_controlnet_model}{Warp ControlNet for Consistent Inpainting}}\label{sec:warpingcontrolnet}
We wish to propagate previous frames' stylization when stylizing new frames. We forward-warp previously stylized frames using their depths $D_{j}^S$ to the current frame, $I_i$, and compute warped frames $I_{ij}^S$, warped depths $D_{ij}^S$, and validity masks $M_{ij}^S$ where $j$ is the index of a previously stylized frame. For each warped reference frame, we composite it with the unstylized current frame to get $I_{ij}^C, D_{ij}^C$. 
% A warped reference view can compete with the unstylized current frame for the same location in the composited frame. While there are existing tactics for collapsing along these features, including using large U-Nets~\cite{hedman2018DeepBlending}, we instead rely on a score derived using frame age and the angle to warped geometry to reduce texture stretch artifacts from large baseline warps and modify masks to be $\hat{M}_{ij}^S$ accordingly. We elaborate on this selection strategy in the supplemental. 
% We compute an overall validity mask as the union of all masks, ${M_i = \bigcup_{j \in N} M_{ij}}$. Our composite color and depths are:
% \begin{equation}
% \centering
%     I_i^C, D_i^C  = \sum_j{(\{I_{ij}^S, D_{ij}^S\} * \hat{M}_{ij})} + \{I_i^R, D_i^R\}*(1-M_i)
% \end{equation}
% \ms{Equation is TMI.} 

%Naively inpainting stylized frames with inpainted missing regions denoted by $M_{ij}$ produces artifacts and leaves texture stretching artifacts unresolved. Instead, we use a specifically-trained custom ControlNet~\cite{zhang2023adding} conditioned on the composites $I_{ij}^C, D_{ij}^C$, the composite mask $M_{ij}^S$, and the input prompt, guiding our RGBD diffusion model to produce harmonized and stylized color $I_i^S$ and depth $D_i^S$. To support multiple frames, we average guidance features from each pass of the ControlNet on every composite from each frame $j$. We elaborate on training in Section~\ref{sec:training}.

%Naively warping a stylized frame and then inpainting missing regions leaves
A naive approach to generating a new frame conditional on a previously stylized frame would be to warp the stylized previous frame, and then inpaint missing regions. However, this leaves warping artifacts. Instead, we create a specifically-trained custom ControlNet~\cite{zhang2023adding} conditioned on the composites $I_{ij}^C, D_{ij}^C$, the composite mask $M_{ij}^S$, and the input prompt. This guides the RGBD diffusion model to correct artifacts in the warped region, and inpaint the rest of the image consistently with warped regions. To support reference frames, we average guidance features from each pass of the ControlNet on every composite from each reference frame $j$. We elaborate on the model training in Section~\ref{sec:training}.

\subsection{\textcolor{feature_sharing}{Depth-Informed Information Sharing}}\label{sec:depth_informed_sharing}
Forward warping of RGB and depth pixels does not preserve fine textures, and does not capture deep features used in previous frames, which our ControlNet does not have access to. To that end, we utilize both feature injection~\cite{tumanyan2023plug} and cross-attention~\cite{cao2023masactrl, wu2024gaussctrl}. These mechanisms help inform layers of the network of how reference frames were stylized. However, cross-attending and injecting features across all image patches might lead to undesirable effects such as duplicated or misplaced aesthetic features, see Figure~\ref{fig:ablations_figure}. While previous work uses epipolar lines to guide this process~\cite{chen2024dge}, we use depth information to more precisely transfer feature information from reference frames to the current frame. We start by building 4D heatmaps $L_{ij}$ computed by forward-warping the reference frame depth $D_j^S$ to the current frame. We use a forward warp to guide the transfer of features from reference to source frames by (a) increasing the strength of cross-attention where a reference frame pixel forward-warps to a target-frame pixel, and (b) directly injecting features from reference-frame pixels to corresponding target-frame pixels. We show this visually in Figure~\ref{fig:feature_sharing}.

We modify the diffusion model's self-attention layers to allow the image to attend to both itself (as in the unmodified self-attention) and to the reference images, by concatenating together the keys from the original image and the reference images. Our cross-attention then becomes: 

\begin{equation}
\centering
    \text{softmax} \left( \frac{Q[K_{\text{self}}, K_{\text{ref}}]^T}{\sqrt{d_k}} + \log \Delta \right) [V_{\text{self}}, V_{\text{ref}}]
    \label{eq:xattn}
\end{equation}
where $\Delta$ is a mask which we use to control the amount of attention between each key-query pair. When the key is in $K_\text{self}$, $\Delta$ is a constant $\lambda_\text{self}$ (which we set to 0.5 everywhere). Where the key is in $K_\text{ref}$, it is equal to $L_{ij}$, allowing us to modulate the strength of the cross-attention based on our knowledge of the geometry.
%\begin{equation}
%M = 
%\begin{cases} 
%\log \lambda_{\text{self}} & \text{for } K_{\text{self}} \\ 
%\log L_{ij} & \text{for } K_{\text{ref}},
%\end{cases}
%\end{equation}
%where $f_{ij}$ is large for key-value pairs which correspond to the same part of the scene. In particular, it is larger when the pixel location of the key token projects close to the pixel location of the value token, given their relative pose and the stylised depth map associated with the reference image. We give the exact definition of $f_{ij}$ in the supplemental. $\lambda_\text{self}$ is a constant which controls the strength of the usual self-attention relative to the cross-attention; we use $\lambda_\text{self} = 0.5$ everywhere.

%At each timestep of the denoising process, it is desirable to use a reference image at the same timestep; otherwise we are injecting features with the wrong amount of noise into the new image. For this reason, we store intermediate latents after each step of the denoising process. Then, at timestep $t$ of image generation, we retrieve the cached reference latents from the corresponding timestep.

During denoising, it is desirable to use a reference image at the same noise level as the image being generated. For this reason, we cache intermediate latents for all frames. At each denoising timestep, we then retrieve the cached reference frame latents from the corresponding timestep and use them for feature-sharing.

\begin{table*}[t]
    \renewcommand{\tabcolsep}{10pt}
    \centering
    \small
    % uncomment this line if the table becomes too wide to fit
    \resizebox{1.0\textwidth}{!} 
    {

    \begin{tabular}{|l|cc|cc|c|}
    \hline
        & \multicolumn{2}{c|}{CLIP} & \multicolumn{2}{c|}{Image Consistency} & Time\\
    \hline
        & CLIP Direction Similarity ↑  & CLIP Direction Consistency ↑ & RMSE ↓ & LPIPS ↓ &\\
    \hline
    Instruct NeRF2NeRF~\cite{haque2023instruct} & .098 & .531 & .0463 & .0540 & $\sim$ hours\\
    Instruct GS2GS~\cite{vachha2024instruct} & .097 & .519 & .0501 & \textit{.0403}  & $\sim$ 30 minutes\\
    GaussCtrl~\cite{wu2024gaussctrl} & \textit{.123} & \textit{.590} & .0471 & .0438 & $\sim$ 10 minutes\\
    DGE~\cite{chen2024dge} & .113 & .565 & \textit{.0384} & .0407 & $\sim$ 10 minutes\\
    \textbf{Ours} & \textbf{.175} & \textbf{.606} & \textbf{.0370} & \textbf{.0378} & $\sim$ 10 minutes\\
    \hline

        % {
    %\begin{tabular}{|l|cc|cc|c|}
    %    \hline
    %        & \multicolumn{2}{c}{CLIP} & \multicolumn{2}{|c|}{Image Consistency} & \multicolumn{1}{c|}{Avg Editing Time} \\
    %    % \cmidrule(l){2-3} \cmidrule(l){4-6}  \cmidrule(l){7-8} 
    %    \hline
    %        & CLIP Direction Similarity ↑  & CLIP Direction Consistency ↑  & LPIPS ↓	& RMSE ↓	& \\
    %    \hline
    %% VicaNeRF~\cite{dong2024vica} & & & & & \\
    %Instruct NeRF2NeRF~\cite{haque2023instruct} & 0.113 & 0.594 & & 0.0471 & $\sim$hours\\
    %Instruct GS2GS~\cite{vachha2024instruct} & 0.112 & 0.569 & & \textit{0.0321} & $\sim$10s of minutes\\
    %GaussCtrl~\cite{wu2024gaussctrl} & \textit{0.139} & \textit{0.629} & & 0.0355 & $\sim$minutes\\
    %DGE~\cite{chen2024dge} & 0.138 & 0.606 & & 0.0342 & $\sim$minutes\\
    %\textbf{Ours} & \textbf{0.177} & \textbf{0.633} & & \textbf{0.0301} &  $\sim$minutes\\
    %    \hline
    %\end{tabular}
    \end{tabular}
    }
    \vspace{-5pt}
    \caption{\textbf{Quantitative Evaluation} We compute metrics for 53 stylizations on a range of published and new scenes. CLIP Direction Similarity measures how well the stylization implied by the prompt is respected. CLIP Direction Consistency measures how consistent this stylization is across frames. We also compute image consistency scores from~\cite{feng2025wave}. Ours outperforms other novel-view stylization methods.}
    \vspace{-10pt}
    \label{tab:main_scores}
\end{table*}

For feature injection, we use an argmax across reference features on $L_{ij}$ to select which reference feature to inject into the network. We do not perform feature injection in the first two and last three layers since we only want higher order semantic information and to prevent texture artifacts.

\begin{figure}[h!]
    \centering
    \vspace{-5pt}
    \renewcommand{\tabcolsep}{2pt}
    \small
    \includegraphics[width=0.8\linewidth]{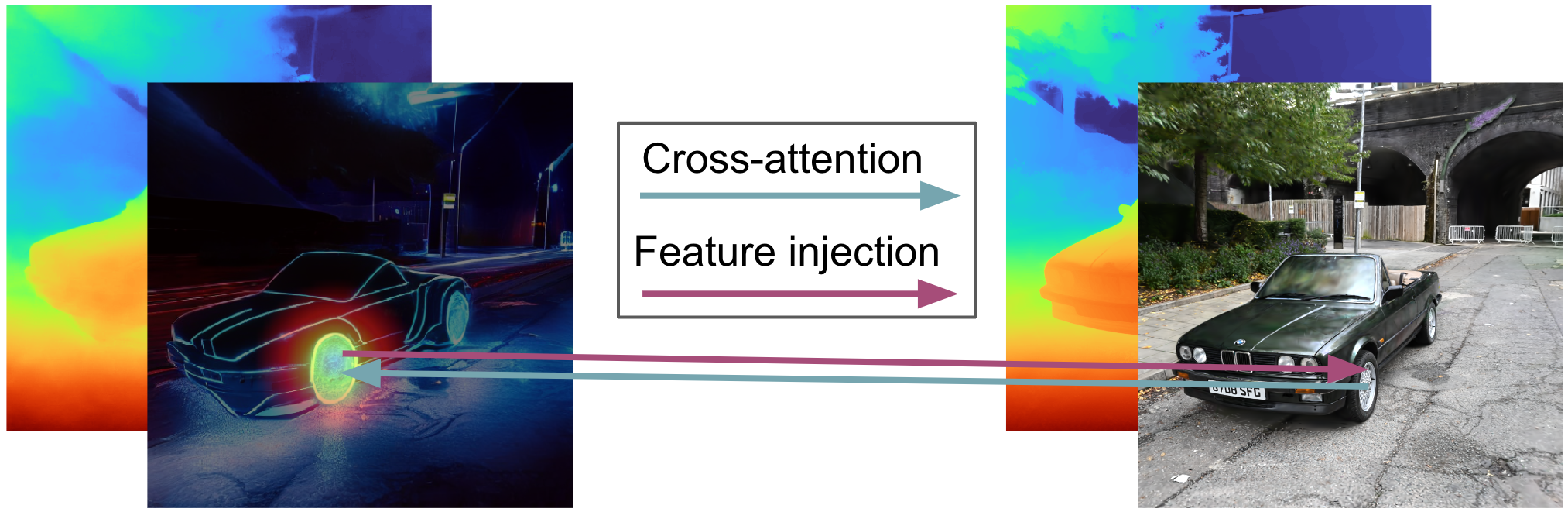}
    
    \vspace{-5pt}
    \caption{\textbf{Feature information sharing} We show a slice through the heatmap $L$ for a single pixel in the target frame.}
    \vspace{-5pt}
    \label{fig:feature_sharing}    
\end{figure}

\section{Implementation Details}
\subsection{Frame Selection, Warping, and Resolution}
Our pipeline takes as input and produces output images at a resolution of $512\times512$. We select a smooth representative trajectory through the 3DGS, so that pose changes from each frame to the next are not too extreme.

We obtain the first frame by running the RGBD model without the ControlNet. For each reference frame, we warp it to the next frame to be generated and composite it with the unstylized new frame, forming the input to our ControlNet. To warp, we construct a mesh by backprojecting stylized depth maps to create vertices, creating mesh edges between neighboring pixels, and cliping edges using a normals check w.r.t the camera look-at. We then render this mesh to the current view using PyTorch3D~\cite{ravi2020pytorch3d}. All results are generated using warps of and featuring sharing from the first and last stylized frames.

\begin{figure}[h!]
    \centering
    \vspace{-5pt}
    \renewcommand{\tabcolsep}{2pt}
    \small
    \includegraphics[width=\linewidth]{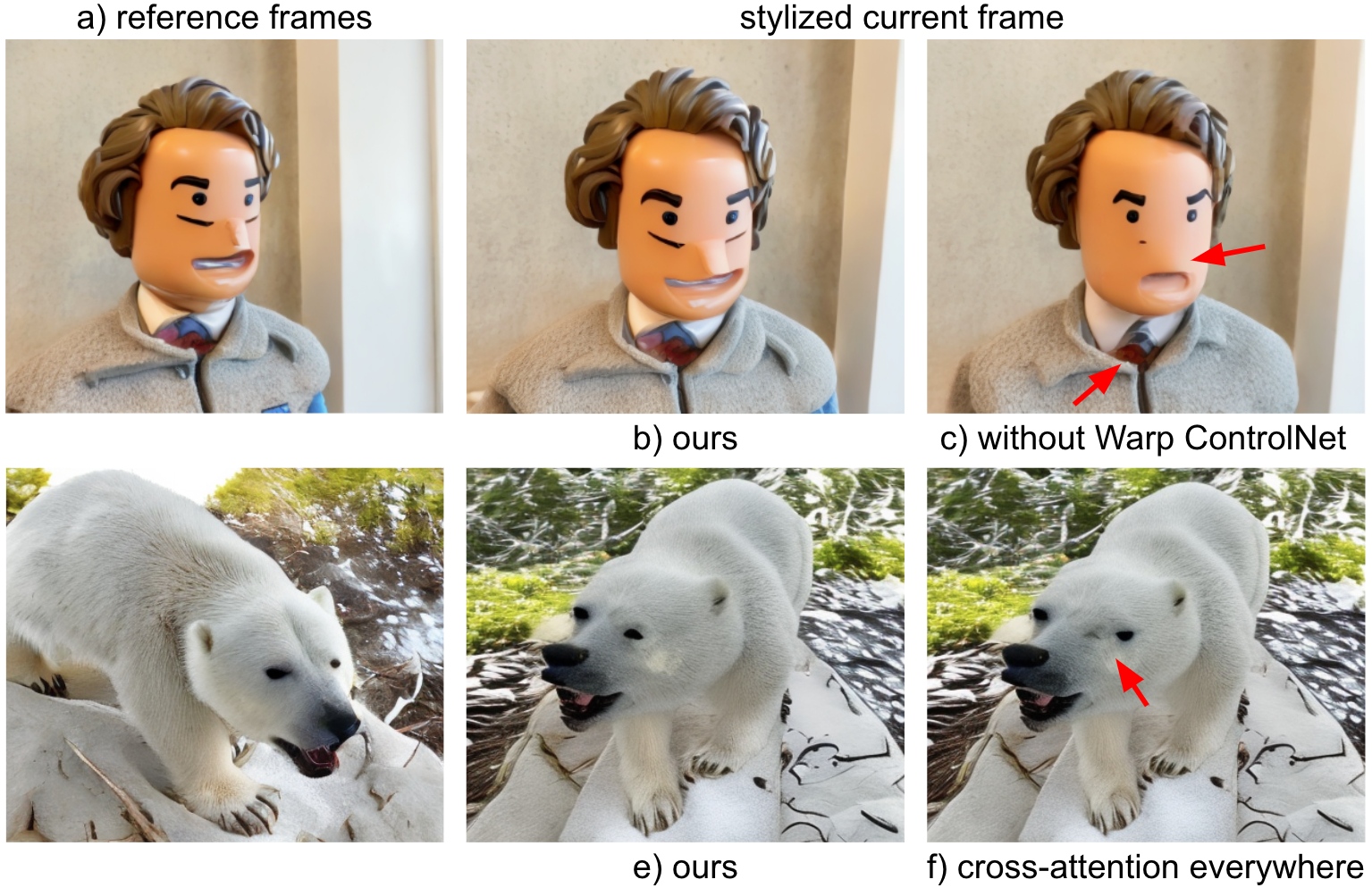}
    
    \vspace{-5pt}
    \caption{\textbf{Qualitative Ablations} We show two a) reference frames, the results of style propagation using ours in b) and e), and then ablations in c) and f). c) without our \textcolor{warp_controlnet_model}{Warp ControlNet}, the geometry and texture on the face and tie are not propagated correctly. f) by cross attending everywhere across the entirety of the frame without \textcolor{feature_sharing}{depth-informed feature sharing}, patches like the bear's eye may be misplaced or repeated leading to inconsistency.}
    \vspace{-10pt}
    \label{fig:ablations_figure}    
\end{figure}

\subsection{Architecture and Training} \label{sec:training}
For our RGBD latent model, we initialize with StableDiffusion 2.1~\cite{rombach2021highresolution} as a base. We encode both RGB and depth images separately. Following~\cite{ke2023repurposing}, we encode the depth map by stacking it three times along the channel dimension and using the image encoder. For our denoising U-Net~\cite{rombach2021highresolution}, the input channels are four encoded features channels and one time-mask channel for each of depth and RGB. We train our RGBD model on 500k images from the aesthetic subset of the Re-LAION-2B dataset~\cite{schuhmann2022laion}. For supervising our model's depth output we generate depth maps for every training-set image using a combination of three off-the-shelf depth models to prevent overfitting to the characteristics of any particular depth model: Depth Anything V2~\cite{depthanything}, Marigold~\cite{ke2024repurposing}, and GeoWizard~\cite{fu2025geowizard}. We apply Gaussian blur to these depth maps 75\% of the time. We use noise scheduler strategies from~\cite{lin2024common}.

Our ControlNet model is based on the original implementation~\cite{zhang2023adding}. We change the hint input channel size to accommodate the 5 channels of our composited RGBD image and associated mask. We also expand the channel sizes of the hint network from $(16, 32, 96, 256)$ to $(48, 96, 192, 384)$ to allow it to better understand the stylization and compositing problem in our input. To train this ControlNet, we generate pairs of training data by predicting depth maps for RGB images, stylizing those frames using our RGBD model (which we train before the ControlNet), warping those stylized frames to an arbitrary camera, and warping back. For our training set, we generate 250k pairs from our RGBD model using 10k prompts. We sample 6 random camera transforms for each. See the supplemental for further details.

\subsection{3DGS Optimization}
We optimize a 3D Gaussian Splatting (3DGS) model for every scene before rendering trajectories for use in our method. To produce 3DGS models with good depth renders, we use depth and normal regularization. We first run a monocular depth model, Metric3D~\cite{yin2023metric}, on each training view, render a depth map from the 3DGS at that view, and median-scale the rendered depth map using the Metric3D depth prediction. We compute a depth loss using a scale-invariant loss~\cite{eigen2014depth}. We compute dot-product and cosine-similarity losses on normal maps from both depth maps made using cross products on local image gradients~\cite{sayed2022simplerecon}. 

The final stage of our pipeline is training a new 3DGS model using stylized output color and depth maps from our method. In this instance, we regularize using our method's depth predictions, as well as normals derived from them. We also use a Total Variation L1 (TVL1)~\cite{chan2005aspects} loss on rendered normals from depth as regularization to reduce floating artifacts. See the supplemental for details.

\section{Experiments}
We evaluate our method both quantitatively (Table~\ref{tab:main_scores}), qualitatively (Figure~\ref{fig:qualitative-comparison}), and with a user study. We evaluate on scenes from Instruct-NeRF2NeRF~\cite{haque2023instruct}, GaussCtrl~\cite{wu2024gaussctrl}, ScanNet++~\cite{yeshwanthliu2023scannetpp}, Mip-NeRF360~\cite{barron2021mip}, and our new scenes. For all baselines, we use official code releases.

\noindent\textbf{Evaluation Frame Selection} 
%The evaluated methods may use a different selection of views for stylization that are often a subset of all available training views used to optimize the unstylized 3D Gaussian Splatting Model (3DGS). This means that their results may lack areas and detail present in the original 3DGS. This would lead to unfair evaluation if the original training views are used for evaluation, especially when a scene is large and complicated with thorough capture. We create a representative trajectory for each scene and 
Our method uses a smooth camera trajectory through the splat, whereas our baselines all take an unordered set of sparse frames (for which we use the original training views of the splat). In order to fairly evaluate our method against our baselines, we find nearest neighbor poses between our trajectory and the original training views used by our baselines. We interpolate halfway between every pair of views. We use these views for evaluating quantitatively and for the user study.

\begin{figure}[h!]
    \centering
    \vspace{-10pt}
    \renewcommand{\tabcolsep}{2pt}
    \small
    \includegraphics[width=\linewidth]{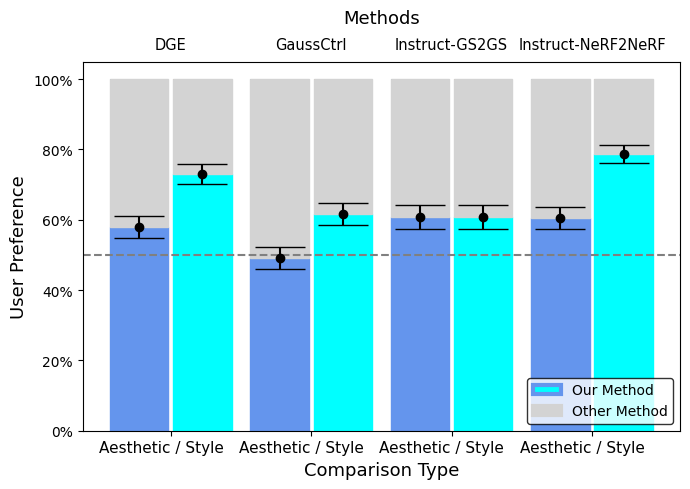}
    
    \vspace{-10pt}
    \caption{\textbf{A/B User Study} 31 participants consistently preferred our method's adherence to \textbf{style} prompts and found it to either match or exceed other methods in \textbf{aesthetic} quality.}
    \vspace{-10pt}
    \label{fig:a_b_study}
\end{figure}

\begin{figure*}
    \vspace{-5pt}
    \centering
    \small
    
    \setlength\tabcolsep{0pt}
\renewcommand{\arraystretch}{0}
\begin{tabular}{ccccccc}
    \centering

     & Original & GaussCtrl~\cite{wu2024gaussctrl} & Instruct-GS2GS~\cite{vachha2024instruct} & Instruct-N2N~\cite{haque2023instruct} & DGE~\cite{chen2024dge} & Ours \\

    \raisebox{1.0\height}{\parbox[t]{12mm}{\rotatebox[origin=c]{90}{\makecell{a photo of a \\ polar bear in \\ the winter forest}}}}
    & \includegraphics[width=0.151\textwidth]{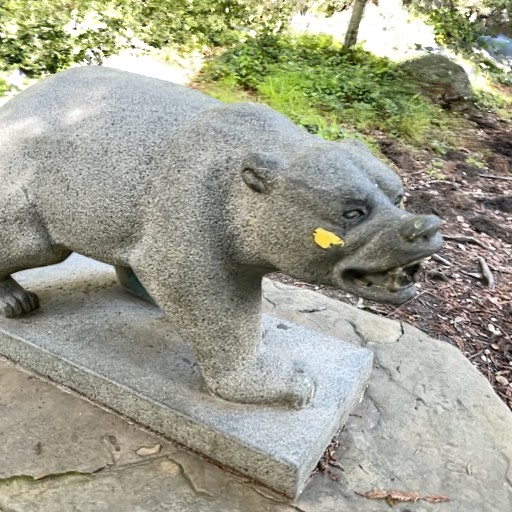}
    & \includegraphics[width=0.151\textwidth]{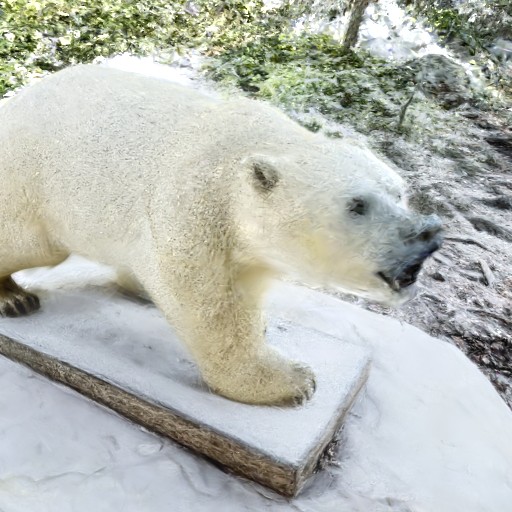}
    & \includegraphics[width=0.151\textwidth]{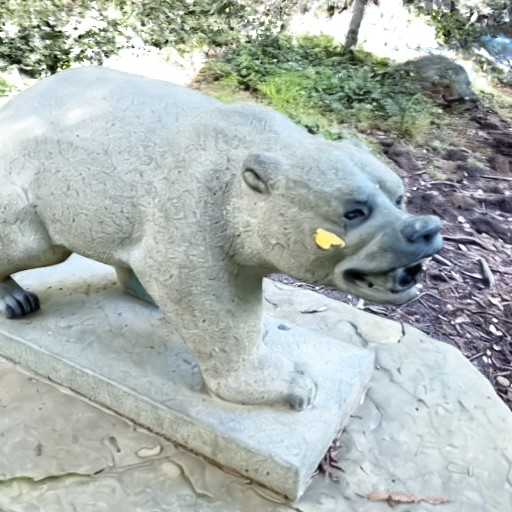}
    & \includegraphics[width=0.151\textwidth]{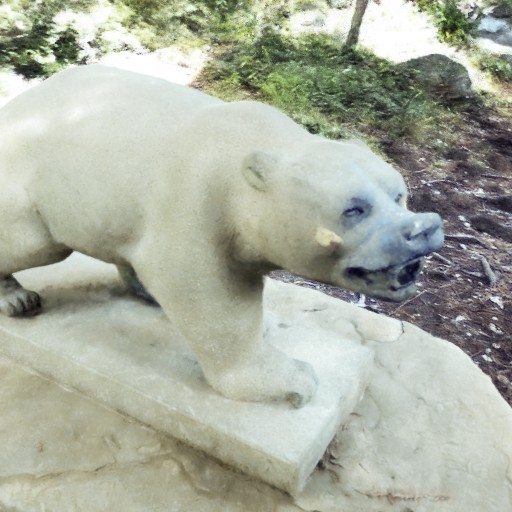}
    & \includegraphics[width=0.151\textwidth]{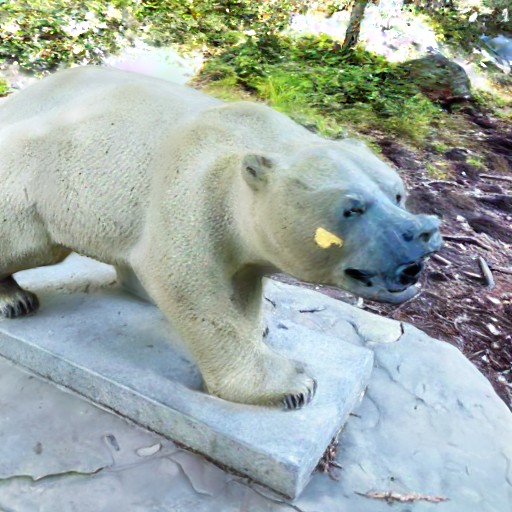}
    & \includegraphics[width=0.151\textwidth]{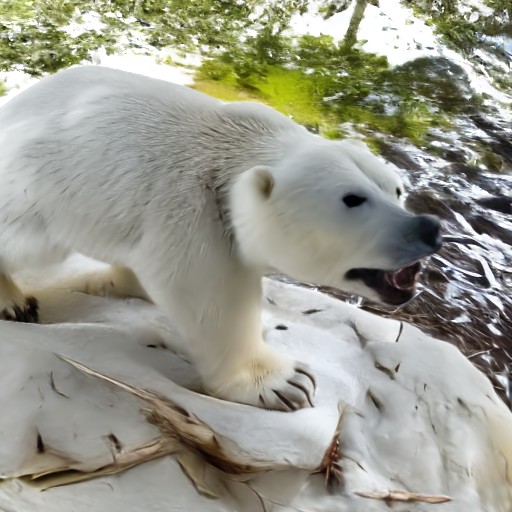}
    \\

    \raisebox{1.0\height}{\parbox[t]{8mm}{\rotatebox[origin=c]{90}{\makecell{a photo of a \\ human skeleton}}}}
    & \includegraphics[width=0.151\textwidth]{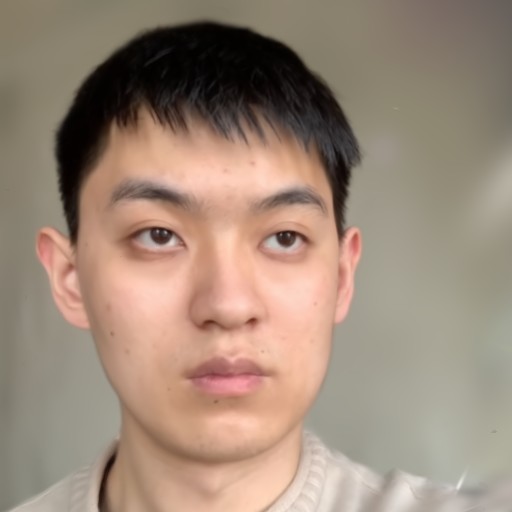}
    & \includegraphics[width=0.151\textwidth]{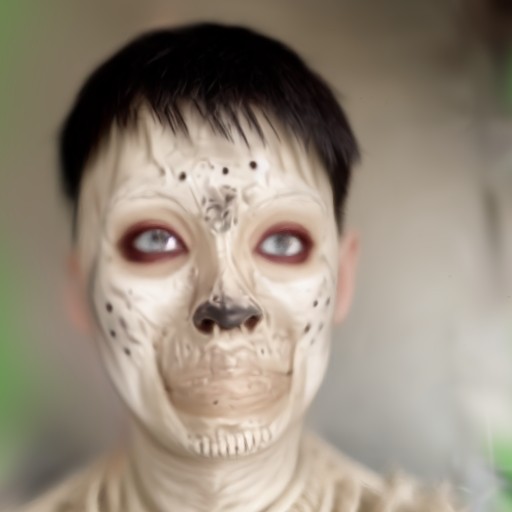}
    & \includegraphics[width=0.151\textwidth]{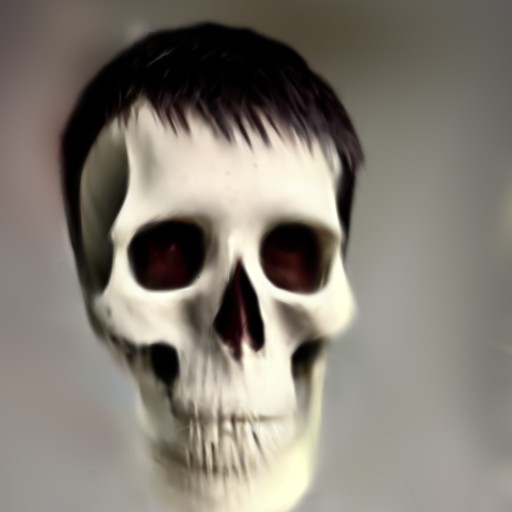}
    & \includegraphics[width=0.151\textwidth]{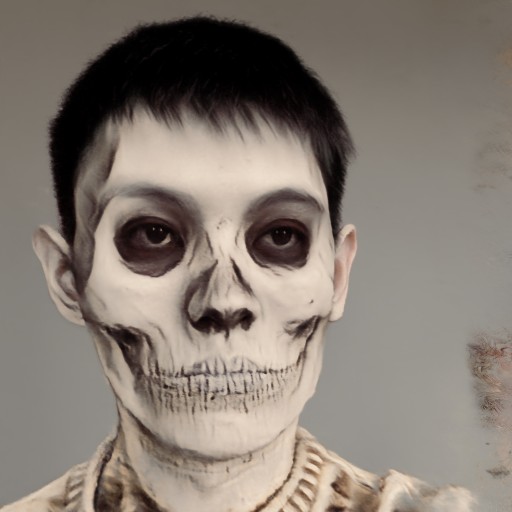}
    & \includegraphics[width=0.151\textwidth]{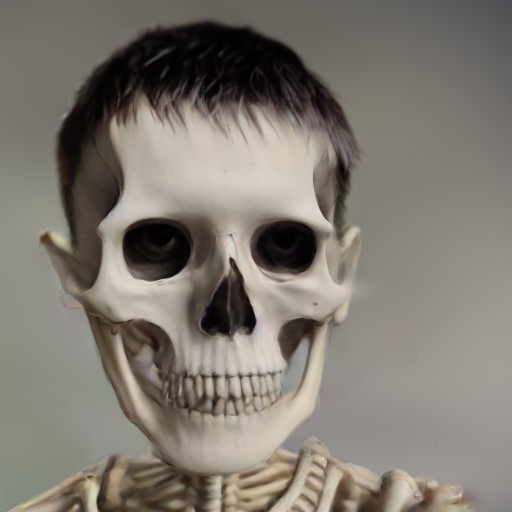}
    & \includegraphics[width=0.151\textwidth]{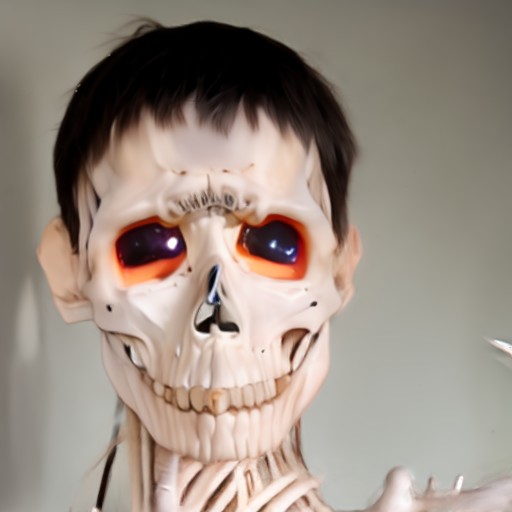}
    \\

    \raisebox{1.3\height}{\parbox[t]{8mm}{\rotatebox[origin=c]{90}{\makecell{a photo of a \\ chimp}}}}
    & \includegraphics[width=0.151\textwidth]{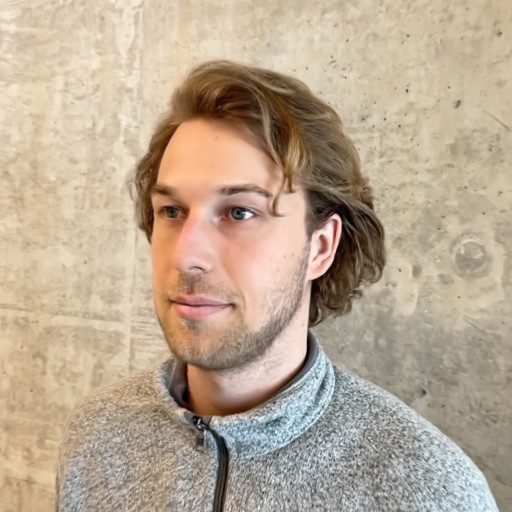}
    & \includegraphics[width=0.151\textwidth]{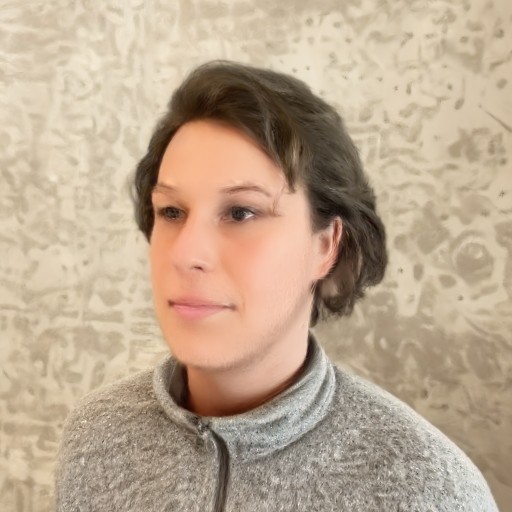}
    & \includegraphics[width=0.151\textwidth]{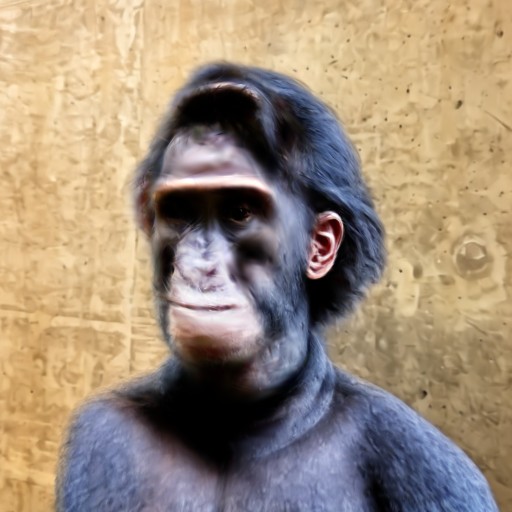}
    & \includegraphics[width=0.151\textwidth]{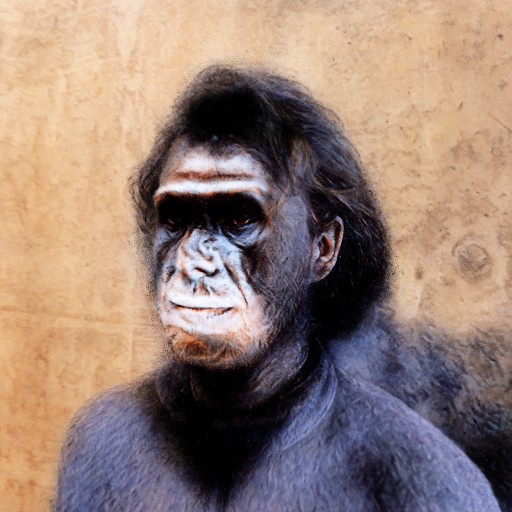}
    & \includegraphics[width=0.151\textwidth]{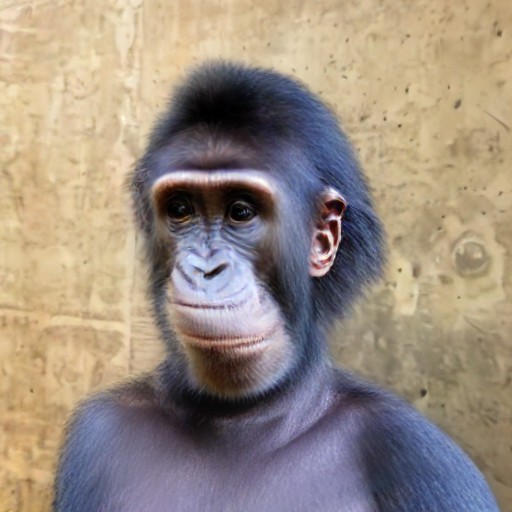}
    & \includegraphics[width=0.151\textwidth]{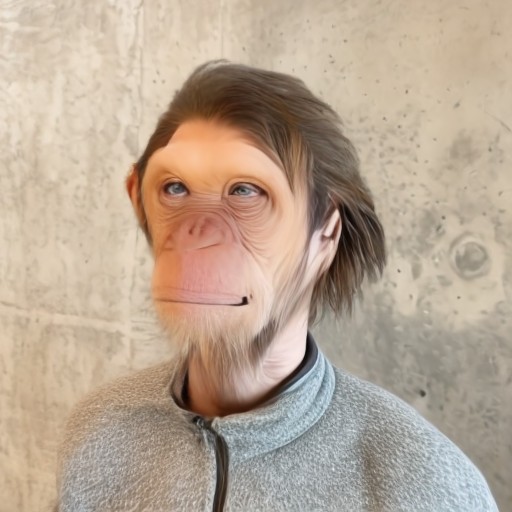}
    \\

    \raisebox{1.0\height}{\parbox[t]{8mm}{\rotatebox[origin=c]{90}{\makecell{a photo of the \\ face of the Hulk}}}}
    & \includegraphics[width=0.151\textwidth]{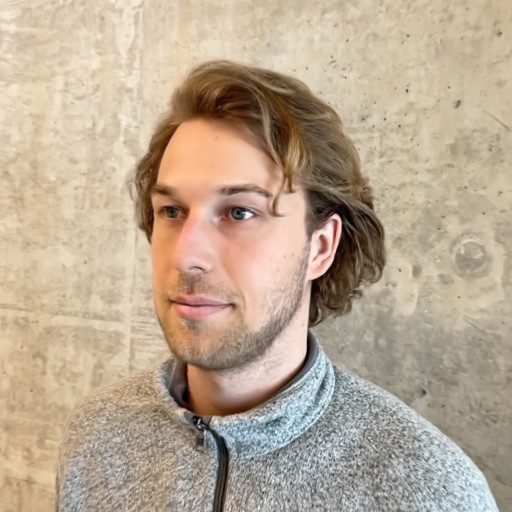}
    & \includegraphics[width=0.151\textwidth]{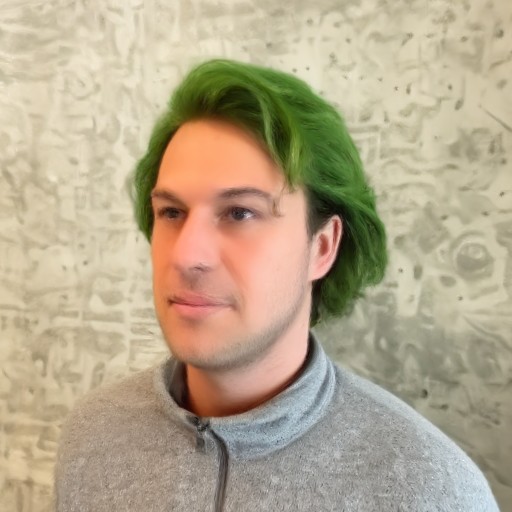}
    & \includegraphics[width=0.151\textwidth]{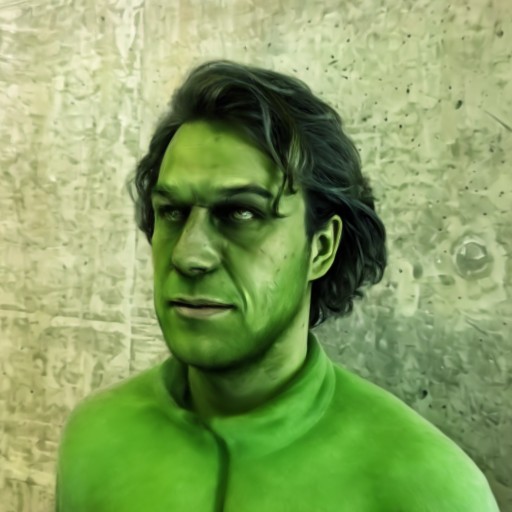}
    & \includegraphics[width=0.151\textwidth]{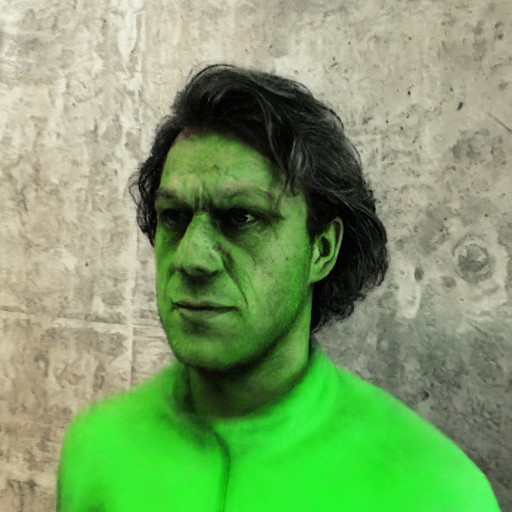}
    & \includegraphics[width=0.151\textwidth]{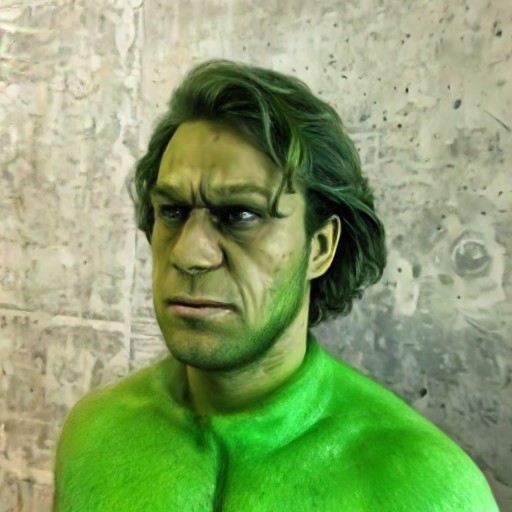}
    & \includegraphics[width=0.151\textwidth]{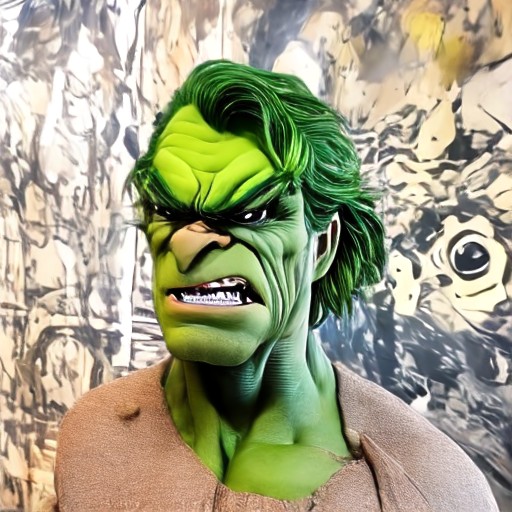}
    \\

    \raisebox{1.0\height}{\parbox[t]{12mm}{\rotatebox[origin=c]{90}{\makecell{a photo of a \\ fountain in the \\ desert}}}}
    & \includegraphics[width=0.151\textwidth]{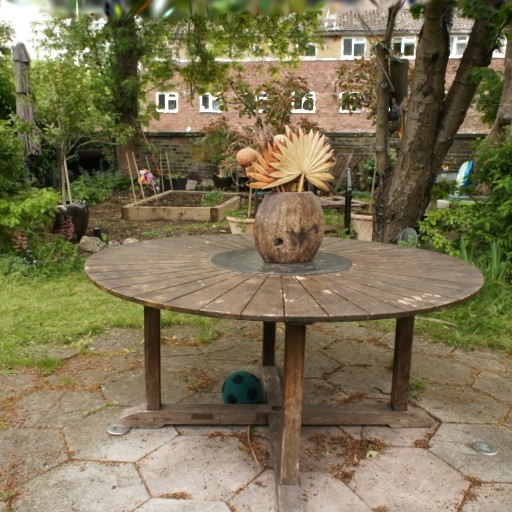}
    & \includegraphics[width=0.151\textwidth]{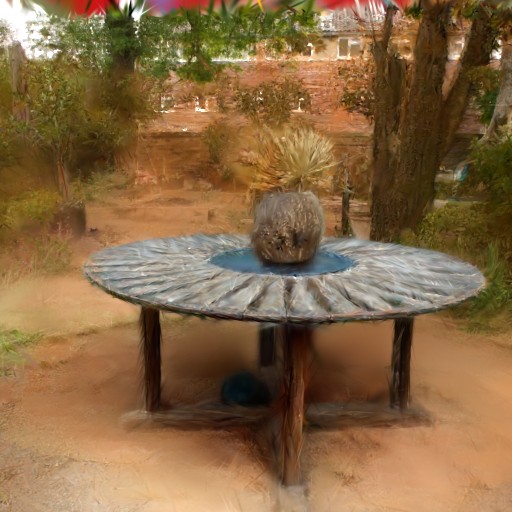}
    & \includegraphics[width=0.151\textwidth]{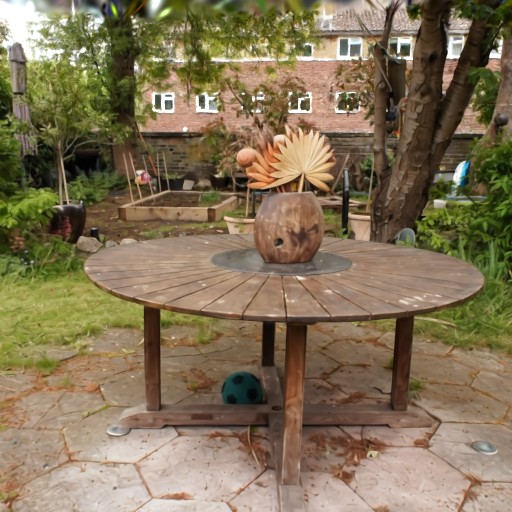}
    & \includegraphics[width=0.151\textwidth]{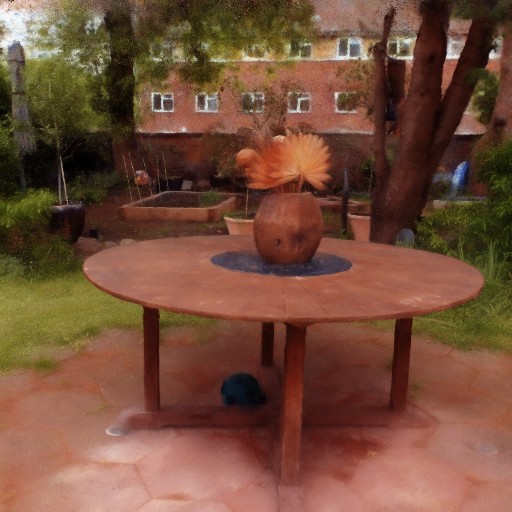}
    & \includegraphics[width=0.151\textwidth]{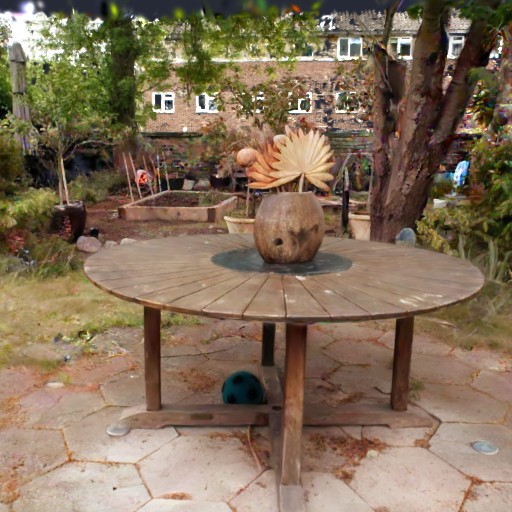}
    & \includegraphics[width=0.151\textwidth]{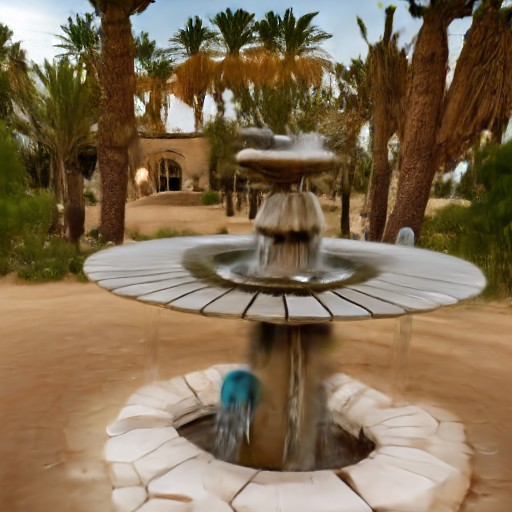}
    \\

    \raisebox{1.2\height}{\parbox[t]{12mm}{\rotatebox[origin=c]{90}{\makecell{Minimalist \\ Japanese tea \\ room  }}}}
    %with \\ tatami mats, \\ low wooden \\ tables, and \\ paper lanterns
    & \includegraphics[width=0.151\textwidth]{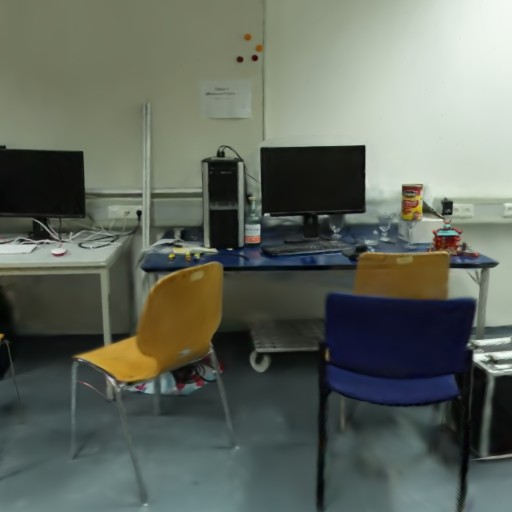}
    & \includegraphics[width=0.151\textwidth]{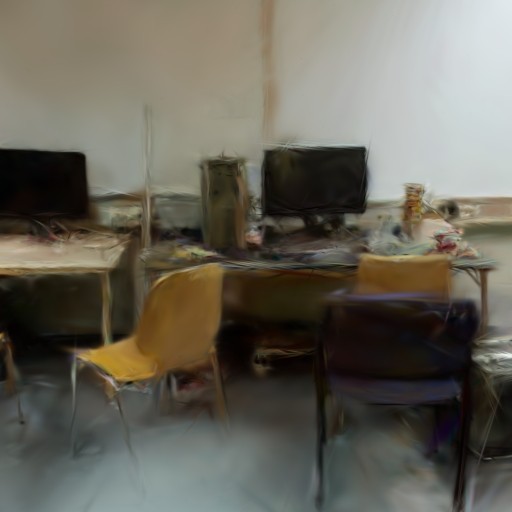}
    & \includegraphics[width=0.151\textwidth]{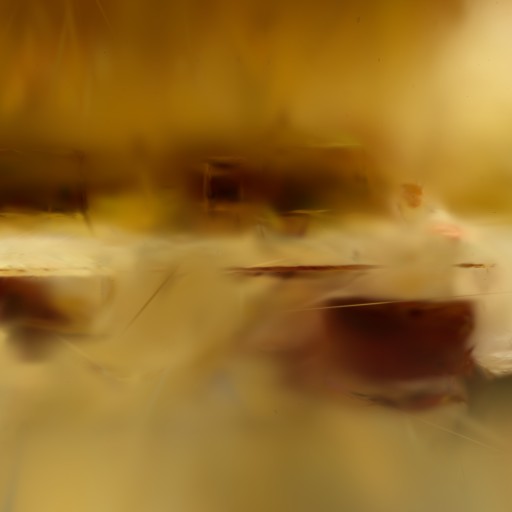}
    & \includegraphics[width=0.151\textwidth]{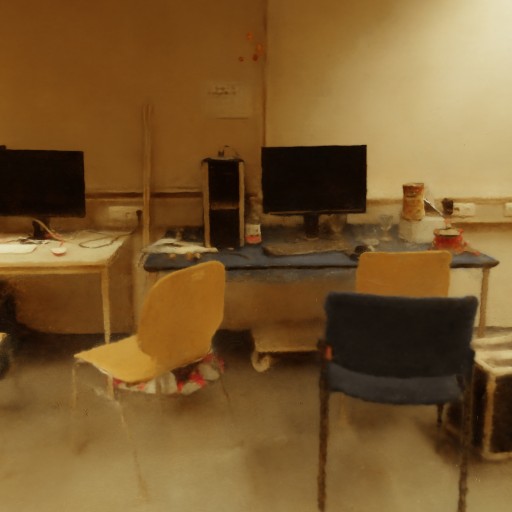}
    & \includegraphics[width=0.151\textwidth]{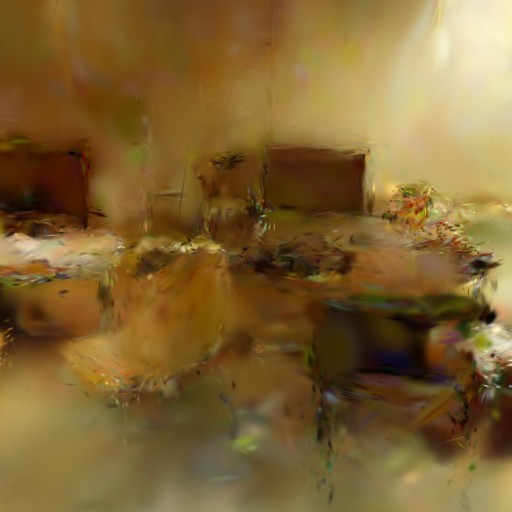}
    & \includegraphics[width=0.151\textwidth]{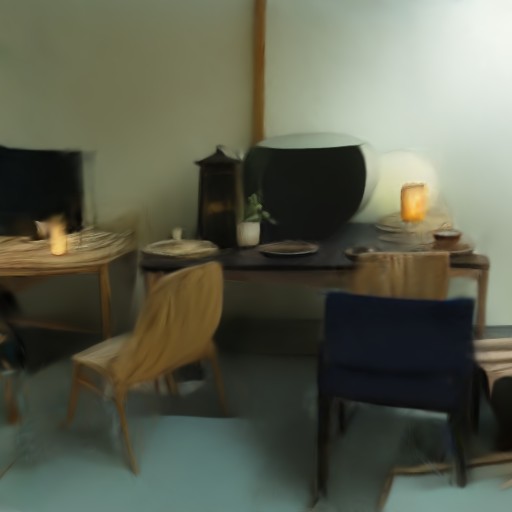}
    \\

    \raisebox{1.05\height}{\parbox[t]{8mm}{\rotatebox[origin=c]{90}{\makecell{Cubist portrait \\ of a horse}}}}
    & \includegraphics[width=0.151\textwidth]{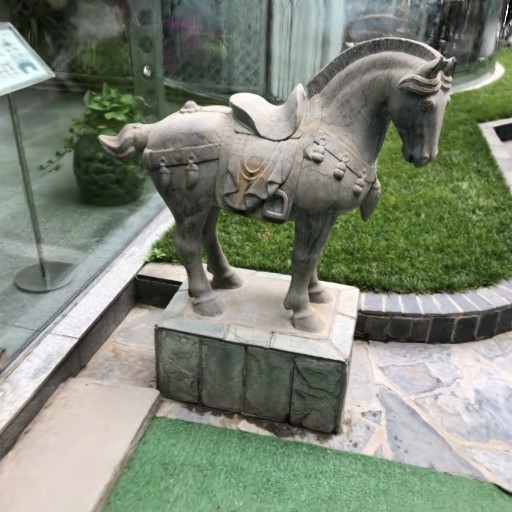}
    & \includegraphics[width=0.151\textwidth]{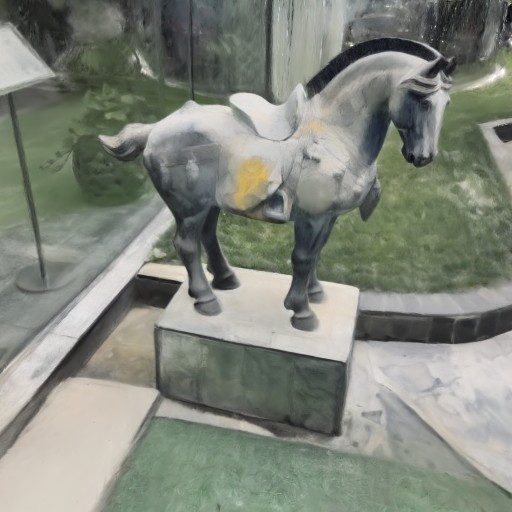}
    & \includegraphics[width=0.151\textwidth]{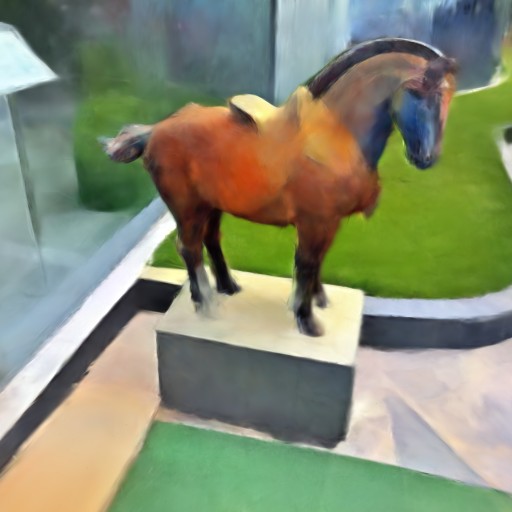}
    & \includegraphics[width=0.151\textwidth]{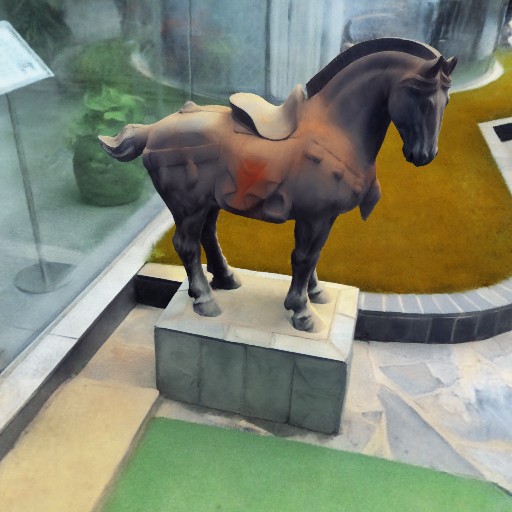}
    & \includegraphics[width=0.151\textwidth]{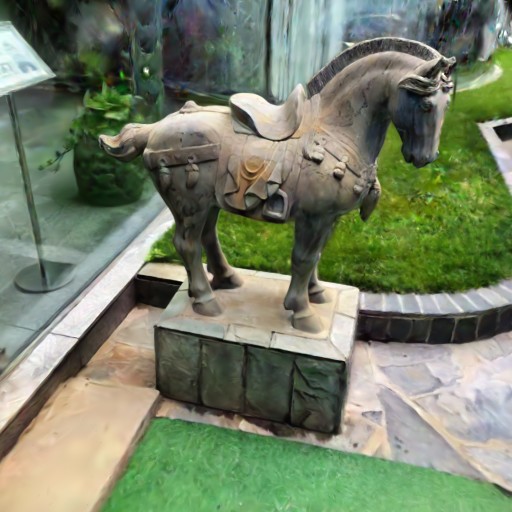}
    & \includegraphics[width=0.151\textwidth]{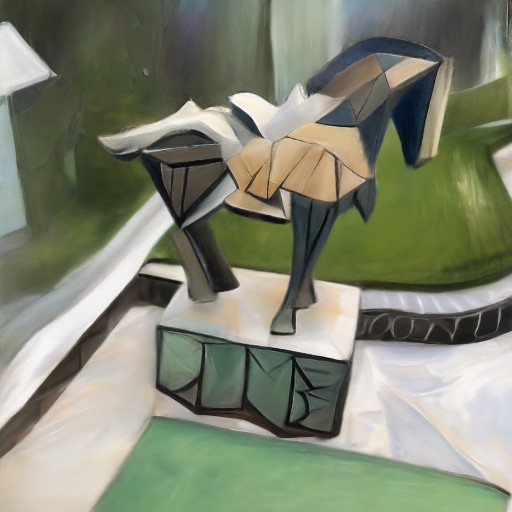}
    \\

    \raisebox{0.85\height}{\parbox[t]{12mm}{\rotatebox[origin=c]{90}{\makecell{Horse statue \\ made from stacked \\ wooden blocks}}}}
    % \\ blocks, rustic \\ and \\ minimalistic
    & \includegraphics[width=0.151\textwidth]{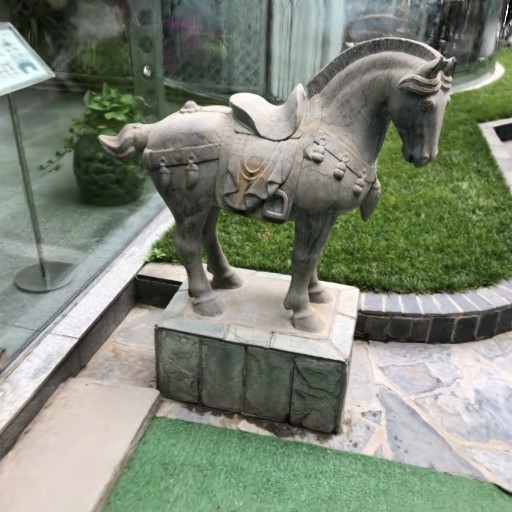}
    & \includegraphics[width=0.151\textwidth]{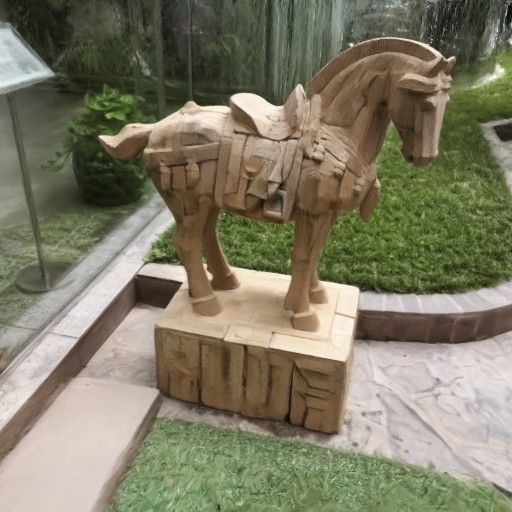}
    & \includegraphics[width=0.151\textwidth]{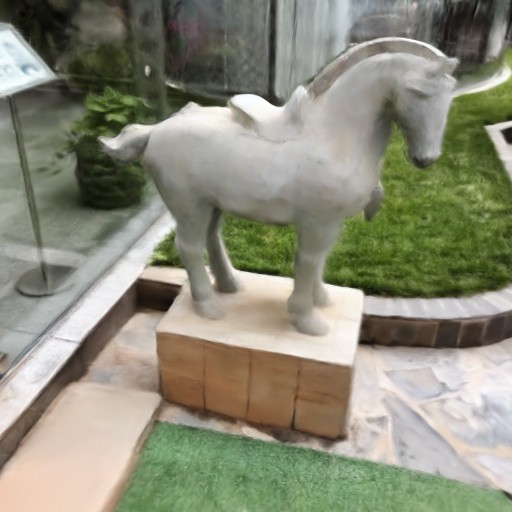}
    & \includegraphics[width=0.151\textwidth]{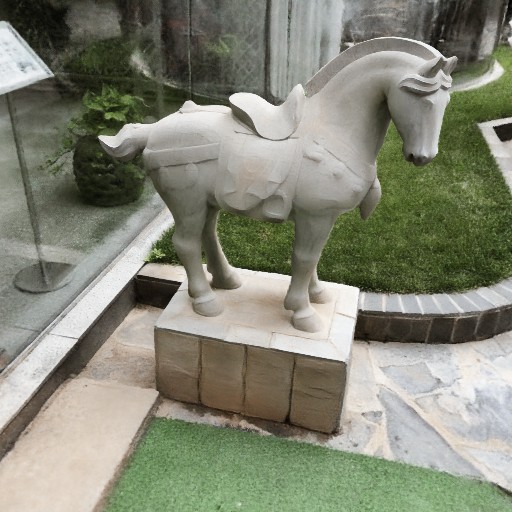}
    & \includegraphics[width=0.151\textwidth]{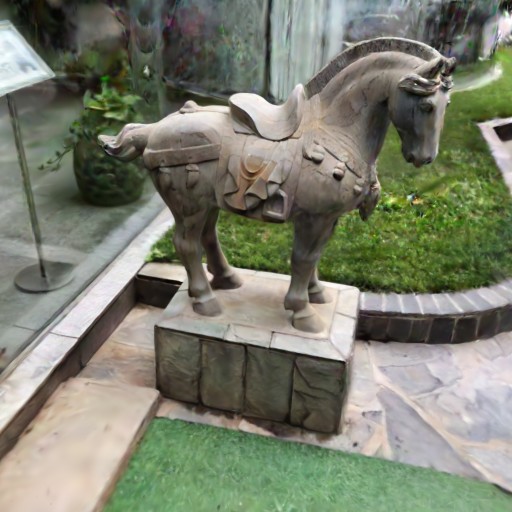}
    & \includegraphics[width=0.151\textwidth]{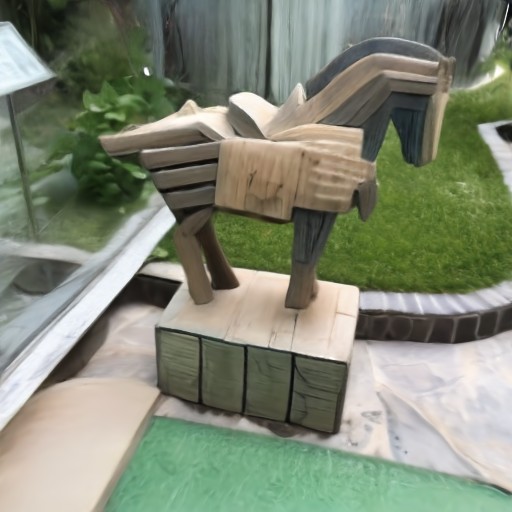}
    \\

\end{tabular}

    \vspace{-5pt}
    \caption{\textbf{Qualitative Comparison from Novel Views} Our method's ability to change the scene's shape allow its stylizations to be more aesthetically pleasing and exhibit more adherence to the style prompt.}
    \vspace{-5pt}
    \label{fig:qualitative-comparison}    
\end{figure*}

\noindent\textbf{Metrics} 
We compute CLIP metrics defined in ~\cite{gal2022stylegan, brooks2023instructpix2pix, haque2023instruct}. Specifically, we first compute the vector direction between the negative prompt and the stylization prompt. We also compute the vector direction between the unstylized and stylized images. Both vector directions should agree if the prompt is adhered to; this metric is labeled CLIP Direction Similarity. We also measure the consistency in stylization by computing the change in image vector directions across frames, CLIP Direction Consistency.

\begin{figure}[h!]
    \centering
    \vspace{-5pt}
    \renewcommand{\tabcolsep}{2pt}
    \small
    \includegraphics[width=.9\linewidth]{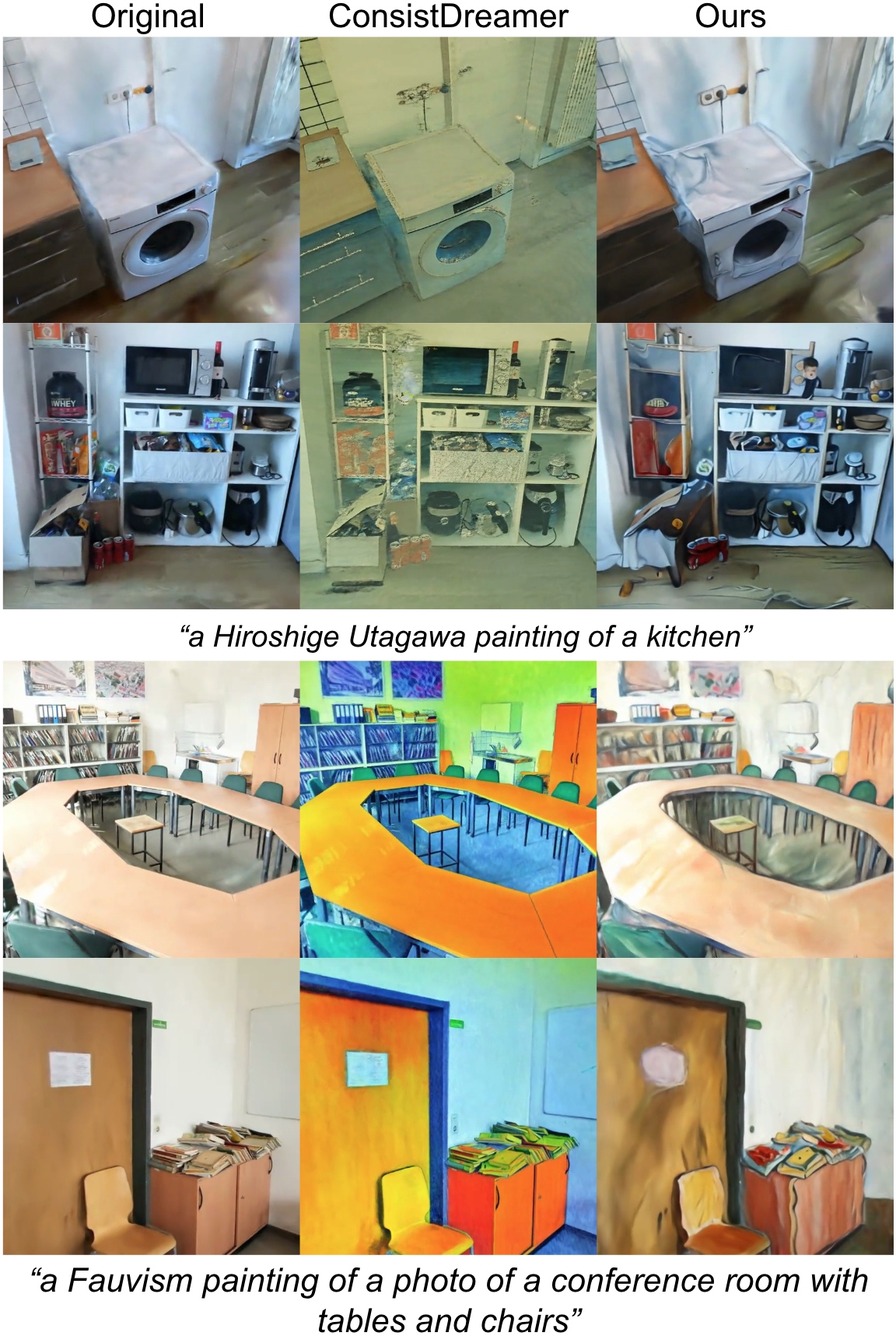}
    
    \vspace{-5pt}
    \caption{Qualitative comparison of our method with ConsistDreamer~\cite{chen2024consistdreamer} since code is not available. Our method alters the scene beyond local texture and color augmentations.}
    \vspace{-10pt}
    \label{fig:consistdreamer}  
\end{figure}

\noindent\textbf{Qualitative Evaluation} We show extensive qualitative results as CLIP metrics alone are not very reliable at judging stylization quality~\cite{wu2024gaussctrl, haque2023instruct}. We experiment with varying the strength of our geometry stylization on both the intermediate RGBD diffusion model and the final 3DGS in Fig.~\ref{fig:splat_stylization_strength}. Notably, in the final 3DGS stylizations, our method is capable of leaving texture information consistent while changing geometry with varying $T_\text{max}^D$. Our method stylizes the scene beyond simple local texture and color edits compared to ConsistDreamer~\cite{chen2024consistdreamer} in Fig.~\ref{fig:consistdreamer}. We show a side-by-side comparison of our model and baselines in Fig.~\ref{fig:qualitative-comparison}; our model consistently outperforms previous methods in both shape editing and overall quality.

\noindent\textbf{User Study}
We report an A/B user study with 31 participants in Figure~\ref{fig:a_b_study}. We use 49 prompt/scene combinations and render A/B comparison videos from a circular wiggle at randomly selected evaluation views of the unstylized splat, our method, and one of each of the baselines from Table~\ref{tab:main_scores}. We randomly select 8 A/B videos for each baseline for a total of 32 videos. We ask users the following questions for each video: \textit{Of the two videos, which one most closely follows the} \textbf{style} \textit{description?} and \textit{Of the two videos, which one is most} \textbf{aesthetically} \textit{pleasing?}. We note that the videos are not cherry picked and are rendered from difficult evaluation views using challenging stylization prompts.

\noindent\textbf{Ablations} 
We ablate our core contributions one by one in Table~\ref{tab:ablations} and show qualitative comparisons in Figure~\ref{fig:ablations_figure}. Inconsistent pipeline output would lead to blurry 3DGS models whose blurry renders exhibit deceptively low RMSE and high consistency. Therefore, we define the metrics \textit{Sequential RMSE} and \textit{Sequential LPIPS} as the RMSE and LPIPS respectively computed between successive pairs of stylized frames from our pipeline (prior to 3DGS retraining), where we warp one to the other. This allows us to quantify how good 3D consistency is between successive views from our pipeline. In Table~\ref{tab:ablations}, (1) is naive stylization with nothing to enforce consistency between frames; (2) uses an RGB inpainting model with a ControlNet conditioned on depth; and (3) adds depth prediction on stylized frames for warping and compositing. (6) Without our \textcolor{warp_controlnet_model}{Warp ControlNet}, stylization is incorrectly propagated and artifacts are left behind, resulting in worse sequential RMSE and LPIPS. Not sharing features (4) or performing cross-attention equally across the whole of the reference images (5) \textcolor{feature_sharing}{without depth guidance} have a similar effect on metrics.

%The table shows that ablating any of our main contributions worsens performance on this metric.

%We note that the `independent stylization' baseline -- on which subsequent frames are wildly different -- scores remarkably well on RMSE. This appears to be because splatting views that are highly inconsistent leads to blurry splats, and RMSE errors tend to be lower when computed between blurry images that lack high-frequency detail. Seq. RMSE bypasses this shortcoming by directly measuring the consistency between generated frames. We define \textit{Sequential RMSE} as the RMSE computed between successive pairs of stylized frames from our pipeline (prior to any resplatting). This directly measures the extend to which 3D-consistency is maintained from one frame to the next. The independent stylization basleine performs very poorly on Seq. RMSE, as expected.
%This directly measures the extend to which 3D-consistency is maintained from one frame to the next, and is more robust to effects where resplatting of images with poor consistency can lead to blurry splats, which then score well on RMSE due to the lack of high-frequency detail in the rendered images. The table shows that ablating any of our main contributions worsens performance on this metric.

%We disable the use of our depth guidance in our feature sharing and note the reduction in stylization quality and consistency.

\begin{table}[h]
    \renewcommand{\tabcolsep}{2pt}
    \centering
    \small
    % uncomment this line if the table becomes too wide to fit
    \resizebox{1.0\linewidth}{!}
    {
\begin{tabular}{|l|cc|cc|}
\hline
    & Similarity ↑  & Consistency ↑ & Seq. RMSE ↓ & Seq. LPIPS ↓ \\
\hline
(1) Single-Frame Independent Stylization with Sec~\ref{sec:rgbd_model} & 0.161 & 0.592 & .1170 & .0941\\
(2) Warp + RGB Inpaint + Depth ControlNet & 0.110 & 0.581 & \textbf{.0585} & .0959 \\
(3) Warp + RGB Inpaint + Depth ControlNet + DAv2~\cite{depthanything} & 0.104 & \textbf{0.629} &  .0776 & .0975 \\
(4) w$\backslash$o feature sharing (\ref{sec:depth_informed_sharing}) & \textit{0.178} & \textit{0.611} & .0817 & .0931\\
(5) w full x-attn, no feat. injection (\ref{sec:depth_informed_sharing}) & \textbf{0.180} & 0.610 & .0730 & \textit{.0914}\\
(6) w$\backslash$o \textcolor{warp_controlnet_model}{Warp ControlNet} (\ref{sec:warpingcontrolnet}) with inpainting & 0.170 & 0.604 & .0834 & .0917 \\
% w$\backslash$o RGBD (\ref{sec:rgbd_model}) w RGB + Depth ControlNet & 0.110 & 0.581 &  - & - \\
(7) Ours & 0.175 & 0.606 & \textit{.0702} & \textbf{.0911}\\
\hline
\end{tabular}

    }
    \vspace{-5pt}
    \caption{\textbf{Quantitative Ablations} We ablate our core contributions and report their effect on scores.}
    \vspace{-10pt}
    \label{tab:ablations}
\end{table}

\section{Limitations}
\begin{figure}[h!]
    \centering
    \vspace{0pt}
    \renewcommand{\tabcolsep}{2pt}
    \small
    \includegraphics[width=.9\linewidth]{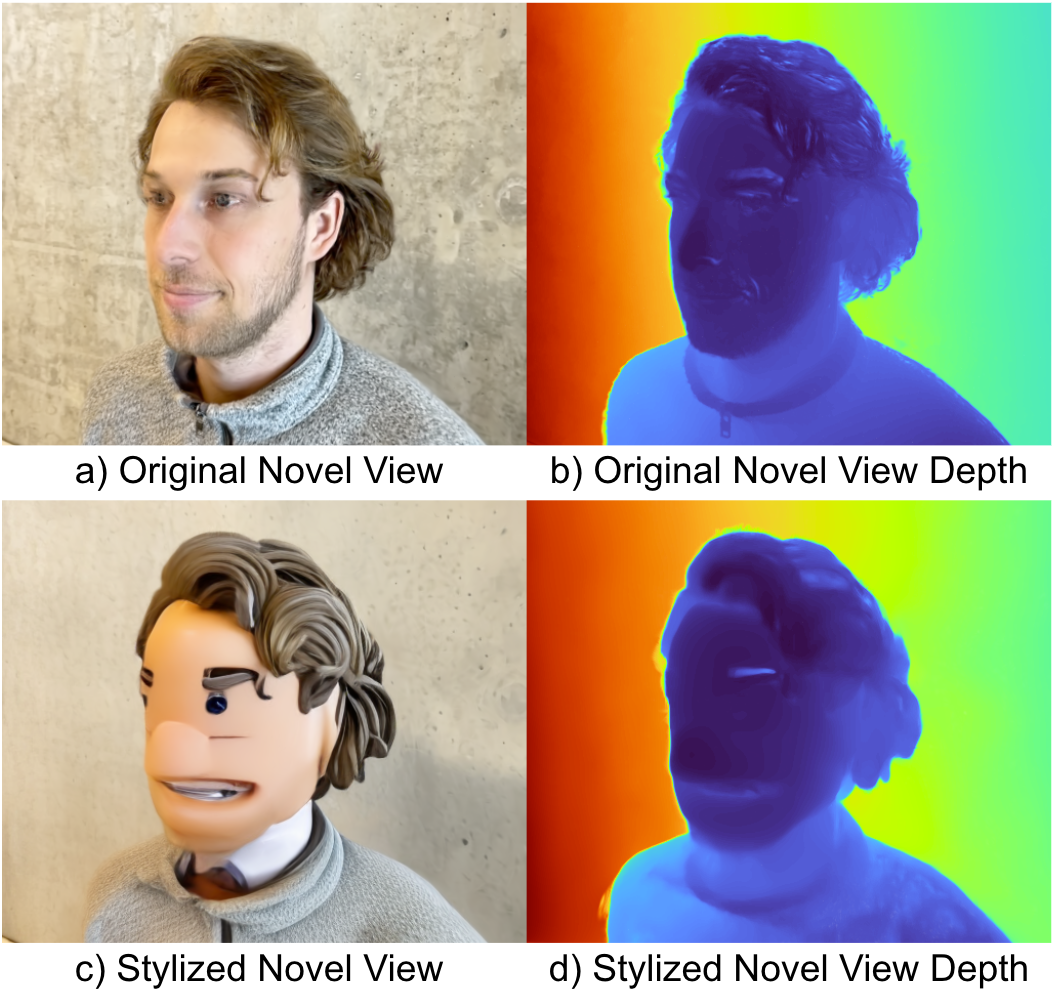}
    
    \vspace{-5pt}
    \caption{\textbf{Limitations} Given errors in the original 3D Gaussian Splat model (see the fuzziness on the right shoulder in a) and the incorrect depth on the eyebrow in b), our method will sometimes inherit these errors in the stylization.}
    \vspace{-10pt}
    \label{fig:limitations}    
\end{figure}

Our method's stylization quality is dependent on trajectory selection. Since our method is reliant on images and depths rendered from the original novel-view synthesis model, we occasionally inherit errors and insert them into stylized output, see Figure~\ref{fig:limitations}.

\section{Conclusion}
We have presented a new method for 3D Gaussian Splat stylization. Our new method utilizes a novel RGBD model for stylization strength control over shape and appearance, a Warp ControlNet for consistently propagating stylizations, and depth-guided feature injection and cross attention. We validated our contributions and the superiority of our method on a user study, a quantitative benchmark, and through qualitative results.

\section{Acknowledgements}
We are grateful to the following colleagues for their support and helpful discussions: Sara Vicente, Saki Shinoda, Stanimir Vichev, Michael Firman, and Gabriel Brostow.

% \begin{figure*}
%     \vspace{-5pt}
%     \centering
%     \small
    
%     \include{figures/comparisons_figure/comparisons_transposed_2}
    
%     \vspace{-5pt}
%     \caption{Qualitative Comparison}
%     \vspace{-5pt}
%     \label{fig:qualitative-comparison-2}    
% \end{figure*}

{
    \small
    \bibliographystyle{ieeenat_fullname}
    \bibliography{main}
}

\clearpage

\noindent \textbf{Supplemental Material}
\maketitle

\appendix
\setcounter{table}{0}
\renewcommand{\thetable}{A\arabic{table}}
\setcounter{figure}{0}
\renewcommand{\thefigure}{A\arabic{figure}}

{
  \hypersetup{linkcolor=blue}
  \tableofcontents
}

%\noindent We will release code, weights, and results upon acceptance.

\section{Note on Results}
All results in the main paper, this supplemental, and the supplemental video are renders from our output stylized 3D Gaussian Splatting models unless stated otherwise.

We provide a supplemental video with extensive qualitative results and an overview of the paper.

\section{Preliminaries}
We use the phrase `target frame' throughout this supplemental to refer to the frame currently being stylized. A `reference frame' is a frame that is being warped to the target frame, or with which we do cross-attention while stylizing the target frame. 

%%%%%%%%%%%%%%%%%%%%%%%%%%%%%%%%%%%%
\section{Splat regularization details}

\subsection{Normals smoothness prior}
\label{subsec:normal-smoothness}

When training our stylized splats, we impose a TVL1 loss to encourage smoothness. To compute this loss, we generate a normal map $N_{i,j} \in \mathbb{R}^{H\times W\times 3}$ from a depth render of the 3D Gaussian Splatting model using cross products on local image gradients~\cite{sayed2022simplerecon}. We define the loss function accumulated across pairs of adjacent pixels:

\begin{equation}
\mathcal{L}_{\text{TVL1}} = \sum_{i,j} \left( \left|\left| N_{i+1,j} - N_{i,j} \right|\right|_1 + \left|\left| N_{i,j+1} - N_{i,j} \right|\right|_1 \right)
\end{equation}

%While training the splat, we add $\mathcal{L}_{\text{TVL1}}$ to the usual photometric loss with a weighting of 0.001.

\subsection{Supervised regularization}
In addition to this smoothness prior, we also use supervised losses on the depths and normals when training splats. This requires ground-truth depth maps $D^\text{gt}$ and normal maps $N^\text{gt}$ for each training view. For the initial splat, we get $D^\text{gt}$ and $N^\text{gt}$ for each frame from Metric3D~\cite{yin2023metric} as discussed in the main paper. For the final stylized splat, we use our stylized depth maps as $D^\text{gt}$ and obtain $N^\text{gt}$ from $D^\text{gt}$ using cross products on local image gradients~\cite{sayed2022simplerecon}.

We compute the scale-invariant depth-loss introduced by~\cite{eigen2014depth} between the rendered depth $D$ and the ground-truth $D^\text{gt}$. We refer to this loss term as $\mathcal{L}_{D}$.

As discussed in \ref{subsec:normal-smoothness}, we also obtain a normals image from the splat. In addition to the smoothness prior introduced in that section, we also supervise those normals using the ground-truth, $N^\text{gt}$, with a dot-product loss (which is equivalent to the cosine distance):
\begin{equation}
    \mathcal{L}_{\text{N}} = - \sum_{i,j} \left( N^\text{gt}_{i,j} \cdot N_{i,j} \right)
\end{equation}

Certain scenes, such as Face (from Instruct-NeRF2NeRF~\cite{haque2023instruct}) and Fangzhou (from GaussCtrl~\cite{wu2024gaussctrl}) still exhibit artifacts such as shimmering and popping due to the people moving during the capture.

\subsection{Hyperparameters}
The total loss when training splats is a sum of the above terms, plus the usual photometric loss $\mathcal{L_\text{phot}}$:
\begin{equation}
    \mathcal{L} = \mathcal{L_\text{phot}} + \lambda_\text{TVL1}\mathcal{L}_\text{TVL1} + \lambda_\text{N}\mathcal{L}_\text{N} + \lambda_\text{D}\mathcal{L}_\text{D}
\end{equation}
While training our stylized splats, we wait for 3000 training steps until enabling our regularization terms, after which we use the following parameters for all scenes: $\lambda_\text{TVL1}=0.05$, $\lambda_\text{N}=0.001$, and $\lambda_\text{D}=0.5$. When training the initial splats which we use as inputs for both our model and our baselines, we vary these parameters for different scenes to obtain higher-quality splats.

%During each step of splat training, we first render a depth map $D$ and normalize it, in order to make the depth supervision scale-invariant:

%\begin{equation}
%    D^\text{normed} = \frac{\text{median}(D^\text{gt})}{\text{median}(D)} D
%\end{equation}

%Then we define the following loss to encourage consistency with the ground-truth depth (TODO cite https://arxiv.org/pdf/1406.2283)

%We then compute the log-difference between predicted and ground-truth depths:
%\begin{equation}
%    \mathcal{\Delta} = \log(D^\text{normed} + \epsilon) + \log(D^\text{gt} + %\epsilon),
%\end{equation}

%\begin{equation}
%    \mathcal{L}_{\text{D}} = \log
%\end{equation}

\begin{figure}[t]
    \centering
    % Centered and evenly spaced text
    \raisebox{.25em}{ % Adjust .25em to control vertical position
        \makebox[\columnwidth]{%
            \textbf{\hspace{-22pt} CCEdit \cite{feng2024cceditcreativecontrollablevideo}} \hspace{80pt} \textbf{Ours} % Adjust spacing as needed
        }
    }\\
    \includegraphics[width=\columnwidth]{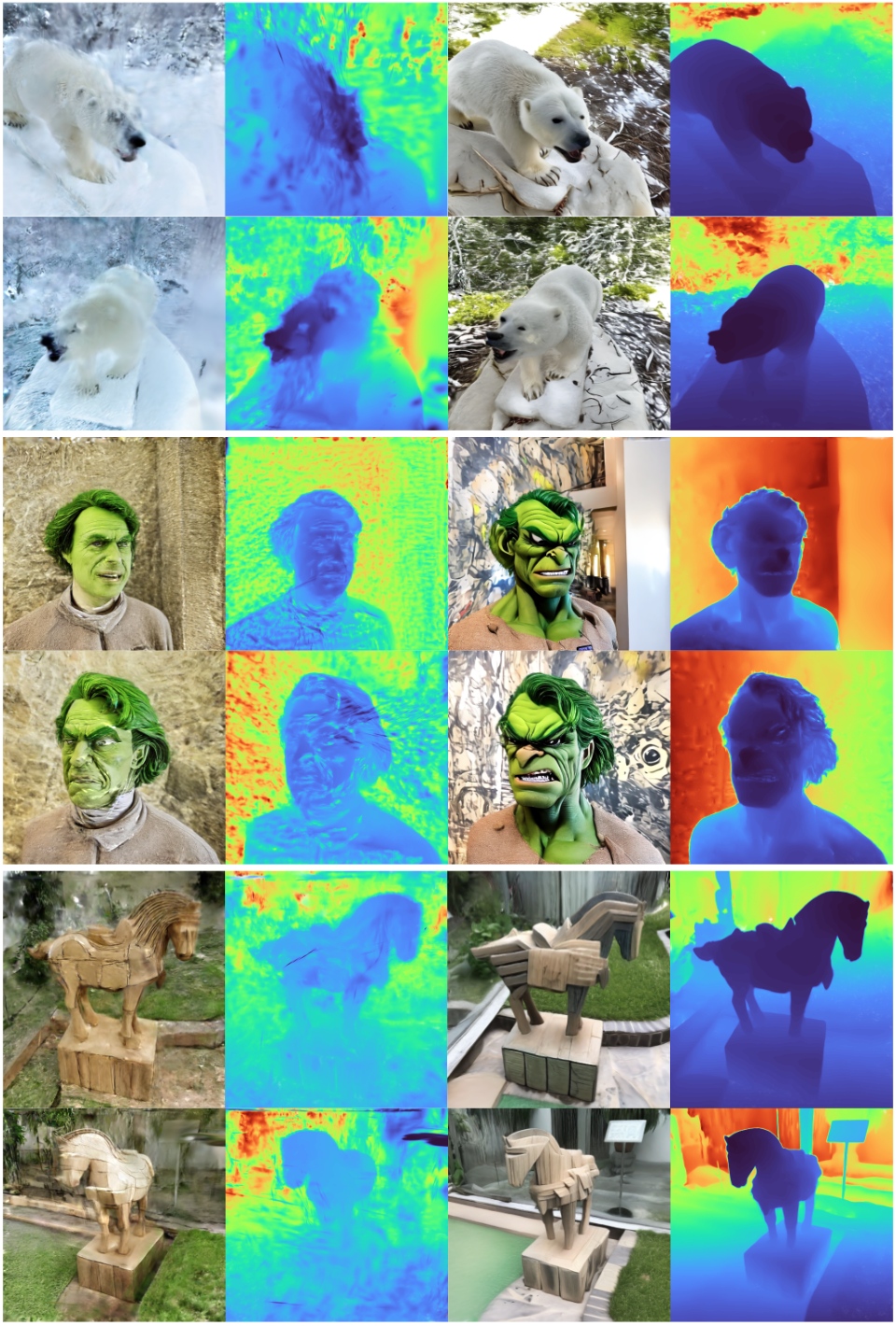}
    \vspace{-15pt}
    \caption{Here, we present output samples from our test set generated using both CCEdit \cite{feng2024cceditcreativecontrollablevideo} and our method. The results demonstrate that our approach achieves significantly better stylization.}
    \label{fig:CCEdit_vs_Ours_fig}
    \vspace{-10pt}
\end{figure}

\subsection{Ablation}
We investigate the impact of the three regularization terms introduced above by switching them all off and recomputing our metrics. We show our scores table in ref.~\ref{tab:tvl1_ablation_scores}. On the CLIP similarity and consistency metrics, our method remains the best, even without these regularisation terms. The increased cloudiness of the geometry unsurprisingly does cause a worsening in the RMSE and LPIPS scores.

\begin{table*}[t]
    \renewcommand{\tabcolsep}{10pt}
    \centering
    \small
    % uncomment this line if the table becomes too wide to fit
    % \resizebox{1.0\textwidth}{!} 
    % {
    %\begin{tabular}{|l|cc|cc|c|}
    %    \hline
    %        & \multicolumn{2}{c}{CLIP} & \multicolumn{2}{|c|}{Image Consistency} & \multicolumn{1}{c|}{Avg Editing Time} \\
    %    % \cmidrule(l){2-3} \cmidrule(l){4-6}  \cmidrule(l){7-8} 
    %    \hline
    %        & CLIP Direction Similarity ↑  & CLIP Direction Consistency ↑  & LPIPS ↓	& RMSE ↓	& \\
    %    \hline
    %% VicaNeRF~\cite{dong2024vica} & & & & & \\
    %Instruct NeRF2NeRF~\cite{haque2023instruct} & 0.113 & 0.594 & & 0.0471 & $\sim$hours\\
    %Instruct GS2GS~\cite{vachha2024instruct} & 0.112 & 0.569 & & \textit{0.0321} & $\sim$10s of minutes\\
    %GaussCtrl~\cite{wu2024gaussctrl} & \textit{0.139} & \textit{0.629} & & 0.0355 & $\sim$minutes\\
    %DGE~\cite{chen2024dge} & 0.138 & 0.606 & & 0.0342 & $\sim$minutes\\
    %\textbf{Ours} & \textbf{0.177} & \textbf{0.633} & & \textbf{0.0301} &  $\sim$minutes\\
    %    \hline
    %\end{tabular}
    \begin{tabular}{|l|cc|cc|}
    \hline
        & \multicolumn{2}{c|}{CLIP} & \multicolumn{2}{c|}{Image Consistency} \\
    \hline
        & CLIP Direction Similarity ↑  & CLIP Direction Consistency ↑ & RMSE ↓ & LPIPS ↓ \\
    \hline
    Instruct NeRF2NeRF~\cite{haque2023instruct} & .098 & .531 & .0463 & .0540 \\
    Instruct GS2GS~\cite{vachha2024instruct} & .097 & .519 & .0501 & \textit{.0403} \\
    GaussCtrl~\cite{wu2024gaussctrl} & .123 & .590 & .0471 & .0438 \\
    DGE~\cite{chen2024dge} & .113 & .565 & \textit{.0384} & .0407 \\
    \hline
    CCEdit~\cite{feng2024cceditcreativecontrollablevideo} w/ our pipeline & .135 & \textbf{.629} & .0915 & .0581 \\
    \hline
    \textbf{Ours w/o resplat reg} & \textit{.171} & \textit{.606} & .0450 & .0454 \\
    \textbf{Ours} & \textbf{.175} & \textit{.606} & \textbf{.0370} & \textbf{.0378} \\
    \hline
\end{tabular}
    % }
    \vspace{-5pt}
    \caption{\textbf{Quantitative Evaluation} We compute metrics for 53 stylizations on a range of published and new scenes. CLIP Direction Similarity measures how well the stylization implied by the prompt is respected. CLIP Direction Consistency measures how consistent this stylization is across frames. We also compute image consistency scores from~\cite{feng2025wave}. Ours outperforms other novel-view stylization methods.}
    \vspace{-10pt}
    \label{tab:tvl1_ablation_scores}
\end{table*}

(We do not include our sequential metrics in this table because they are computed on the stylized images that our pipeline generates rather than from the splat, and therefore they are unaffected by this ablation.)

%%%%%%%%%%%%%%%%%%%%%%%%%%%%%%%%%%%%

%%%%%%%%%%%%%%%%%%%%%%%%%%%%%%%%%%%%

\section{Video Diffusion Baseline}
Video diffusion models have shown great promise for consistent frame-to-frame generation and editing. However, they cannot stylize NVS models on their own. For evaluation, we take an off-the-shelf video editing model, CCEdit~\cite{feng2024cceditcreativecontrollablevideo}, and use it to stylize frames within our pipeline. We report scores in Table~\ref{tab:tvl1_ablation_scores} and a qualitative comparison in Fig.~\ref{fig:CCEdit_vs_Ours_fig}.

By default, CCEdit uses a stylized reference frame as guidance when editing the rest of a video sequence. We use StableDiffusion 2.1~\cite{rombach2021highresolution} to generate the stylized reference using their parameters and the prompt our method uses.

CCEdit is limited to generating a sequence of 17 frames at a time. While we could limit 3DGS renders to just 17 frames, CCEdit cannot handle large frame-to-frame changes given large scenes with large camera baselines. To remedy this, we take inspiration from CCEdit's description for handling longer sequences. We start with the exact same rendered frames we use in our pipeline. We pass chunks of 17 overlapping frames through CCEdit. The last stylized frame in each chunk serves as the style guidance frame in the midpoint of the next chunk. We continue this until all frames are exhausted. We pass the stylized frames through the retraining phase of our pipeline.

CCEdit underperforms compared to our approach across all metrics, except for CLIP Direction Consistency. The subjective quality of the CCEdit results is poor, as indicated in fig.~\ref{fig:CCEdit_vs_Ours_fig}.

\section{Evaluation Metrics}
For the sequential RMSE and sequential LPIPS metrics, we use our mesh warper to warp one stylized frame from our pipeline to the next. In the case of RMSE, we warp the RGB image; for LPIPS, we use a pretrained VGG16 network to extract features, normalize them along the channel dimension, and then warp those features. In both cases we measure the inter-frame consistency by computing the RMSE between the warped previous frame and the next frame, excluding any pixels which lie outside the valid region of the warp.

Mesh warping inherently creates artifacts and so these metrics inherit them. This explains why our `Warp + RGB Inpaint + Depth ControlNet' ablation in Table 2 outperforms our full pipeline on the Seq. RMSE metric: it is rewarded by the RMSE metric for not correcting the warping artifacts.
% (We note that the bold and italic formatting in the Consistency and Seq. RMSE columns of Table 2 were erroneously not applied correctly.)
Because warping artifacts tend to look perceptually unlike the original image, this ablation still underperforms our full method quite substantially on the sequential LPIPS metric. This suggests that sequential LPIPS may be a better measure of inter-frame consistency, because it is less likely to inherit issues with warping artifacts from the underlying warping method.

Our method outperforms all of our competitors on the CLIP similarity and consistency metrics (Table 1), and this aligns with our qualitative figures (Figures 7 and \ref{fig:qualitative-comparison-supp}), in which our method generally adheres better to the prompt. However, Table 2 indicates that our `without feature sharing' and `full x-attn, no feature injection' ablations sometimes outperform our full pipeline on the CLIP metrics. We attribute this finding to a trade-off between \textit{per-frame prompt adherence} and \textit{cross-frame consistency}: stylizing a frame conditional on some previously stylized frames is a more constrained problem then simply stylizing the frame on its own. This likely accounts for why these ablations outperform our full method on CLIP similarity but underperform it on sequential RMSE and sequential LPIPS: they are less constrained by previous frames and therefore more free to align with the prompt. However, since we train a splat on the stylized frames, we prefer the `more consistency' side of this trade-off, to avoid blurriness, floaters and view-dependent artifacts in the final splat.

%%%%%%%%%%%%%%%%%%%%%%%%%%%%%%%%%%%%

%%%%%%%%%%%%%%%%%%%%%%%%%%%%%%%%%%%%
\section{Compositing details}

As discussed in the main paper, we form composites for each reference frame by converting the RGBD frame to a mesh, rendering it from the view of a target frame to be stylized, and compositing them together. For each pixel inside the valid region of the warp, we must choose whether to use either the warped reference frame or the unstylized new frame. We do this by computing a compositing score $S_{i,j}$ for each pixel:

\begin{equation}
    S_{i,j} = \lambda_1 \left|\sin(\theta_{i,j})\right| - \lambda_2 \frac{d'_{i,j}}{d_{i,j}} + \lambda_3 \left(\text{idx} - \text{idx}'\right)
\end{equation}

In the first term above, $\theta_{i,j}$ is the angle between the mesh normal at target-frame pixel $i,j$ and the camera look-at vector of the \emph{reference} frame (even though we are compositing in the target frame). This term rewards geometry that is viewed face-on in the reference frame (and is therefore less likely to have warping artifacts).

In the second term, $d'_{i,j}$ is the depth of pixel $i,j$ in the warped stylized reference frame, and $d_{i,j}$ is the depth of that same pixel in the unstylized target frame. This is an occlusion term which tends to favour whichever frame is occluding the other.

In the final term, $\text{idx}$ and $\text{idx'}$ are the indices of the target and reference frame respectively within our camera trajectory. This term tends to favour reference frames that are older, because earlier reference frames are less likely to contain artifacts arising from repeated warping. Note that because of the autoregressive nature of our pipeline, $\text{idx} - \text{idx}'$ is always positive.

The three $\lambda$ parameters are weights for each of these factors. We use the same values everywhere: $\lambda_1=1.0$, $\lambda_2=3.0$, and $\lambda_3=0.02$.

%%%%%%%%%%%%%%%%%%%%%%%%%%%%%%%%%%%%

%%%%%%%%%%%%%%%%%%%%%%%%%%%%%%%%%%%%
\section{Full list of prompts}

We list all scenes and prompts used for the evaluation and user study in Table \ref{tab:evaluation_scenes}, \ref{tab:evaluation_scenes_3} and \ref{tab:evaluation_scenes_2}. For reproducibility purposes, we include information on the positive, negative, and inversion prompts together with guidance scale, color, and depth strengths.

\begin{table*}[t]
    \renewcommand{\tabcolsep}{6pt}
    \renewcommand{\arraystretch}{1.2}
    \centering
    \small
    % uncomment this line if the table becomes too wide to fit
    \resizebox{1.0\textwidth}{!} {
    \begin{tabular}{|p{0.1\textwidth}|p{0.1\textwidth}|p{0.22\textwidth}|p{0.22\textwidth}|p{0.22\textwidth}|p{0.07\textwidth}|p{0.07\textwidth}|p{0.07\textwidth}|p{0.07\textwidth}|}
        \toprule
 &  &  & Negative Prompt & Inversion Prompt & Guidance Scale & Color Strength & Depth Strength & Used in the User Study \\
Dataset & Scene Name & Positive Prompt &  &  &  &  &  &  \\
\midrule
\multirow[t]{3}{*}{BlendedMVS} & \multirow[t]{3}{*}{Dinosaur} & ``Photo of a robot dinosaur" & ``a photo of a dinosaur statue on the road side lowres, grainy, blurry" & ``a photo of a dinosaur statue on the road side" & 7.50 & 0.50 & 0.40 & 55 No \\
 &  & ``Photo of an ostrich" & ``a photo of a dinosaur statue on the road side lowres, grainy, blurry" & ``a photo of a dinosaur statue on the road side" & 5.00 & 0.50 & 0.30 & 51 Yes \\
 &  & ``a photo of a giraffe on a desert" & ``a photo of a dinosaur statue on the road side lowres, grainy, blurry" & ``a photo of a dinosaur statue on the road side" & 5.00 & 0.50 & 0.30 & 51 Yes \\
\cline{1-9} \cline{2-9}
\multirow[t]{13}{*}{GaussCtrl} & \multirow[t]{5}{*}{Fangzhou} & ``a photo of a face of a man with a moustache" & ``a photo of a face of a man lowres, grainy, blurry" & ``a photo of a face of a man" & 7.50 & 0.70 & 0.70 & 55 No \\
 &  & ``a photo of a face of an old man with wrinkles" & ``a photo of a face of a man lowres, grainy, blurry" & ``a photo of a face of a man" & 7.50 & 0.70 & 0.70 & 55 No \\
 &  & ``a photo of a human skeleton" & ``a photo of a face of a man lowres, grainy, blurry" & ``a photo of a face of a man" & 10.00 & 0.80 & 0.80 & 51 Yes \\
 &  & ``a photo of a man wearing a pair of glasses" & ``a photo of a face of a man lowres, grainy, blurry" & ``a photo of a face of a man" & 7.50 & 0.70 & 0.70 & 55 No \\
 &  & ``a photo of a man with a hat on" & ``a photo of a face of a man lowres, grainy, blurry" & ``a photo of a face of a man" & 7.50 & 0.60 & 0.80 & 55 No \\
\cline{2-9}
 & \multirow[t]{8}{*}{Stone Horse} & ``Cubist portrait of a horse" & ``a photo of a stone horse in front of the museum lowres, grainy, blurry" & ``a photo of a stone horse in front of the museum" & 7.50 & 0.60 & 0.40 & 55 No \\
 &  & ``Horse statue made from stacked wooden blocks, rustic and minimalistic" & ``a photo of a stone horse in front of the museum lowres, grainy, blurry" & ``a photo of a stone horse in front of the museum" & 5.00 & 0.60 & 0.40 & 51 Yes \\
 &  & ``Horse statue made from very large lego bricks in front of a museum" & ``a photo of a stone horse in front of the museum lowres, grainy, blurry" & ``a photo of a stone horse in front of the museum" & 10.00 & 0.70 & 0.30 & 55 No \\
 &  & ``Photo of a dinosaur" & ``a photo of a stone horse in front of the museum lowres, grainy, blurry" & ``a photo of a stone horse in front of the museum" & 7.50 & 0.60 & 0.40 & 55 No \\
 &  & ``Steampunk horse statue with brass gears, pipes, and mechanical parts" & ``a photo of a stone horse in front of the museum lowres, grainy, blurry" & ``a photo of a stone horse in front of the museum" & 5.00 & 0.60 & 0.40 & 55 No \\
 &  & ``a photo of a cyberpunk sci robot horse" & ``a photo of a stone horse in front of the museum lowres, grainy, blurry" & ``a photo of a stone horse in front of the museum" & 5.00 & 0.60 & 0.40 & 51 Yes \\
 &  & ``a photo of a giraffe in zoo" & ``a photo of a stone horse in front of the museum lowres, grainy, blurry" & ``a photo of a stone horse in front of the museum" & 7.50 & 0.65 & 0.65 & 51 Yes \\
 &  & ``a photo of a medieval horse in front of the castle" & ``a photo of a stone horse in front of the museum lowres, grainy, blurry" & ``a photo of a stone horse in front of the museum" & 10.00 & 0.60 & 0.40 & 51 Yes \\
\cline{1-9} \cline{2-9}
\multirow[t]{12}{*}{\raisebox{-0.5\height}{\makecell{Instruct\\Nerf2Nerf}}} & \multirow[t]{4}{*}{Bear} & ``a photo of a hippopotamus next to a river" & ``a photo of a bear statue in the forest lowres, grainy, blurry" & ``a photo of a bear statue in the forest" & 7.50 & 0.60 & 0.35 & 55 No \\
 &  & ``a photo of a panda in the bamboo forest" & ``a photo of a bear statue in the forest lowres, grainy, blurry" & ``a photo of a bear statue in the forest" & 7.50 & 0.50 & 0.50 & 51 Yes \\
 &  & ``a photo of a polar bear in the winter forest" & ``a photo of a bear statue in the forest lowres, grainy, blurry" & ``a photo of a bear statue in the forest" & 5.00 & 0.70 & 0.10 & 51 Yes \\
 &  & ``a photo of a rhino" & ``a photo of a bear statue in the forest lowres, grainy, blurry" & ``a photo of a bear statue in the forest" & 4.50 & 0.70 & 0.60 & 51 Yes \\
\cline{2-9}
 & \multirow[t]{8}{*}{Face} & ``Portrait of a 19th-century aristocrat with a powdered wig and elaborate clothing" & ``a photo of a face of a man lowres, grainy, blurry" & ``a photo of a face of a man" & 7.50 & 0.75 & 0.60 & 51 Yes \\
 &  & ``Viking warrior portrait with a braided beard, fur cloak, and a fierce expression" & ``a photo of a face of a man lowres, grainy, blurry" & ``a photo of a face of a man" & 7.50 & 0.75 & 0.75 & 51 Yes \\
 &  & ``a photo of a chimp" & ``a photo of a face of a man lowres, grainy, blurry" & ``a photo of a face of a man" & 5.00 & 0.50 & 0.70 & 51 Yes \\
 &  & ``a photo of a face of an old man with wrinkles" & ``a photo of a face of a man lowres, grainy, blurry" & ``a photo of a face of a man" & 5.00 & 0.50 & 0.60 & 51 Yes \\
 &  & ``a photo of a man wearing a hat" & ``a photo of a face of a man lowres, grainy, blurry" & ``a photo of a face of a man" & 5.00 & 0.50 & 0.70 & 51 Yes \\
 &  & ``a photo of a man wearing a moustache" & ``a photo of a face of a man lowres, grainy, blurry" & ``a photo of a face of a man" & 5.00 & 0.50 & 0.65 & 51 Yes \\
 &  & ``a photo of a man wearing a pair of glasses" & ``a photo of a face of a man lowres, grainy, blurry" & ``a photo of a face of a man" & 5.00 & 0.50 & 0.70 & 55 No \\
 &  & ``a photo of a man wearing steampunk goggles" & ``a photo of a face of a man lowres, grainy, blurry" & ``a photo of a face of a man" & 7.50 & 0.60 & 0.60 & 51 Yes \\
\cline{1-9} \cline{2-9}
    \end{tabular}
    }
    \vspace{-5pt}
    \caption{\textbf{Evaluation Prompts} Full list of prompts used for evaluation and user study.}
    \vspace{-10pt}
    \label{tab:evaluation_scenes}
\end{table*}

\begin{table*}[t]
    \renewcommand{\tabcolsep}{6pt}
    \renewcommand{\arraystretch}{1.2}
    \centering
    \small
    % uncomment this line if the table becomes too wide to fit
    \resizebox{1.0\textwidth}{!} {
    \begin{tabular}{|p{0.1\textwidth}|p{0.1\textwidth}|p{0.22\textwidth}|p{0.22\textwidth}|p{0.22\textwidth}|p{0.07\textwidth}|p{0.07\textwidth}|p{0.07\textwidth}|p{0.07\textwidth}|}
        \toprule
 &  &  & Negative Prompt & Inversion Prompt & Guidance Scale & Color Strength & Depth Strength & Used in the User Study \\
Dataset & Scene Name & Positive Prompt &  &  &  &  &  &  \\
\midrule

\multirow[t]{18}{*}{ScanNet++} & \multirow[t]{5}{*}{A~\cite{chen2024consistdreamer}} & ``Ancient alchemist's lab with potion bottles, parchment, and mystical symbols" & ``a photo of a microbiology laboratory with microscopes at a university lowres, grainy, blurry" & ``a photo of a microbiology laboratory with microscopes at a university" & 7.50 & 0.70 & 0.40 & 51 Yes \\
 &  & ``Medieval herbalist lab with dried plants, wooden tables, and candle lighting" & ``a photo of a microbiology laboratory with microscopes at a university lowres, grainy, blurry" & ``a photo of a microbiology laboratory with microscopes at a university" & 7.50 & 0.70 & 0.40 & 51 Yes \\
 &  & ``Steampunk laboratory with brass instruments, gears, and retro scientific equipment" & ``a photo of a microbiology laboratory with microscopes at a university lowres, grainy, blurry" & ``a photo of a microbiology laboratory with microscopes at a university" & 7.50 & 0.70 & 0.40 & 55 No \\
 &  & ``a Picasso painting of a microbiology laboratory with microscopes at a university" & ``a photo of a microbiology laboratory with microscopes at a university lowres, grainy, blurry" & ``a photo of a microbiology laboratory with microscopes at a university" & 10.00 & 0.70 & 0.50 & 55 No \\
 &  & ``an Edvard Munch painting of a microbiology laboratory with microscopes at a university" & ``a photo of a microbiology laboratory with microscopes at a university lowres, grainy, blurry" & ``a photo of a microbiology laboratory with microscopes at a university" & 7.50 & 0.60 & 0.10 & 51 Yes \\
\cline{2-9}
 & \multirow[t]{10}{*}{036bce3393} & ``1950s diner-style lounge with checkered floors, red vinyl seats, and jukebox" & ``a photo of a robotics laboratory with computers and a sofa at a university lowres, grainy, blurry" & ``a photo of a robotics laboratory with computers and a sofa at a university" & 10.00 & 0.70 & 0.40 & 51 Yes \\
 &  & ``Jungle-themed break room with plants, natural wood furniture, and tropical decor" & ``a photo of a robotics laboratory with computers and a sofa at a university lowres, grainy, blurry" & ``a photo of a robotics laboratory with computers and a sofa at a university" & 10.00 & 0.70 & 0.40 & 51 Yes \\
 &  & ``Medieval tavern with wooden tables, stone walls, and candle lighting" & ``a photo of a robotics laboratory with computers and a sofa at a university lowres, grainy, blurry" & ``a photo of a robotics laboratory with computers and a sofa at a university" & 10.00 & 0.70 & 0.40 & 51 Yes \\
 &  & ``Minimalist Japanese tea room with tatami mats, low wooden tables, and paper lanterns" & ``a photo of a robotics laboratory with computers and a sofa at a university lowres, grainy, blurry" & ``a photo of a robotics laboratory with computers and a sofa at a university" & 10.00 & 0.70 & 0.40 & 55 No \\
 &  & ``Modern art studio with paint splatters, canvases, and abstract sculptures" & ``a photo of a robotics laboratory with computers and a sofa at a university lowres, grainy, blurry" & ``a photo of a robotics laboratory with computers and a sofa at a university" & 10.00 & 0.70 & 0.40 & 55 No \\
 &  & ``Retro arcade room with classic game machines, colorful lights, and vintage posters" & ``a photo of a robotics laboratory with computers and a sofa at a university lowres, grainy, blurry" & ``a photo of a robotics laboratory with computers and a sofa at a university" & 10.00 & 0.70 & 0.40 & 51 Yes \\
 &  & ``Victorian parlor with ornate furniture, dark wallpaper, and chandeliers" & ``a photo of a robotics laboratory with computers and a sofa at a university lowres, grainy, blurry" & ``a photo of a robotics laboratory with computers and a sofa at a university" & 10.00 & 0.70 & 0.40 & 55 No \\
 &  & ``a Picasso painting of a robotics laboratory with computers and a sofa at a university" & ``a photo of a robotics laboratory with computers and a sofa at a university lowres, grainy, blurry" & ``a photo of a robotics laboratory with computers and a sofa at a university" & 10.00 & 0.70 & 0.50 & 51 Yes \\
 &  & ``a Van Gogh painting of a robotics laboratory with computers and a sofa at a university" & ``a photo of a robotics laboratory with computers and a sofa at a university lowres, grainy, blurry" & ``a photo of a robotics laboratory with computers and a sofa at a university" & 7.50 & 0.60 & 0.20 & 55 No \\
 &  & ``a near-futuristic night city Cyberpunk2077 style picture of a robotics laboratory with computers and a sofa at a university" & ``a photo of a robotics laboratory with computers and a sofa at a university lowres, grainy, blurry" & ``a photo of a robotics laboratory with computers and a sofa at a university" & 7.50 & 0.80 & 0.10 & 51 Yes \\
\cline{2-9}
 & 1b75758486 & ``Fauvist painting of a room with a table and chairs" & ``a photo of a conference room with tables and chairs lowres, grainy, blurry" & ``a photo of a conference room with tables and chairs" & 12.50 & 0.70 & 0.10 & 55 No \\
\cline{2-9}
 & 0cf2e9402d & ``a Hiroshige Utagawa painting of a kitchen" & ``a photo of a kitchen lowres, grainy, blurry" & ``a photo of a kitchen" & 7.50 & 0.60 & 0.20 & 51 Yes \\
\cline{2-9}
 & 3db0a1c8f3 & ``a Bastion game style picture of a flat with a bathroom, living room and a kitchen" & ``a photo of a flat with a bathroom, living room and a kitchen lowres, grainy, blurry" & ``a photo of a flat with a bathroom, living room and a kitchen" & 7.50 & 0.60 & 0.20 & 55 No \\
\cline{1-9} \cline{2-9}
    \end{tabular}
    }
    \vspace{-5pt}
    \caption{\textbf{Evaluation Prompts (continued)} Full list of prompts used for evaluation and user study.}
    \vspace{-10pt}
    \label{tab:evaluation_scenes_3}
\end{table*}

\begin{table*}[t]
    \renewcommand{\tabcolsep}{6pt}
    \renewcommand{\arraystretch}{1.2}
    \centering
    \small
    % uncomment this line if the table becomes too wide to fit
    \resizebox{1.0\textwidth}{!} {
    \begin{tabular}{|p{0.1\textwidth}|p{0.1\textwidth}|p{0.22\textwidth}|p{0.22\textwidth}|p{0.22\textwidth}|p{0.07\textwidth}|p{0.07\textwidth}|p{0.07\textwidth}|p{0.07\textwidth}|}
        \toprule
 &  &  & Negative Prompt & Inversion Prompt & Guidance Scale & Color Strength & Depth Strength & Used in the User Study \\
Dataset & Scene Name & Positive Prompt &  &  &  &  &  &  \\
\midrule

\multirow[t]{3}{*}{Mip-NeRF 360} & \multirow[t]{3}{*}{Garden} & ``a photo of a fountain in the desert" & ``a photo of a fake plant on a table in the garden lowres, grainy, blurry" & ``a photo of a fake plant on a table in the garden" & 10.00 & 0.80 & 0.50 & 55 No \\
 &  & ``a photo of a halloween garden with pumpkins" & ``a photo of a fake plant on a table in the garden lowres, grainy, blurry" & ``a photo of a fake plant on a table in the garden" & 10.00 & 0.65 & 0.40 & 51 Yes \\
 &  & ``a photo of a snowman on a table in the garden in the snow" & ``a photo of a fake plant on a table in the garden lowres, grainy, blurry" & ``a photo of a fake plant on a table in the garden" & 7.50 & 0.70 & 0.40 & 55 No \\
\cline{1-9} \cline{2-9}
\multirow[t]{4}{*}{Our Capture} & \multirow[t]{4}{*}{Car} & ``80s retro car on a neon grid landscape with pink and blue glowing lights" & - & ``Photo of a car" & 10.00 & 0.85 & 0.30 & 51 Yes \\
 &  & ``Monster truck car with oversized wheels and a lifted body" & - & ``Photo of a car" & 10.00 & 0.70 & 0.30 & 51 Yes \\
 &  & ``Photo of a tron legacy nightrider neon futuristic night" & - & ``Photo of a car" & 5.00 & 0.95 & 0.20 & 51 Yes \\
 &  & ``Tank with treads and heavy armor in a rugged environment" & - & ``Photo of a car" & 10.00 & 0.70 & 0.30 & 55 No \\
\cline{1-9} \cline{2-9}

    \end{tabular}
    }
    \vspace{-5pt}
    \caption{\textbf{Evaluation Prompts (continued)} Full list of prompts used for evaluation and user study.}
    \vspace{-10pt}
    \label{tab:evaluation_scenes_2}
\end{table*}

%%%%%%%%%%%%%%%%%%%%%%%%%%%%%%%%%%%%
\begin{figure*}
    \vspace{-5pt}
    \centering
    \small

    \setlength\tabcolsep{2pt}
\renewcommand{\arraystretch}{0}

\newcommand{\qualimwidth}{0.13333333333333333\textwidth}
\newcommand{\addArrowToImage}[5]{ % parameters: image file, x1, y1, x2, y2
    % arguments are: {image_path}{arrow start x}{arrow start y}{arrow end x}{arrow end y}
    % coordinates go from bottom left to top right.
    \begin{tikzpicture}
        \node[anchor=south west,inner sep=0] (image) at (0,0) {\includegraphics[width=\qualimwidth]{#1}};
        \begin{scope}[x={(image.south east)},y={(image.north west)}]
            \draw[-latex, ultra thick, blue] (#2,#3) -- (#4,#5);
        \end{scope}
    \end{tikzpicture}
}
\newcommand{\addTwoArrowToImage}[9]{ % parameters: image file, x1, y1, x2, y2
    % arguments are: {image_path}{arrow start x}{arrow start y}{arrow end x}{arrow end y}
    % coordinates go from bottom left to top right.
    \begin{tikzpicture}
        \node[anchor=south west,inner sep=0] (image) at (0,0) {\includegraphics[width=\qualimwidth]{#1}};
        \begin{scope}[x={(image.south east)},y={(image.north west)}]
            \draw[-latex, ultra thick, orange] (#2,#3) -- (#4,#5);
            \draw[-latex, ultra thick, orange] (#6,#7) -- (#8,#9);
        \end{scope}
    \end{tikzpicture}
}
\begin{tabular}{ccccccc}

    \centering

     & Original & \makecell{Initial Stylized \\ Frame} & \makecell{No \\ Cross-attention} & \makecell{Full Image \\ Cross Attention} & Ours \\

    \raisebox{1\height}{\parbox[t]{12mm}{\rotatebox[origin=c]{90}{\makecell{a photo of a \\ cute dog on \\ a stone pedestal}}}}
    & \includegraphics[width=0.13333333333333333\textwidth]{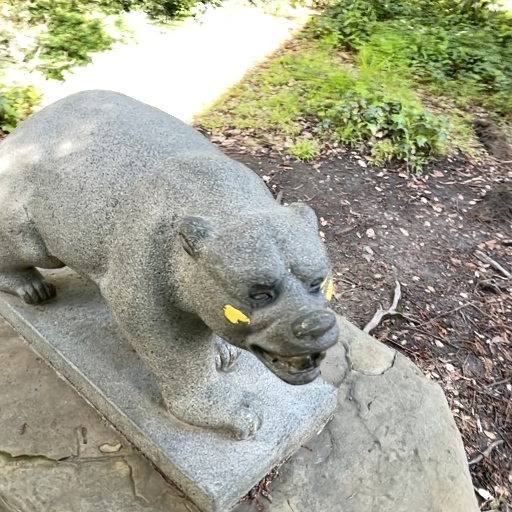}
    & \includegraphics[width=0.13333333333333333\textwidth]{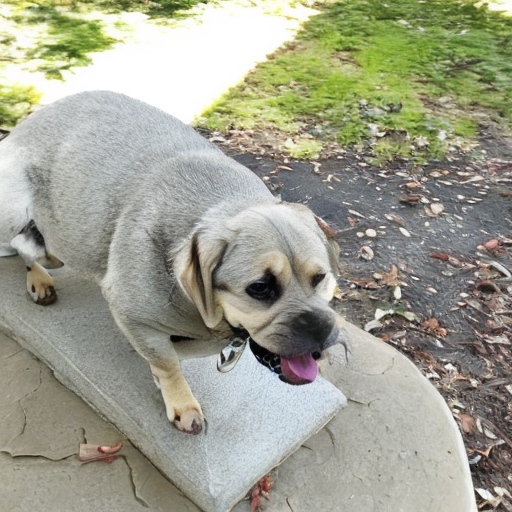}
    & \includegraphics[width=0.13333333333333333\textwidth]{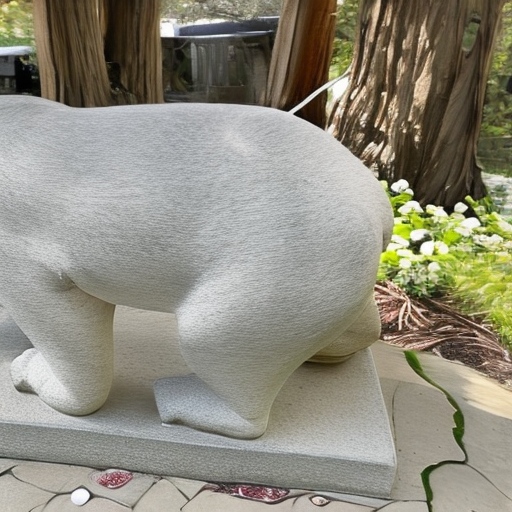}
    & \addTwoArrowToImage{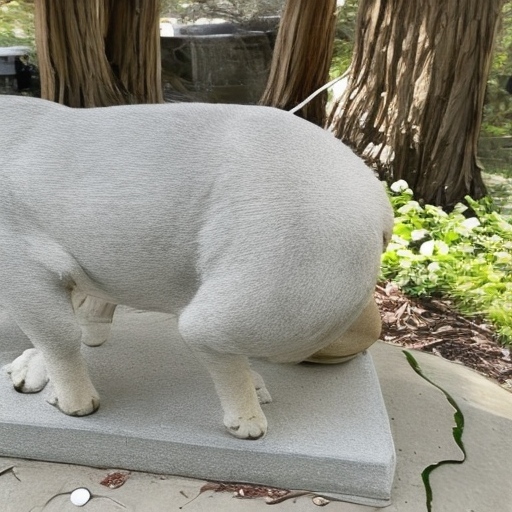}{0.75}{0.3}{0.55}{0.2} {0.42}{0.31}{0.22}{0.21}
    & \includegraphics[width=0.13333333333333333\textwidth]{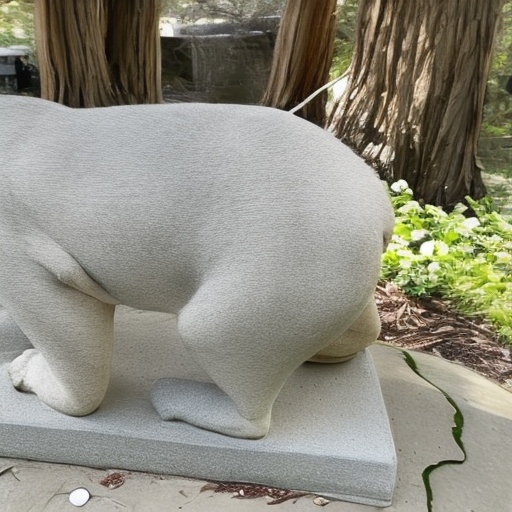}
    \\
  
    \raisebox{0.9\height}{\parbox[t]{12mm}{\rotatebox[origin=c]{90}{\makecell{Steampunk horse \\ with brass gears\\ mechanical parts}}}}
    & \includegraphics[width=0.13333333333333333\textwidth]{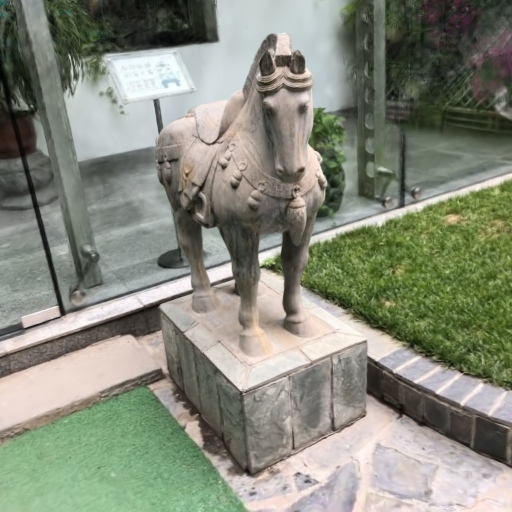}
    & \includegraphics[width=0.13333333333333333\textwidth]{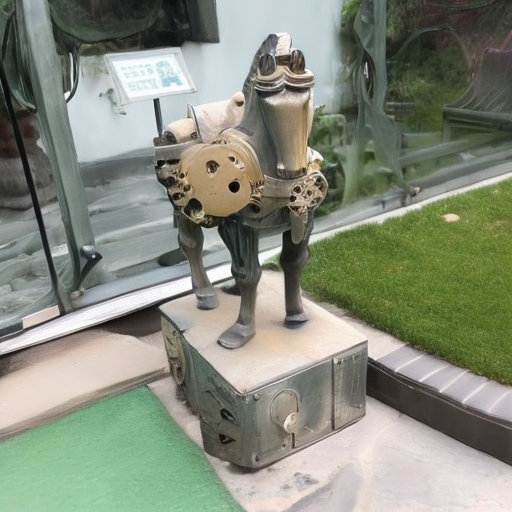}
    & \addArrowToImage{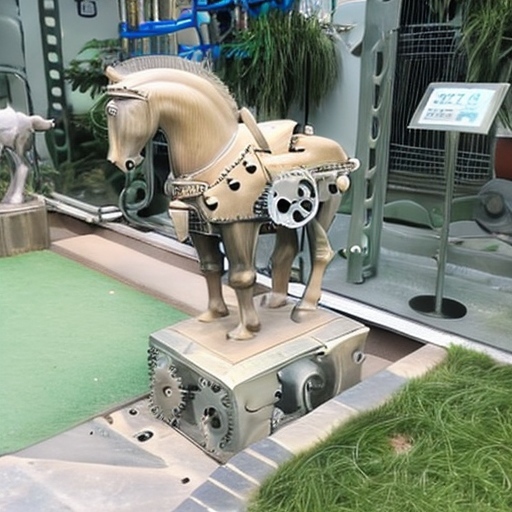}{0.1}{0.35}{0.3}{0.25}
    & \includegraphics[width=0.13333333333333333\textwidth]{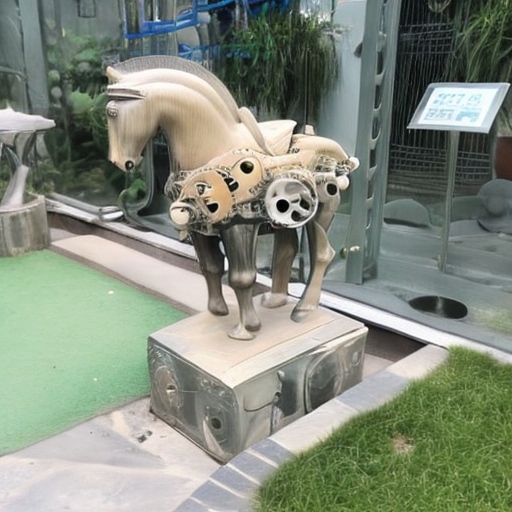}
    & \includegraphics[width=0.13333333333333333\textwidth]{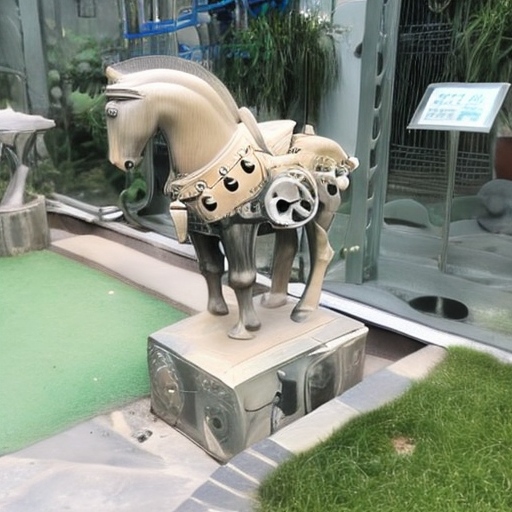}
    \\

\end{tabular}
    
    \vspace{-5pt}
    \caption{\textbf{Qualitative Comparison from Novel Views for Ablations} Here we ablate our feature-sharing contributions. With no cross-attention, the cogs are copied over from the body of the horse onto the pedestal's base (blue arrow). With full-image cross-attention (without our depth-informed heatmaps), the legs of the bear point the wrong way and are misshaped (orange arrows).}
    \vspace{-5pt}
    \label{fig:ablations-qualitative-comparison}    
\end{figure*}

%%%%%%%%%%%%%%%%%%%%%%%%%%%%%%%%%%%%

%%%%%%%%%%%%%%%%%%%%%%%%%%%%%%%%%%%%
\section{Fixed depth noise vs RGB noise}

\begin{figure*}[h!]
    \centering
    \vspace{-5pt}
    \renewcommand{\tabcolsep}{2pt}
    \small
    \includegraphics[width=\linewidth]{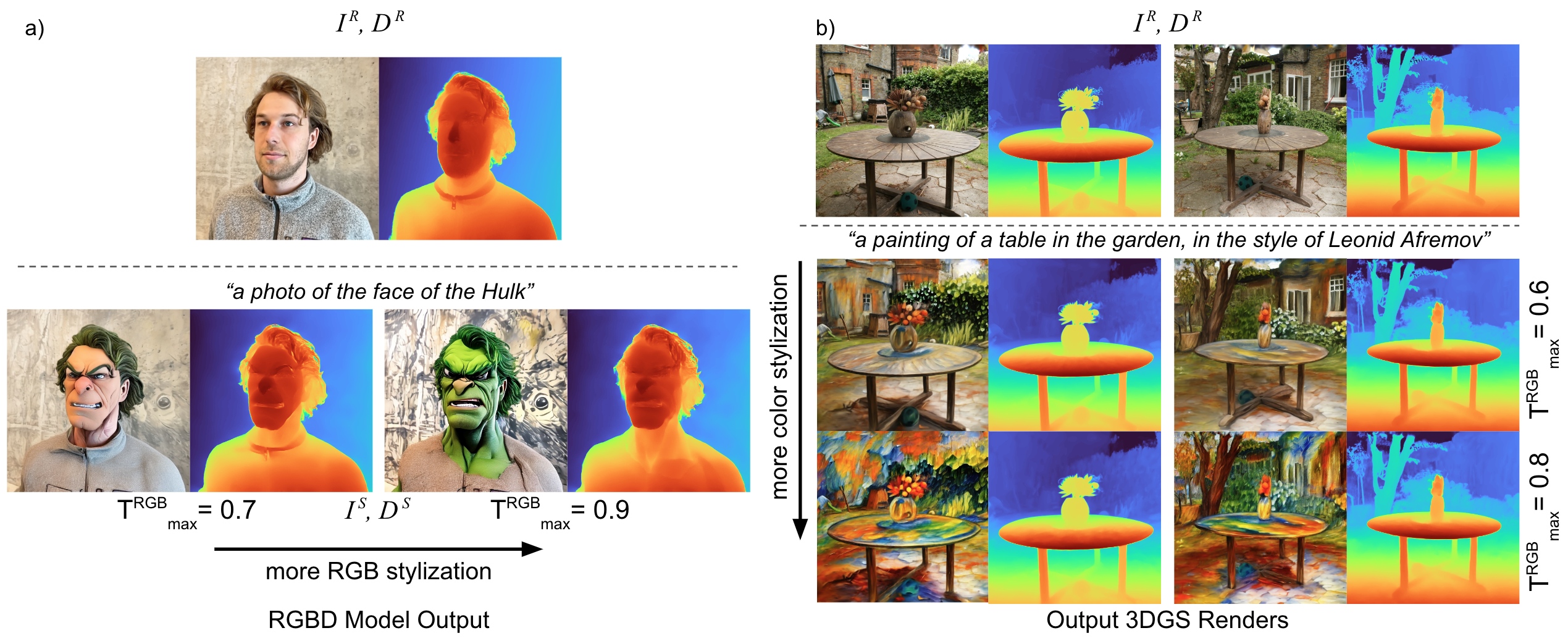}
    
    \vspace{-5pt}
    \caption{For the same prompt, we vary stylization strength for RGB. a) we show the output of our RGBD model for the same stylization prompt but with varying RGB stylization strengths. b) We show the effect of color stylization in output 3DGS models from our method.}
    \label{fig:rgb_stylization_strength}    
\end{figure*}

In the main paper we showed our RGBD model's ability to change depths for varying depth stylization while preserving the the overall color gamut. In Figure \ref{fig:rgb_stylization_strength} we show the opposite and demonstrate our model's ability to change the colors for varying color stylization without considerably changing geometry.

%%%%%%%%%%%%%%%%%%%%%%%%%%%%%%%%%%%%

%%%%%%%%%%%%%%%%%%%%%%%%%%%%%%%%%%%%
\section{\textcolor{feature_sharing}{Cross Attention} implementation details}

\subsection{Construction of the heatmaps}

Our depth-guided attention is based on heatmaps $L_{i,j}$ encoding whether a token $i$ from the reference image forward-warps to the location of token $j$ in the target image. We perform a forward-warp from the reference frame to the target frame by converting the reference RGBD image to a mesh and rendering it into the target frame, as discussed in the main paper. Using this we construct a flow field encoding how a reference-frame pixel $(u,v)$ maps to a target-frame pixel $(u',v')$ under the forward warp:

\begin{equation}
\mathbf{F} \in \mathbb{R}^{H \times W \times 2}, \quad \mathbf{F}_{u,v} = (u', v')
\end{equation}

We then expand this flow field into a 4D tensor:
\begin{equation}
    A_{u,v,u',v'} = \delta \left( \text{round}\left(\mathbf{F}_{u,v}^T\right) - \begin{pmatrix} u' \\ v' \end{pmatrix} \right)
\end{equation}
This tensor $A_{u,v,u',v'}$ will be unity only where the reference-frame coordinates $(u,v)$ map to the target-frame coordinates $(u',v')$ under the forward-warp, and zero elsewhere.  

Finally, we blur this tensor along the spatial axes of the reference. This will have the effect of causing a target-frame feature to attend not just to the reference feature that warps to it, but also to all other features in its immediate vicinity in the reference image. We do this by convolving with a blur kernel:
\begin{equation}
    L_{u, v, u', v'} = A_{u, v, u', v'} * K
\end{equation}
where $*$ is the convolution operator (which we apply only along the $u$ and $v$ axes, i.e. the reference axes). For the blur kernel $K$, we choose a simple linear decay towards zero:
\begin{equation}
    K(\Delta u, \Delta v) = \max\left({1 - \frac{||\Delta u, \Delta v||_2}{d_\text{max}}, 0}\right),
\end{equation}
where $d_\text{max}$ is the number of pixels over which the attention decays to zero. We use $d_\text{max}=100\,\text{px}$ everywhere.
The heatmap $L$ now encodes our desire that target-frame pixels should attend only to reference-frame pixels that map to them. In practice we then clip this so that we always have nonzero attention with all reference-frame features:
\begin{equation}
    L_{u, v, u', v'} \leftarrow \max\left( L_{u, v, u', v'}, L_\text{min} \right)
\end{equation}
This allows the target frame to be stylistically guided by the reference frame even in areas of the target frame that were not visible from the reference. The constant $L_\text{min}$ is a hyperparameter controlling the locality of the cross-attention; when $L_\text{min}$ is unity, we attend equally to the entire reference frame, whereas in the limit $L_\text{min}\rightarrow 0$ we completely disregard reference-frame features that do not warp to a given target-frame feature's location. We choose $L_\text{min}=0.5$ for all forward-facing scenes and $L_\text{min}=0.3$ for all other scenes.

Since the attention layers inside the diffusion U-Net operate at downscaled resolutions, we must also downscale $L_{i,j}$. We do this by reshaping it to a 4D tensor indexed by pixel coordinates $u,v$ in the reference frame and $u',v'$ in the target frame, $L_{u,v,u',v'}$ (as opposed to a 2D tensor indexed by reference- and target-frame tokens). Then we max-pool across the $u$ and $v$ dimensions, and then -- in a separate max-pool -- across the $u'$ and $v'$ dimensions. This allows us to downscale the entire tensor across the spatial axes. We then reshape back to a 2D tensor $L'_{i,j}$ of the correct dimensionality for the downscaled feature maps on which the attention layer acts.

In the following sections we describe how in practice we use the above matrices to control the attention layers inside the diffusion U-Net.

\begin{figure*}[t]
    \vspace{-5pt}
    \centering
    \small
    
    \includegraphics[width=2\columnwidth]{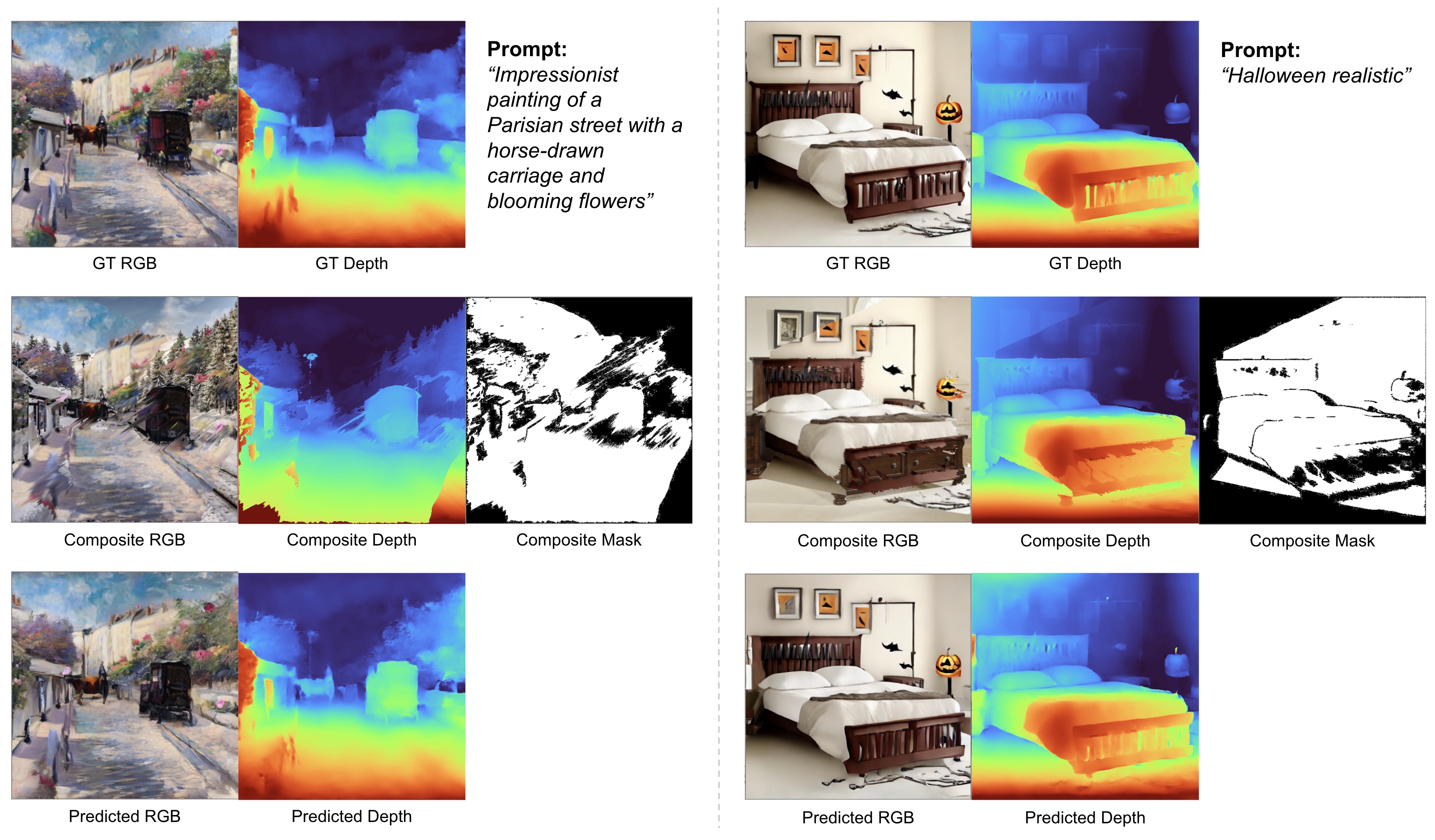}
    
    \vspace{-5pt}
    \caption{\textbf{ControlNet training samples.} \textbf{Top row} depicts the stylized RGBD images as well as the prompts used to stylize the RGBD images; These are the images we are training the ControlNet to predict. \textbf{Middle row} shows the composite RGBD images and composite masks we give the ControlNet as conditioning. Notice the warping artifacts in the composite images. \textbf{Bottom row} shows the predicted RGBD image from the ControlNet for the given conditioning images and noised target images. The model successfully preserves the warped pixels while correcting warping artifacts.}
    \vspace{-5pt}
    \label{fig:controlnet-training-sample}    
\end{figure*}

\subsection{Depth-conditioned cross-attention}
As described in the main paper (eq.\,1), we modify the U-Net's cross-attention layers so that the keys (and values) are the concatenation of the target-frame keys with all of the reference-frame keys (for however many reference frames we have). %The heatmap $L$ is reshaped by flattening the reference-frame spatial dimensions $(u, v)$ into a single dimension $i$, and likewise the target-frame spatial dimensions $u, v$ are flattened into a single dimension $j$. 
We then construct the heatmap $L_{i, j}$ as described in the previous section, and it enters the mask $\Delta$ in eq.\,1 of the main paper as described in the paragraph following that equation. We modify all self-attention layers of the RGBD diffusion model U-Net in this way, though we leave the ControlNet unchanged.

\subsection{Feature injection}
Feature injection is a more direct means of sharing features between reference and target images. Here we simply mix in reference features directly, rather than having the target frame attend to them. This mixing is performed on the final hidden states after each depth-conditioned cross-attention operation.

We first compute a mixing matrix, M:
\begin{equation}
    M_{i,j} = \text{softmax}_i \left(\frac{L_{i,j}}{T}\right)
\end{equation}
This softmax increases the `locality' of the mixing matrix, suppressing elements that were not near unity. It also ensures normalization over the reference-frame axis. This is important because we wish to take weighted averages of reference- and target-frame features, not sums. We choose a very low value for the temperature $T$ of $0.001$, which effectively turns this into an argmax operation. We do this because we only want to inject reference features that warp directly to a given target feature; otherwise the result tends to be somewhat blurry.

We define a weight for injection into a given target hidden state $j$ as follows:
\begin{equation}
    w_j = \max_i L_{i, j} \lambda_\text{inject},
\end{equation}
where the constant $\lambda_\text{inject}$ is a hyperparameter, which we set to 0.15 for all forward-facing scenes and 0.2 for all other scenes. This weight allows us to suppress feature injection in cases where no reference-frame features warp to a given target-frame feature.

Then we use the mixing matrix M to inject reference-frame hidden states into the target-frame hidden states, with the amount of mixing per hidden state weighted by $w_j$:
\begin{equation}
    h_j' = (1 - w_j) h_j + w_j \sum_j h_i^\text{ref} M_{ij}
\end{equation}

%%%%%%%%%%%%%%%%%%%%%%%%%%%%%%%%%%%%

%%%%%%%%%%%%%%%%%%%%%%%%%%%%%%%%%%%%
\section{\textcolor{rgbd_model}{RGBD Diffusion Model} details}

The RGBD diffusion model stylizes RGB and disparity maps. We use disparity maps as opposed to depth since disparity has higher fidelity closer to the camera. 

\subsection{Training the RGBD model for separate controls
over geometry and appearance}
To train our RGBD model and our Warp ControlNet for separate control over geometry and appearance, we introduce time parameters $T^{\text{RGB}}_\text{max}$ and $T^\text{D}_\text{max}$ to independently control the amount of noise applied to the RGB and depth channels; these are defined relative to the maximum training timestep, and so vary from zero (no noise) to one (pure Gaussian noise). We also add two new channels to the UNet input to receive masks that inform the model of whether its predicted step will be discarded or retained for each of the RGB and depth channels. This is similar to the way that inpainting models are trained, except that our masks do not vary across the spatial dimensions of the image. We train the model as follows:

\begin{enumerate}
    \item Randomly select $T^{\text{RGB}}_\text{max}$ and $T^\text{D}_\text{max}$.
    %\item Define $t$ that is the larger of $t_{\text{rgb}}$ and $t_\text{d}$.
    \item Noise the RGB channels up to $T^{\text{RGB}}_\text{max}$, and noise the depth channels up to $T^\text{D}_\text{max}$.
    \item Generate two binary masks: RGB mask, where the value is 0 if $T^\text{RGB}_\text{max} < t$ and 1 otherwise (a value of zero indicating updates to RGB should be ignored at this point), and a depth mask, which is 0 if $T^\text{D}_\text{max} < t$ and 1 otherwise.
    \item Pass these masks into the U-Net as part of the input. This way, the model knows which updates will be discarded, allowing it to condition its processing accordingly.
    \item Run the U-Net as usual and compute the loss based on the denoising tasks it performs. Training is guided by standard supervision losses.
\end{enumerate}
Half of the time we randomly sample values $T^{\text{RGB}}_\text{max}$ and $T^\text{D}_\text{max}$ independently from a uniform distribution between 0 and 1. For the other half of the time, we choose $T^{\text{RGB}}_\text{max}=T^\text{D}_\text{max}$.

\begin{figure*}[t]
    \vspace{-5pt}
    \centering
    \small
    
    \includegraphics[width=0.7\linewidth]{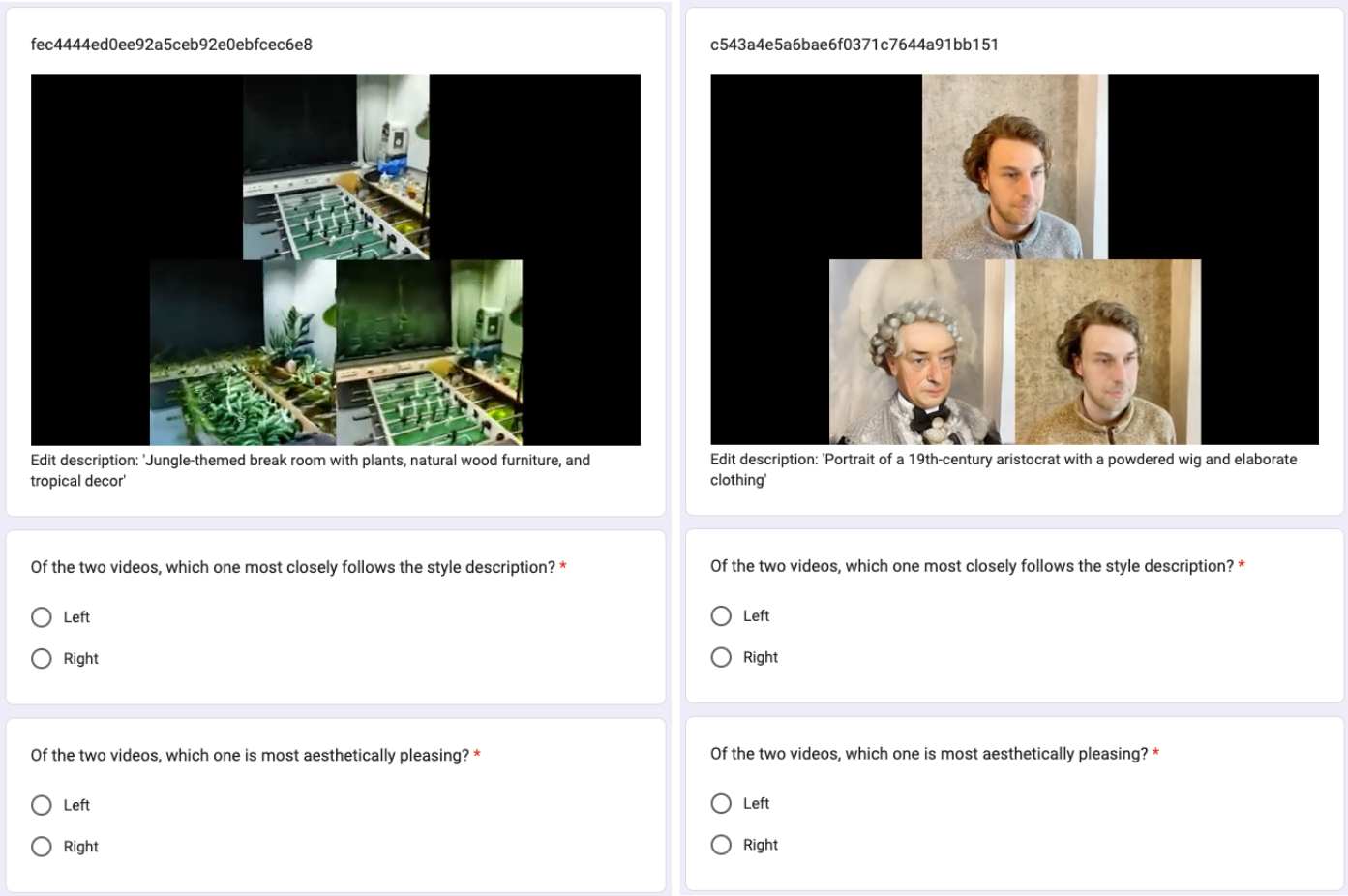}
    
    \vspace{-5pt}
    \caption{\textbf{User Study Prompt} Sample A/B video and associated questions from our user study. The original render is on top, and two methods are on the bottom. The method order is randomized.}
    \vspace{-5pt}
    \label{fig:user-study-sample}    
\end{figure*}

\subsection{Training hyper-parameters}
We train our RGBD model for 30k steps, batch size 4 and accumulation of 8 batches, for an effective batch size of 32. We use an exponential learning rate scheduler which updates at every step. We use a learning rate of 3e-5 with a warmup of 100 steps, decaying to 3e-7 by 25k steps.

\subsection{Inference}
Congruently with our training regime, at inference time we discard the model's RGB updates when $T^{\text{RGB}}_\text{max} < t$, and we discard its depth updates when $T^\text{D}_\text{max} < t$. This allows us to use $T^{\text{RGB}}_\text{max}$ and $T^\text{D}_\text{max}$ to control the extent to which RGB and depth are edited.

%%%%%%%%%%%%%%%%%%%%%%%%%%%%%%%%%%%%

%%%%%%%%%%%%%%%%%%%%%%%%%%%%%%%%%%%%

\section{\textcolor{warp_controlnet_model}{Warp ControlNet} training details}

The Warp ControlNet is conditioned on the composite of the target frame and a warped stylized reference frame alongside the composite mask, and then guides the RGBD diffusion model when stylizing the target frame.

\subsection{RGBD training image composites}
%We generate synthetic data pairs for training our Warp ControlNet. We start with a dataset of generated RGBD pairs of unstylized and stylized frames using our RGBD Diffusion model. We forward-warp the stylized RGBD image to an arbitrary camera view and then warp it back to the original identity camera to form $\text{RGBD}_{\text{warped}}$. We then create composites for conditioning using the validity mask from the second forward-warp. 

As discussed in the main paper, our Warp ControlNet takes a composite image containing a stylized region warped from a previous frame and an unstylized region from the new frame. Its task is: 1) to stylize the unstylized region consistently with the stylized region, and 2) to correct warping artifacts within the stylized region that were introduced by the warp from the previous frame. Here we expand on the procedure by which we train the Warp ControlNet for this task.

To train the ControlNet, we need a dataset of composite images. We generate these composites as follows:
\begin{enumerate}
\item{Sample an image $I$ from the Re-LAION-2B~\cite{schuhmann2022laion} dataset.}
\item{Generate a monocular depth map $D$ for the image with an off-the-shelf detector.}
\item{Stylize the image with a random prompt using our RGBD diffusion model (with no ControlNet), producing a stylized image $I_S$ and depth map $D_S$.}
\item{Forward-warp $I_S$ and $D_S$ to a randomly chosen new camera view using the generated depth map, producing warped stylized RGBD $I_S'$ and $D_S'$.}
\item{Forward-warp $I_S'$ and $D_S'$ back to the original view, producing doubly-warped RGB $I_S''$ and depth $D_S''$.}
\item{Composite together $I_S''$ and $D_S''$ with $I$ and $D$ to produce a new RGB $I_C$ and depth $D_C$.}
\end{enumerate}
Fig.~\ref{fig:controlnet-training-sample} shows example images from this process. The top row of the figure shows stylized RGBDs $I_S$ and $D_S$ generated by step (3) above. The second row shows the final composite, $I_C$ and $D_C$, arising from step (6).

One can see from Fig.~\ref{fig:controlnet-training-sample} that unlike the original $I_S$ and $D_S$, the doubly warped $I_S''$ and $D_S''$ contain both warping artifacts and gaps where stylized information is not available. By training the ControlNet to try and reconstruct the original stylized RGBD $I_S$ and $D_S$ by using $I_S''$ and $D_S''$ as input, the ControlNet learns to fix warping artifacts and to stylize newly visible areas of the scene consistently with the previous frame. This is exactly what we need for our task of stylizing a new view consistently with some previously stylized view of the same scene. Moreover, even though this is a multiview task, this method of generating the training data means we can learn to solve this task using only monocular datasets such as Re-LAION-2B.

%When compositing depth, we need to scale the depths of $\text{RGBD}_{\text{warped}}$to the target depth map. While scaling, we wish to prioritize relevant areas with contextual overlap between the unstylized depth map and the $\text{RGBD}_{\text{warped}}$ depths to reduce the impact of outliers like large flat regions. In order to do this, we only use values near the edges of the invalid regions.

When compositing depth, we need to scale the depths of $D_S''$ to match the original depth map $D$, because our RGBD model only predicts depths up to scale. We compute the scale factor using only pixels that lie near the boundaries between the stylized and unstylized regions in the composited depth map, to minimize the extent of any depth discontinuities between stylized and unstylized regions. One can still see seams between $D$ and $D_S''$ in the composites on Fig.~\ref{fig:controlnet-training-sample} (which the `Predicted Depth' images show that the ControlNet learns to fix), but this approach helps to reduce their extent.
%While scaling, we wish to prioritize relevant areas with contextual overlap between the $D$ and $D_S''$ depths to reduce the impact of outliers like large flat regions. In order to do this, we only use values near the edges of the invalid regions.

\subsection{Stylization prompts}
The stylization prompts for the RGBD training pairs were obtained by querying Llama-3.1-405b-instruct ~\cite{dubey2024llama3herdmodels}, we share some example prompts below:
\begin{enumerate}
\item ``\textit{Ethereal, mystical landscape with glowing, luminescent plants}"
\item ``\textit{A futuristic cyberpunk cityscape at dusk}"
\item ``\textit{Victorian-era style illustration of a fantastical steampunk airship soaring above the clouds}"
\end{enumerate}

\subsection{Training hyper-parameters}
We train our Warp ControlNet for 180k steps, batch size 6 and accumulation of 2 batches for effective batch size of 12. We use a constant learning rate of 1e-4.

% \begin{enumerate}
% \item We create a composite mask based on the valid pixels in $\text{RGBD}_{\text{warped}}$.
% \item The composite RGB is created using this mask where valid pixels from $\text{RGBD}_{\text{warped}}$ are used and the invalid pixels are filled with the original unstylized image.
% \item Similarly, the depth composite must fill the invalid pixels in the $\text{RGBD}_{\text{warped}}$ depth map from the original unstylized depth map. We align the $\text{RGBD}_{\text{warped}}$ depth map to the original unstylized depth map by computing a scale and shift based on values near the edges of the invalid regions. This alignment strategy prioritizes relevant areas with contextual overlap between the unstylized depth map and the $\text{RGBD}_{\text{warped}}$ depths, reducing the impact of outliers like large flat regions
% \end{enumerate}

%%%%%%%%%%%%%%%%%%%%%%%%%%%%%%%%%%%%
\section{Details of inversion}

When stylizing a new frame, we begin by inverting that frame to noise using DDIM inversion. This leads to more consistent results because, unlike simply adding Gaussian noise, it does not destroy the information content of the original RGBD image. During inversion we condition on a prompt describing the content of the unstylized scene (which we refer to elsewhere as the `inversion prompt'), although one can use an empty prompt with little impact on the quality of the results.

We use a standard procedure for DDIM inversion, based on the approximation that the steps predicted by the diffusion model will be similar to latents at successive timesteps: $\epsilon({x_\text{t+1}}) \approx \epsilon({x_\text{t}})$. This then leads to a simple DDIM inversion scheme in which one inverts an image to noise by repeatedly obtaining a step from the diffusion model and then stepping in the opposite direction.

In practice, we find that, owing to our use of depth models at training time, our model is somewhat sensitive to the high-frequency characteristics of the RGBD frame which is used as input, leading to visual artifacts in cases where splat artifacts are present in the input. To improve robustness to these high-frequency characteristics, we instead perform a `partial inversion' in which we first add noise to the input RGBD to get to some noise level $T_\text{noise}$, and then use the above-mentioned inversion procedure for the remaining steps. This allows us to destroy the high-frequency components in the first few steps, avoiding the issue with the model's sensitivity to them. We use a relatively conservative amount of random noise, $T_\text{noise}=0.05$, throughout, which ensures that we do not destroy too much of the lower-frequency signal in the depth map.

For the case where $T^\text{RGB}_\text{max} \neq T^\text{D}_\text{max}$, we adapt the inversion as follows: if $T^\text{RGB}_\text{max} > T^\text{D}_\text{max}$, we perform partial inversion as usual up until $T^\text{D}_\text{max}$ is reached, after which we continue inverting but discard the updates to the depth channels until $T^\text{RGB}_\text{max}$ is reached. If $T^\text{RGB}_\text{max} < T^\text{D}_\text{max}$, we do the converse. 

We note that the relatively non-destructive inversion procedure we use often has the desirable result of tending to preserve view-dependent effects, such as reflections, from the original frame.

%%%%%%%%%%%%%%%%%%%%%%%%%%%%%%%%%%%%

%%%%%%%%%%%%%%%%%%%%%%%%%%%%%%%%%%%%

\begin{figure*}[h!]
    \centering
    \vspace{0pt}
    \renewcommand{\tabcolsep}{2pt}
    \small
    \includegraphics[width=0.8\linewidth]{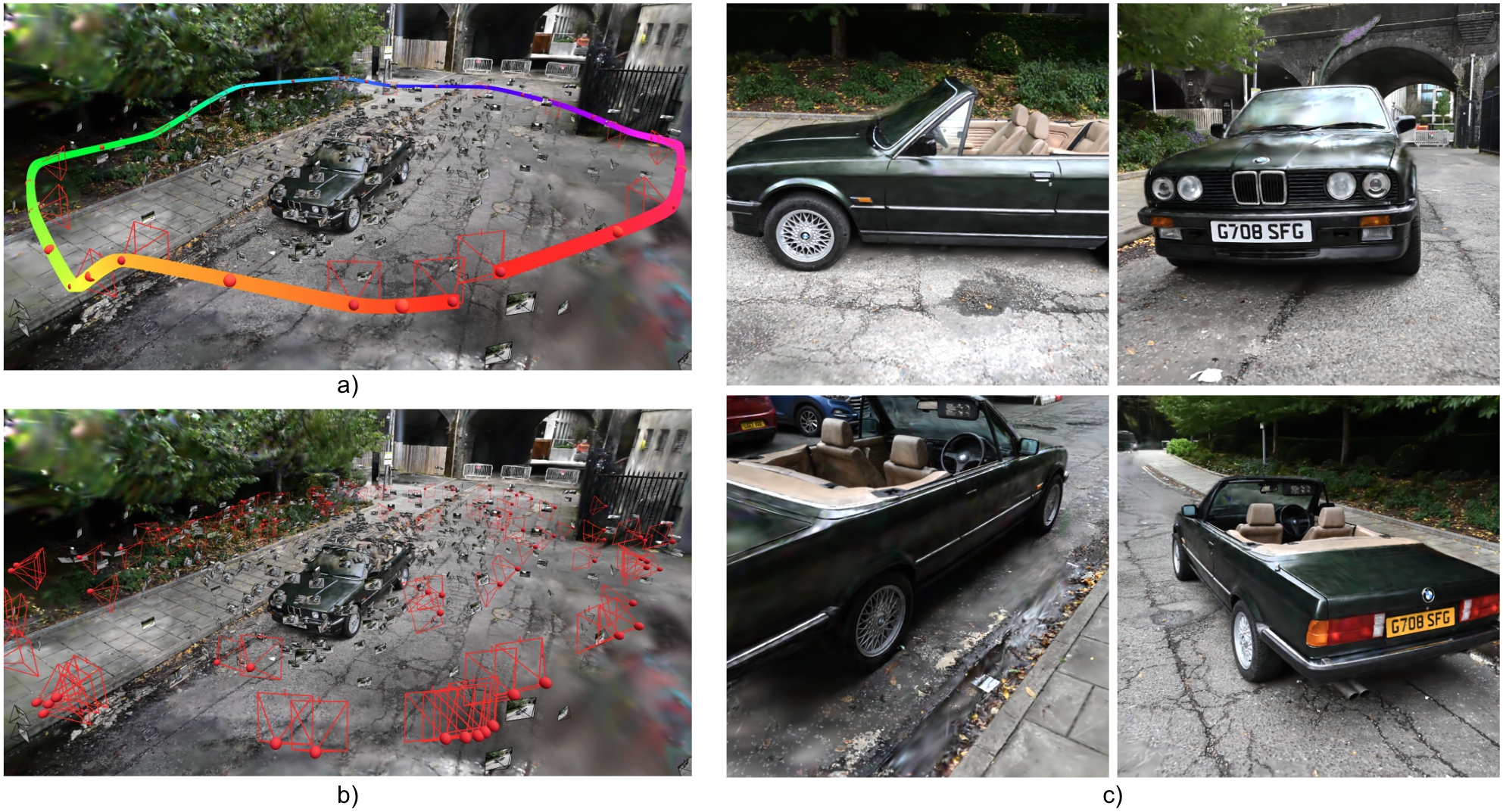}
    
    \vspace{-5pt}
    \caption{\textbf{Training and evaluation camera trajectories} a) Smooth camera trajectory used by our method together with set of camera views used for the input 3DGS model training b) All camera views used for evaluation of our method on this scene c) Example rendered views used for the evaluation}
    \vspace{-10pt}
    \label{fig:train_eval_camera_trajectories}    
\end{figure*}

%%%%%%%%%%%%%%%%%%%%%%%%%%%%%%%%%%%%
\section{Details on the user study}
We sampled views for the user study by selecting a random selection from the prompts in our evaluation set (see Table~\ref{tab:evaluation_scenes}), and then -- for each prompt -- selecting a random evaluation view. In some cases there are evaluation views that are particularly uninteresting, or which give poor coverage of the scene. For example, in the Scannet++ scenes, some of the evaluation views just show a wall at close range. In these cases, we manually eliminate the views from the user study.

For each evaluation view selected for the study, we rendered a circular trajectory in the camera plane centred on that view, which helps to convey a sense of the geometry of the scene from that evaluation view.

The participants in the user study were presented with three splat renders for the circular trajectory: the original splat, our stylized splat, and the stylized splat from one of our baselines. They were asked to indicate preferences between the two stylized splats for prompt adherence and for aesthetic value. A screenshot of the questions for two videos is shown in Figure~\ref{fig:user-study-sample}. Our study had 31 participants, and contained 32 questions.

%%%%%%%%%%%%%%%%%%%%%%%%%%%%%%%%%%%%

%%%%%%%%%%%%%%%%%%%%%%%%%%%%%%%%%%%%
\section{Details of our train view trajectories}

The 3DGS models that we use as an input to our method are trained on all camera views available in the source dataset. However, those views are usually an independent set of photos with corresponding poses that do not follow any smooth camera motion. As our method relies on a continuous trajectory, we create a new smooth sequence of camera views that pass through the original set of training views. We visualize an example of such a trajectory in Fig. \ref{fig:train_eval_camera_trajectories}(a).

%%%%%%%%%%%%%%%%%%%%%%%%%%%%%%%%%%%%

%%%%%%%%%%%%%%%%%%%%%%%%%%%%%%%%%%%%
\section{Further details on evaluation views}
As discussed in the main paper, we select evaluation views that are fair to both our method (which uses smooth camera trajectories) and our baselines (which use the original training views of the splat) by choosing a view from each set and then interpolating halfway between them. 

Half of the time we select a view from our trajectory and then interpolate halfway to the nearest original training view; the other half of the time we do the reverse, selecting one of the original training views and interpolating halfway to the nearest view in our trajectories. 

We linearly interpolate the camera positions and slerp the camera orientations. Sometimes a pair of views will have the cameras looking in very different directions, and then interpolating them may lead to an evaluation view that does not look somewhere sensible. For this reason we reject pairs of views whose camera orientations differ by more than $90^\circ$.

We limit the number of evaluation views to at most 100, in order to avoid excessive evaluation runtimes. We select the evaluation views once per scene and use them everywhere, across all prompts for that scene.

%%%%%%%%%%%%%%%%%%%%%%%%%%%%%%%%%%%%
\section{Further qualitative results}
\begin{figure*}
    \vspace{-5pt}
    \centering
    \small
    
    \setlength\tabcolsep{0pt}
\renewcommand{\arraystretch}{0}
\begin{tabular}{ccccccc}
    \centering

     & Original & GaussCtrl~\cite{wu2024gaussctrl} & Instruct-GS2GS~\cite{vachha2024instruct} & Instruct-N2N~\cite{haque2023instruct} & DGE~\cite{chen2024dge} & Ours \\

    \raisebox{1.1\height}{\parbox[t]{12mm}{\rotatebox[origin=c]{90}{\makecell{a man wearing \\ steampunk \\ goggles}}}}
    & \includegraphics[width=0.151\textwidth]{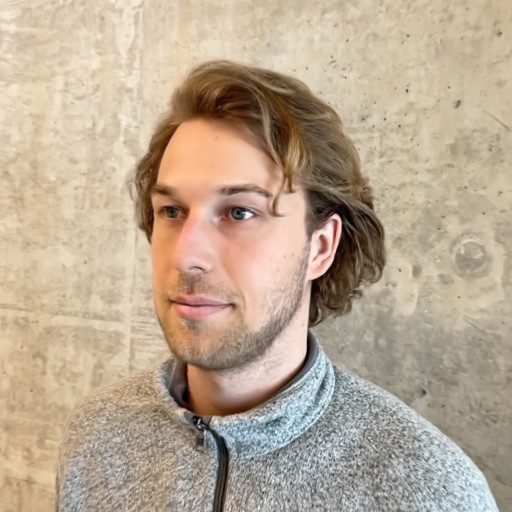}
    & \includegraphics[width=0.151\textwidth]{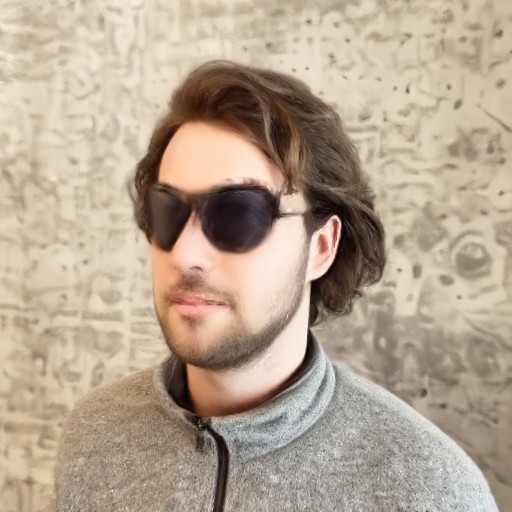}
    & \includegraphics[width=0.151\textwidth]{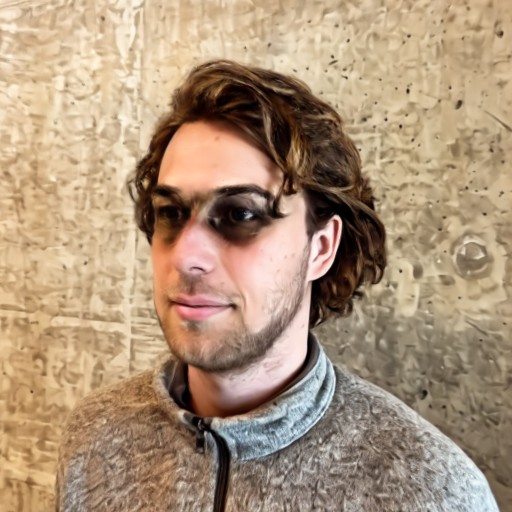}
    & \includegraphics[width=0.151\textwidth]{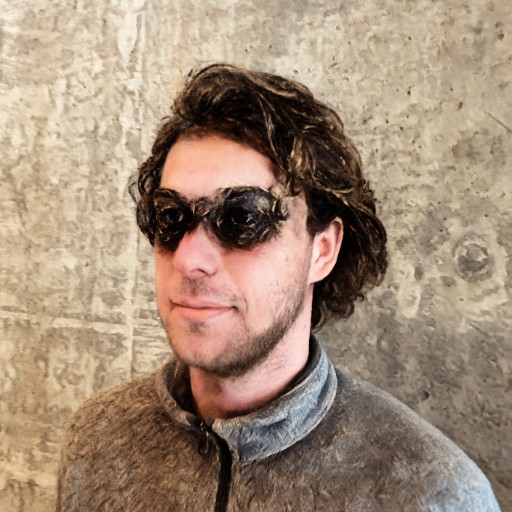}
    & \includegraphics[width=0.151\textwidth]{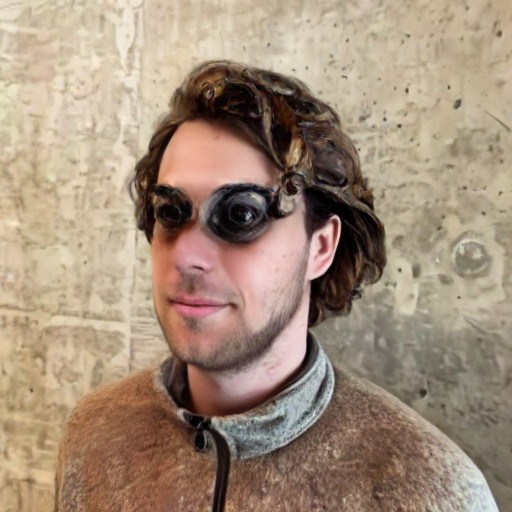}
    & \includegraphics[width=0.151\textwidth]{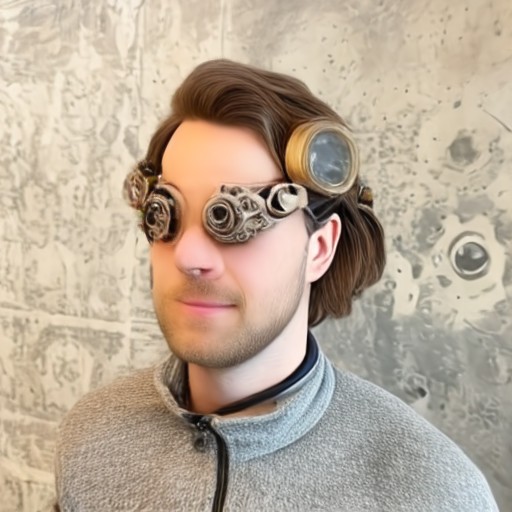}
    \\

    % \raisebox{1.0\height}{\parbox[t]{12mm}{\rotatebox[origin=c]{90}{\makecell{a photo of a \\ face with a hat \\ on}}}}
    % & \includegraphics[width=0.151\textwidth]{figures/comparisons_figure/raw_jpg_small/face_a_photo_of_a_face_with_a_hat_on_original.jpg}
    % & \includegraphics[width=0.151\textwidth]{figures/comparisons_figure/raw_jpg_small/face_a_photo_of_a_face_with_a_hat_on_gaussctrl.jpg}
    % & \includegraphics[width=0.151\textwidth]{figures/comparisons_figure/raw_jpg_small/face_a_photo_of_a_face_with_a_hat_on_instruct_gs2gs.jpg}
    % & \includegraphics[width=0.151\textwidth]{figures/comparisons_figure/raw_jpg_small/face_a_photo_of_a_face_with_a_hat_on_instruct_nerf2nerf.jpg}
    % & \includegraphics[width=0.151\textwidth]{figures/comparisons_figure/raw_jpg_small/face_a_photo_of_a_face_with_a_hat_on_DGE.jpg}
    % & \includegraphics[width=0.151\textwidth]{figures/comparisons_figure/raw_jpg_small/face_a_photo_of_a_face_with_a_hat_on_ours.jpg}
    \\
    \raisebox{1.0\height}{\parbox[t]{12mm}{\rotatebox[origin=c]{90}{\makecell{Portrait of a \\ 19th-century \\ aristocrat}}}}
    & \includegraphics[width=0.151\textwidth]{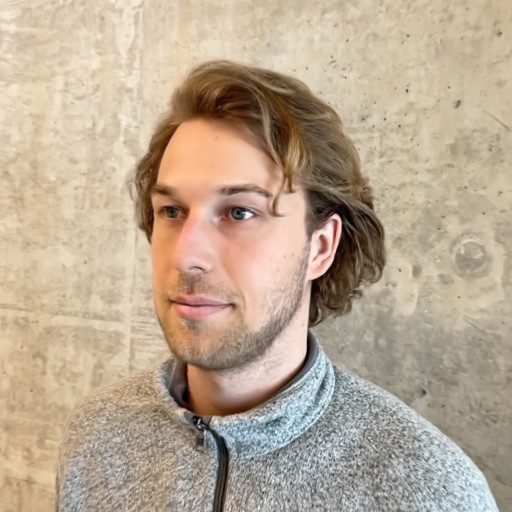}
    & \includegraphics[width=0.151\textwidth]{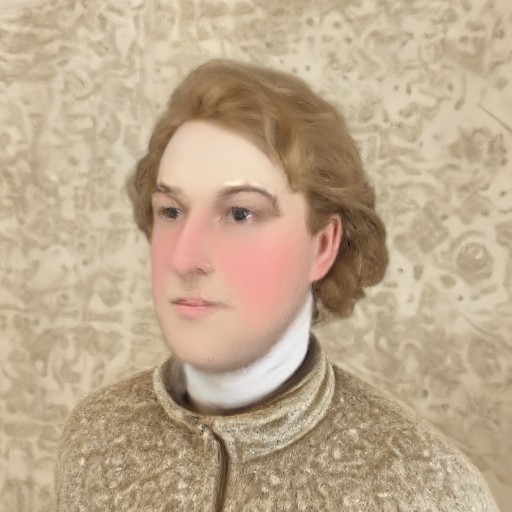}
    & \includegraphics[width=0.151\textwidth]{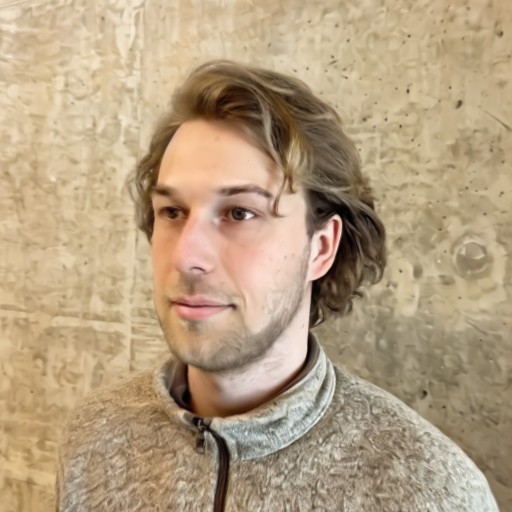}
    & \includegraphics[width=0.151\textwidth]{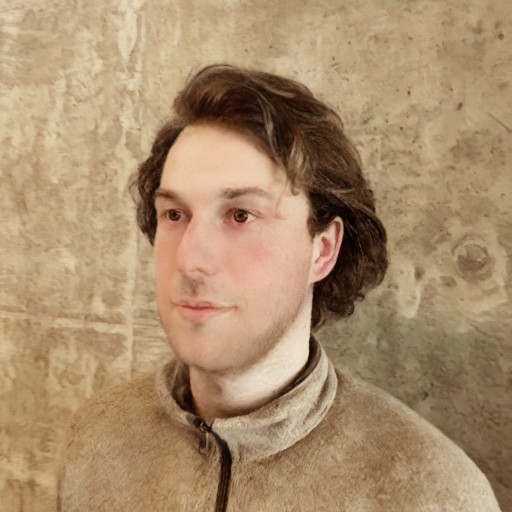}
    & \includegraphics[width=0.151\textwidth]{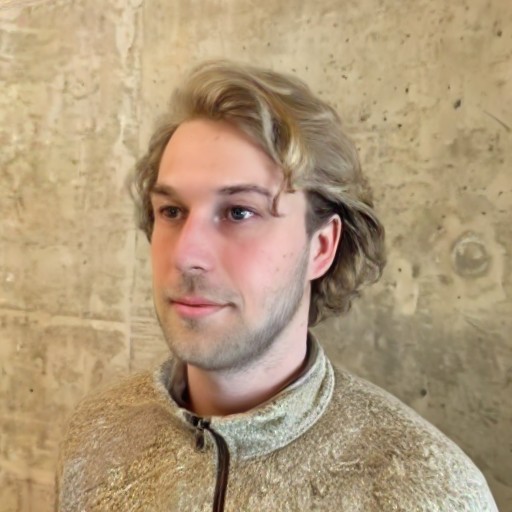}
    & \includegraphics[width=0.151\textwidth]{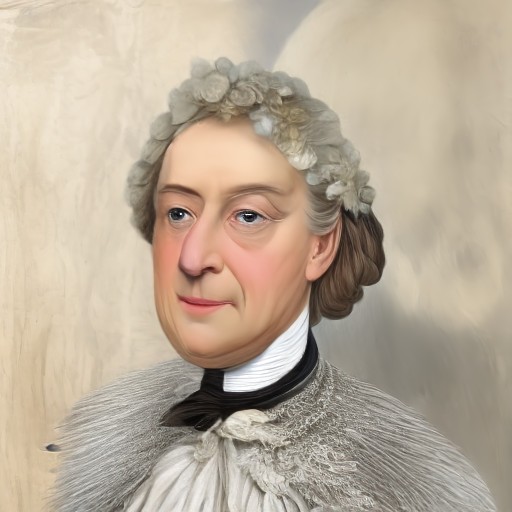}
    \\

    \raisebox{1.1\height}{\parbox[t]{12mm}{\rotatebox[origin=c]{90}{\makecell{Viking warrior \\ portrait with a \\ braided beard}}}}
    & \includegraphics[width=0.151\textwidth]{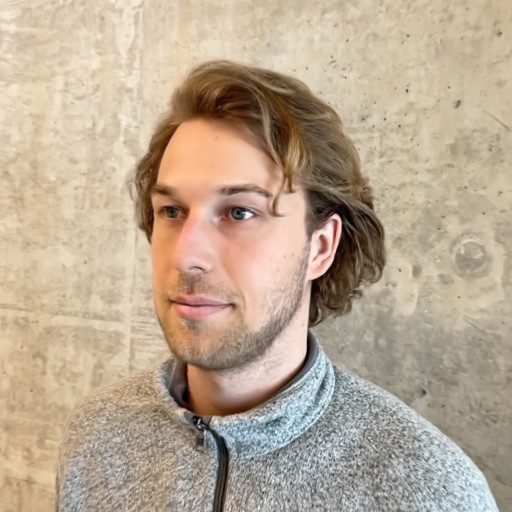}
    & \includegraphics[width=0.151\textwidth]{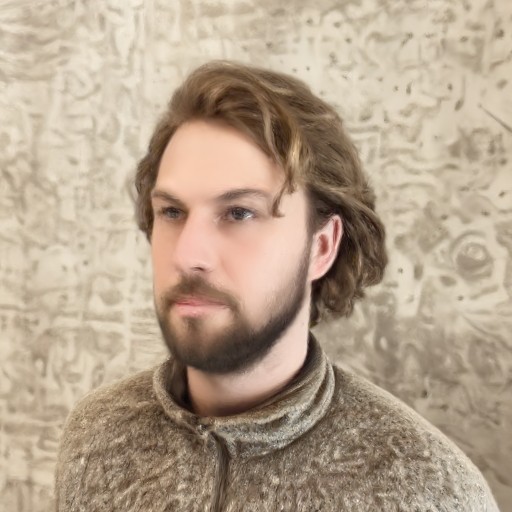}
    & \includegraphics[width=0.151\textwidth]{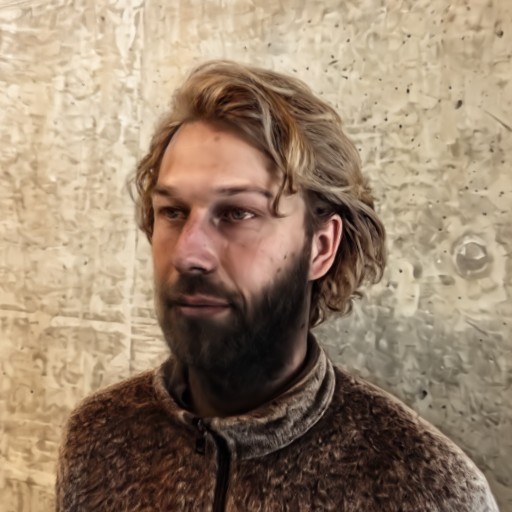}
    & \includegraphics[width=0.151\textwidth]{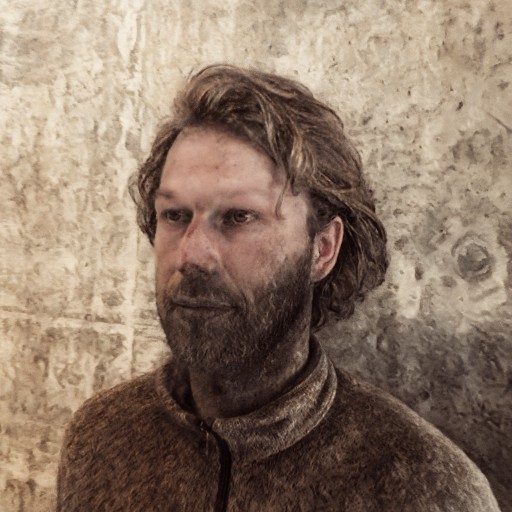}
    & \includegraphics[width=0.151\textwidth]{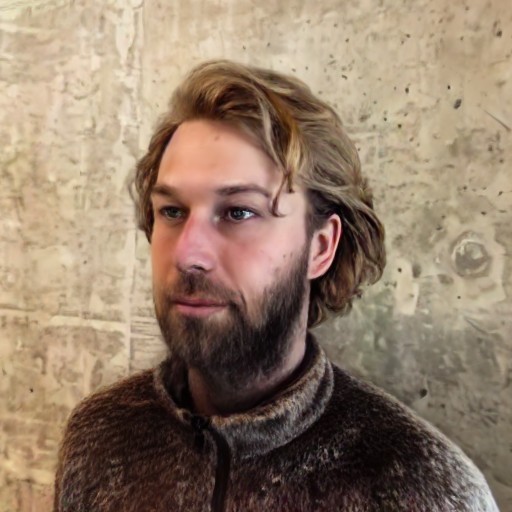}
    & \includegraphics[width=0.151\textwidth]{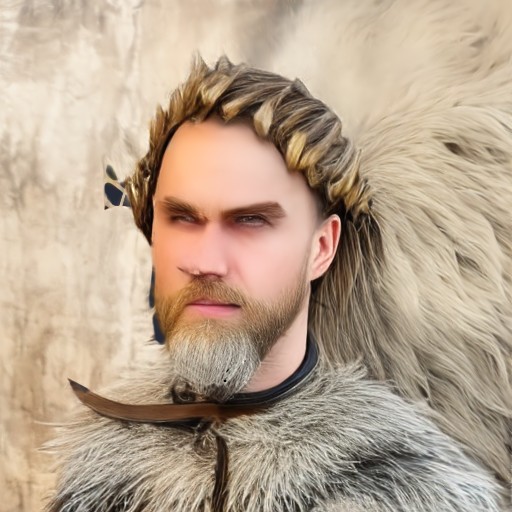}
    \\

    \raisebox{1.0\height}{\parbox[t]{12mm}{\rotatebox[origin=c]{90}{\makecell{a snowman on a \\ table in the \\ garden in snow}}}}
    & \includegraphics[width=0.151\textwidth]{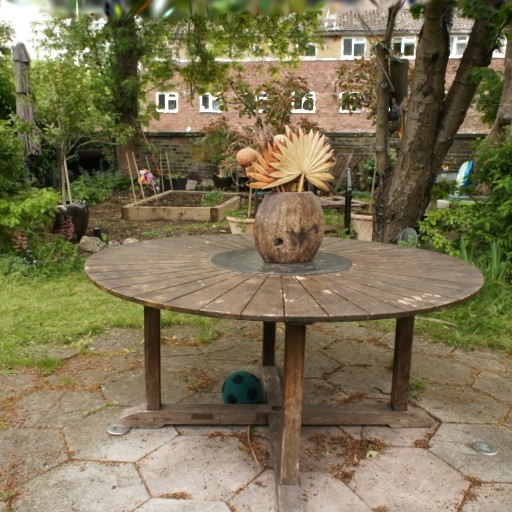}
    & \includegraphics[width=0.151\textwidth]{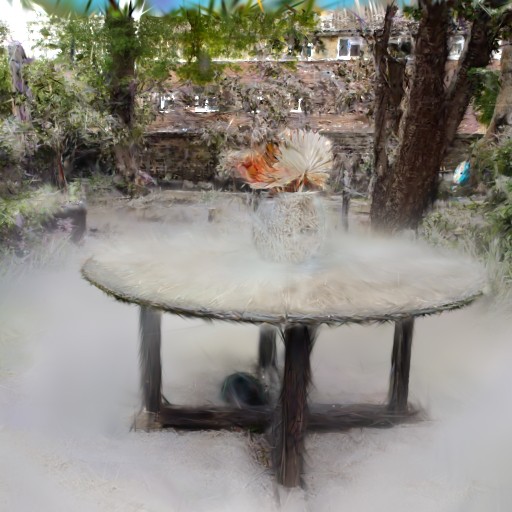}
    & \includegraphics[width=0.151\textwidth]{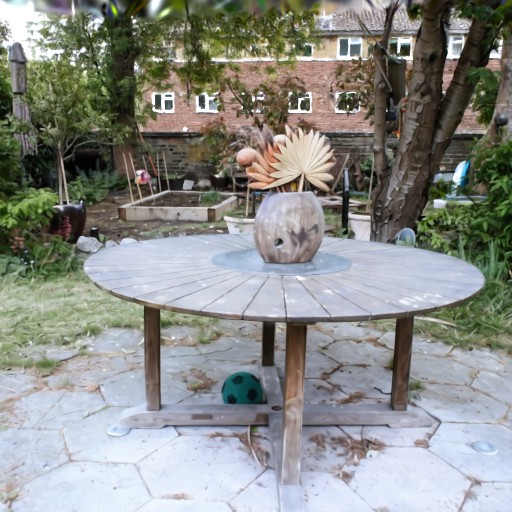}
    & \includegraphics[width=0.151\textwidth]{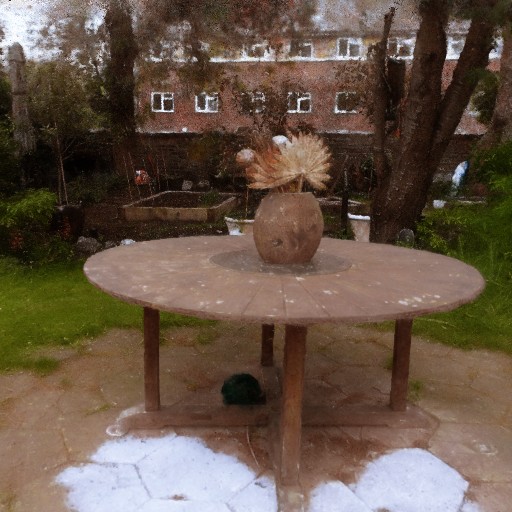}
    & \includegraphics[width=0.151\textwidth]{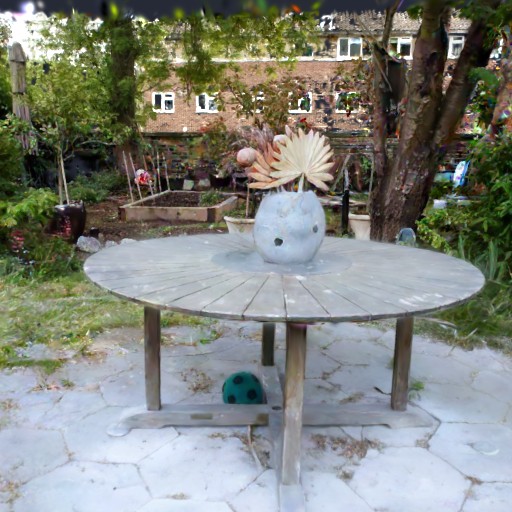}
    & \includegraphics[width=0.151\textwidth]{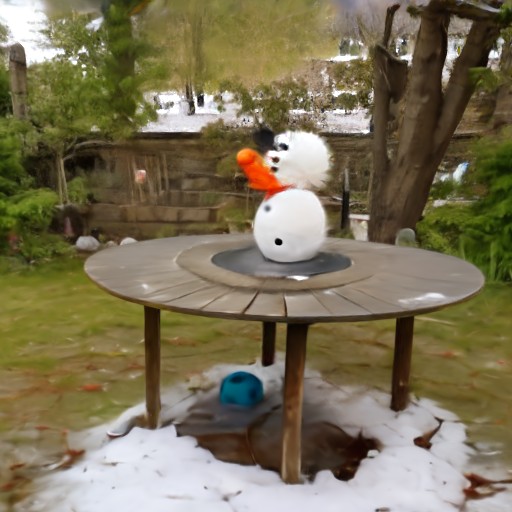}
    \\

    \raisebox{1.1\height}{\parbox[t]{12mm}{\rotatebox[origin=c]{90}{\makecell{a halloween \\ garden with \\ pumpkins}}}}
    & \includegraphics[width=0.151\textwidth]{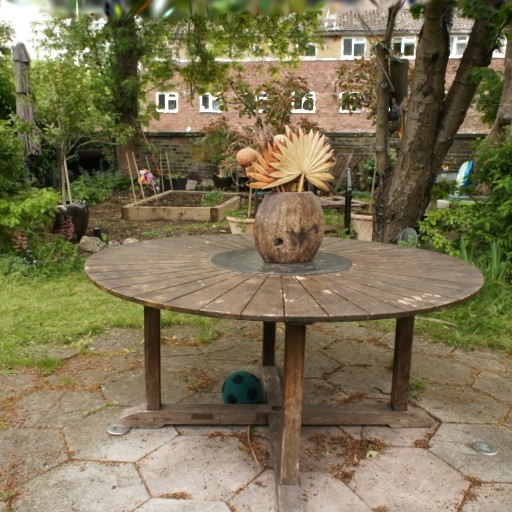}
    & \includegraphics[width=0.151\textwidth]{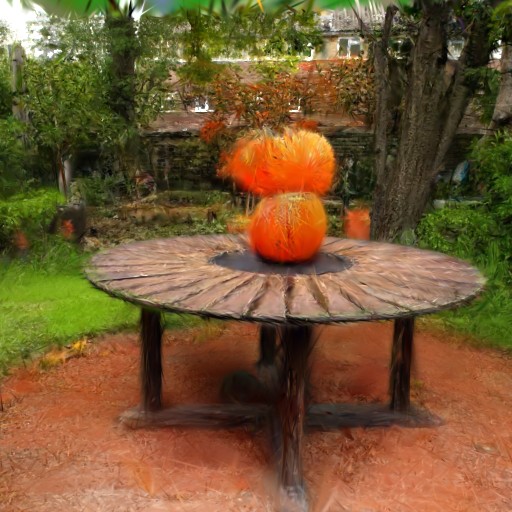}
    & \includegraphics[width=0.151\textwidth]{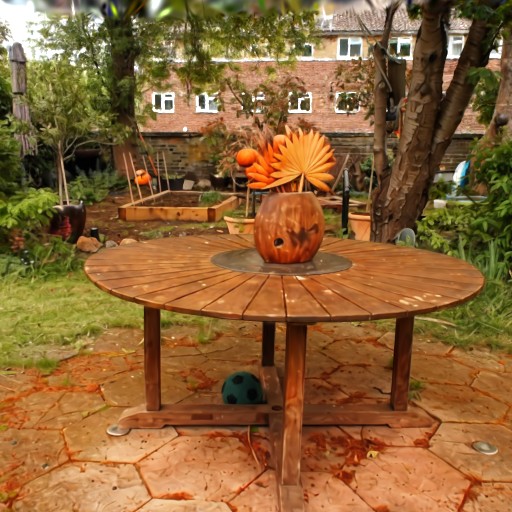}
    & \includegraphics[width=0.151\textwidth]{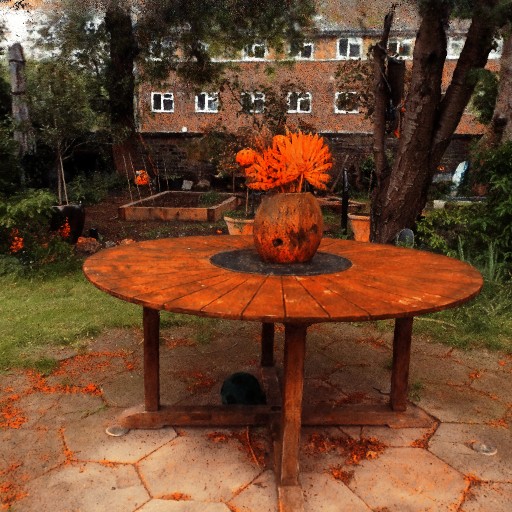}
    & \includegraphics[width=0.151\textwidth]{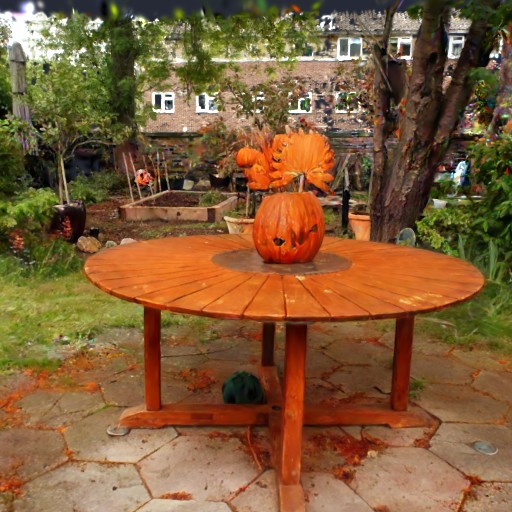}
    & \includegraphics[width=0.151\textwidth]{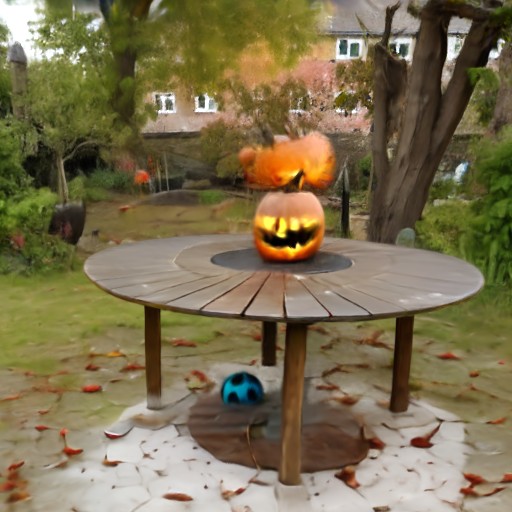}
    \\

    % \raisebox{1.0\height}{\parbox[t]{12mm}{\rotatebox[origin=c]{90}{\makecell{a photo of a \\ fountain in the \\ desert}}}}
    % & \includegraphics[width=0.151\textwidth]{figures/comparisons_figure/raw_jpg_small/mipnerf_garden_a_photo_of_a_fountain_in_the_desert_original.jpg}
    % & \includegraphics[width=0.151\textwidth]{figures/comparisons_figure/raw_jpg_small/mipnerf_garden_a_photo_of_a_fountain_in_the_desert_gaussctrl.jpg}
    % & \includegraphics[width=0.151\textwidth]{figures/comparisons_figure/raw_jpg_small/mipnerf_garden_a_photo_of_a_fountain_in_the_desert_instruct_gs2gs.jpg}
    % & \includegraphics[width=0.151\textwidth]{figures/comparisons_figure/raw_jpg_small/mipnerf_garden_a_photo_of_a_fountain_in_the_desert_instruct_nerf2nerf.jpg}
    % & \includegraphics[width=0.151\textwidth]{figures/comparisons_figure/raw_jpg_small/mipnerf_garden_a_photo_of_a_fountain_in_the_desert_DGE.jpg}
    % & \includegraphics[width=0.151\textwidth]{figures/comparisons_figure/raw_jpg_small/mipnerf_garden_a_photo_of_a_fountain_in_the_desert_ours.jpg}
    % \\

    \raisebox{1.0\height}{\parbox[t]{12mm}{\rotatebox[origin=c]{90}{\makecell{a Picasso \\ painting of a \\ microbiology lab}}}}
    & \includegraphics[width=0.151\textwidth]{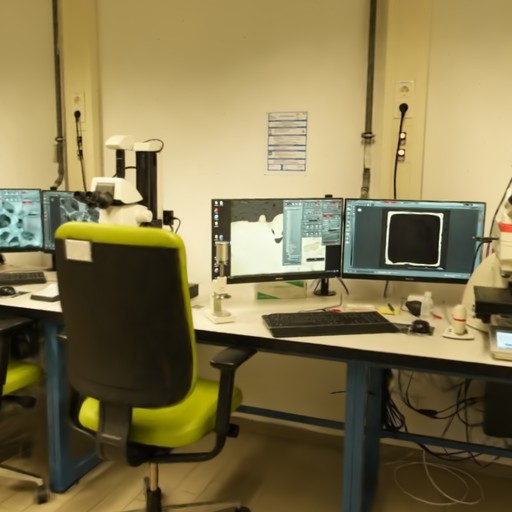}
    & \includegraphics[width=0.151\textwidth]{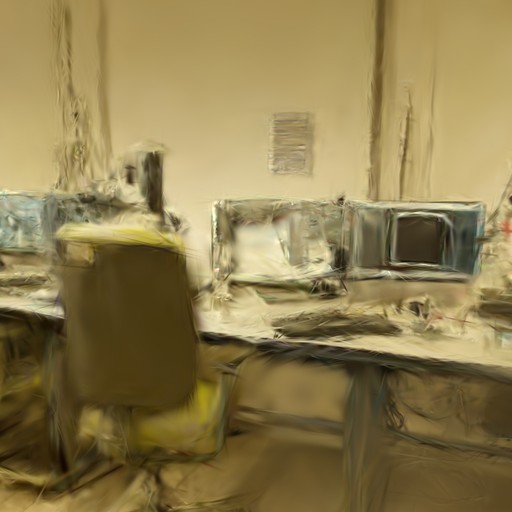}
    & \includegraphics[width=0.151\textwidth]{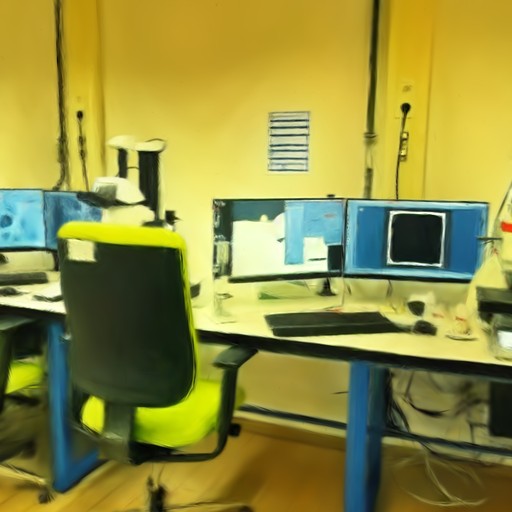}
    & \includegraphics[width=0.151\textwidth]{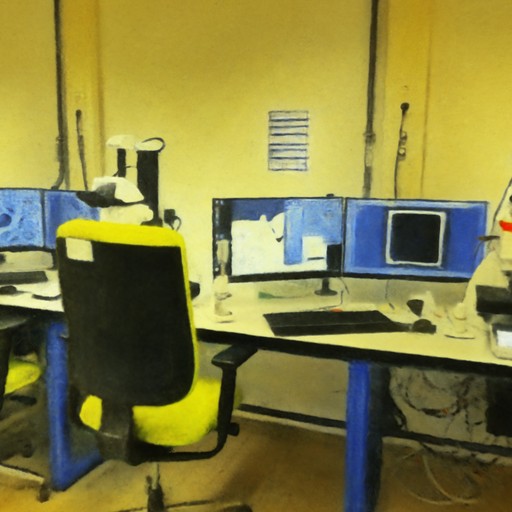}
    & \includegraphics[width=0.151\textwidth]{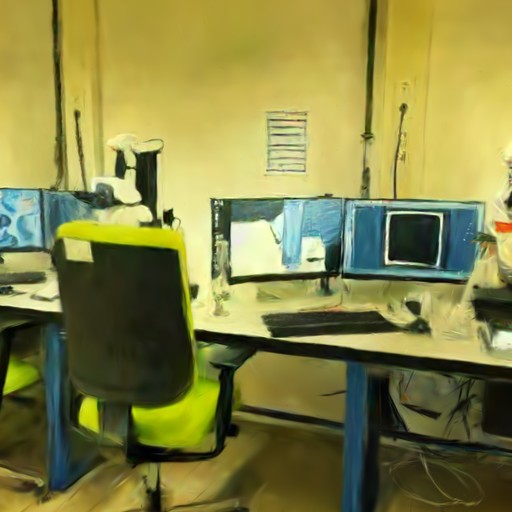}
    & \includegraphics[width=0.151\textwidth]{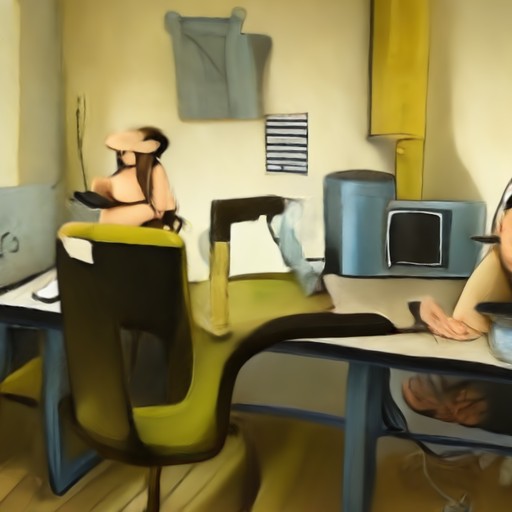}
    \\

    \raisebox{1.0\height}{\parbox[t]{12mm}{\rotatebox[origin=c]{90}{\makecell{futuristic night \\ city Cyberpunk \\ style robotics lab}}}}
    & \includegraphics[width=0.151\textwidth]{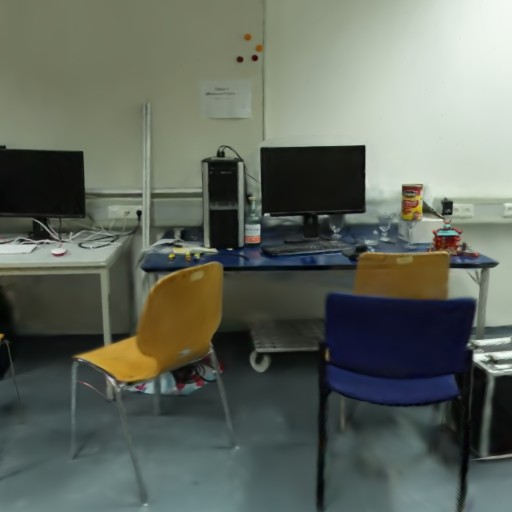}
    & \includegraphics[width=0.151\textwidth]{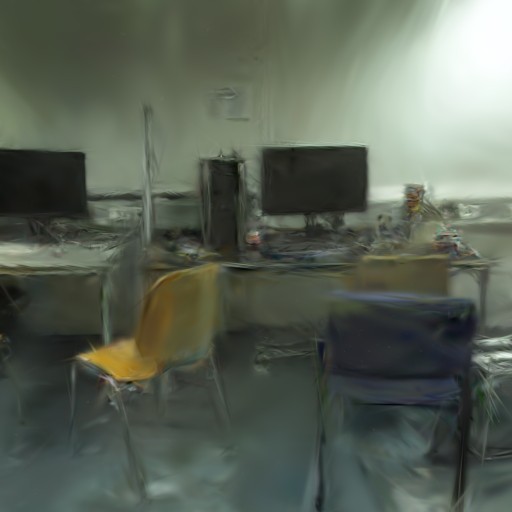}
    & \includegraphics[width=0.151\textwidth]{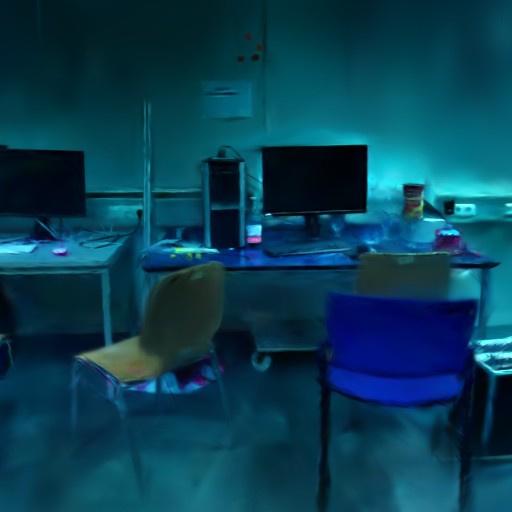}
    & \includegraphics[width=0.151\textwidth]{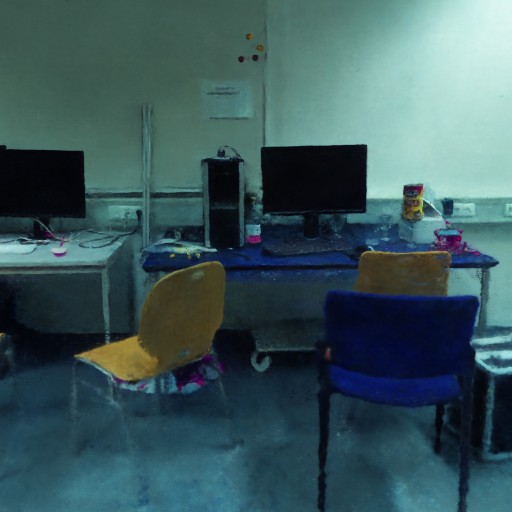}
    & \includegraphics[width=0.151\textwidth]{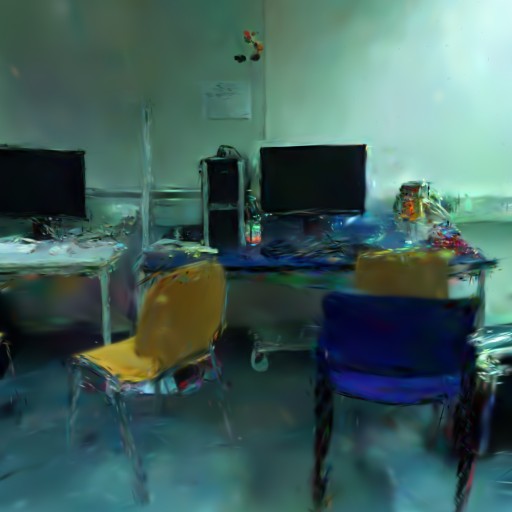}
    & \includegraphics[width=0.151\textwidth]{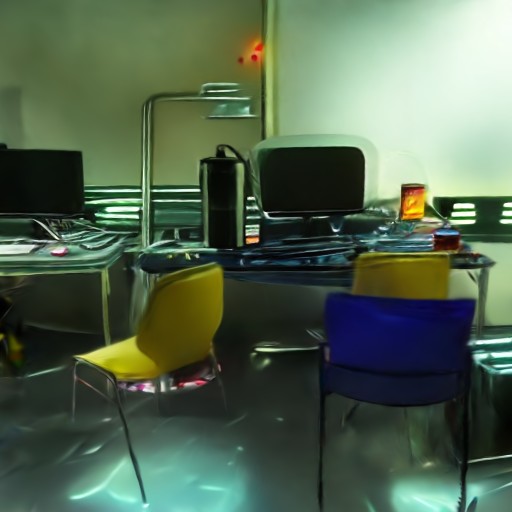}
    \\

    \raisebox{1.0\height}{\parbox[t]{12mm}{\rotatebox[origin=c]{90}{\makecell{1950s diner, \\ checkered floors, \\ red vinyl seats}}}}
    & \includegraphics[width=0.151\textwidth]{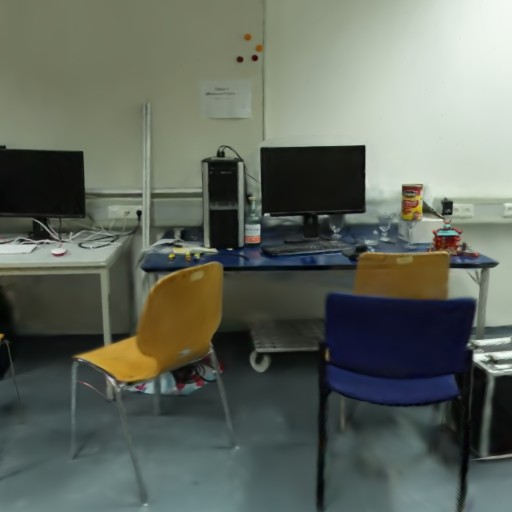}
    & \includegraphics[width=0.151\textwidth]{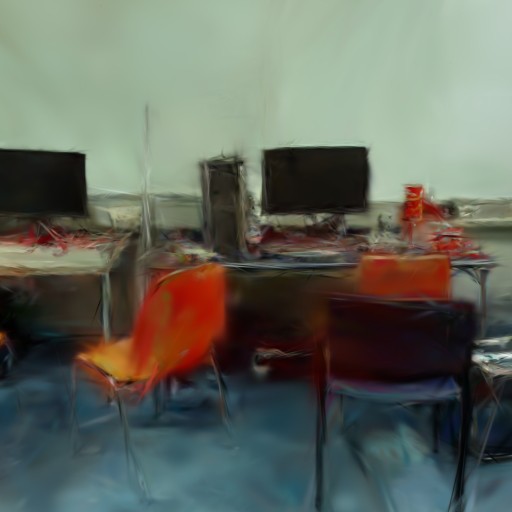}
    & \includegraphics[width=0.151\textwidth]{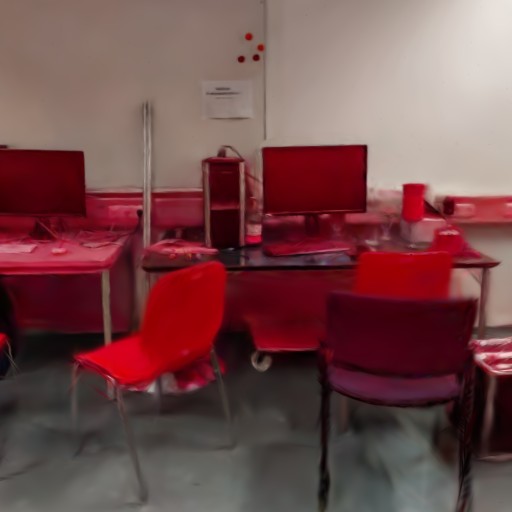}
    & \includegraphics[width=0.151\textwidth]{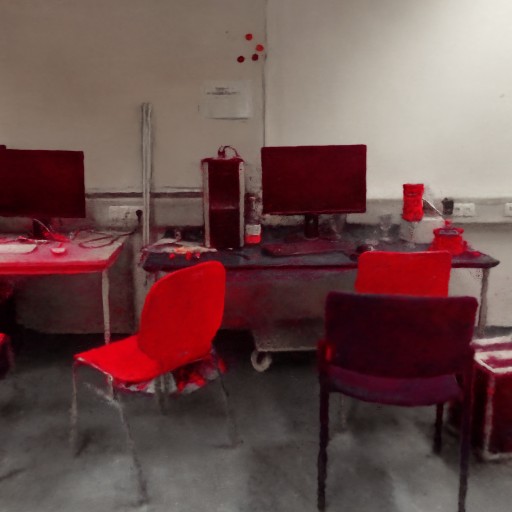}
    & \includegraphics[width=0.151\textwidth]{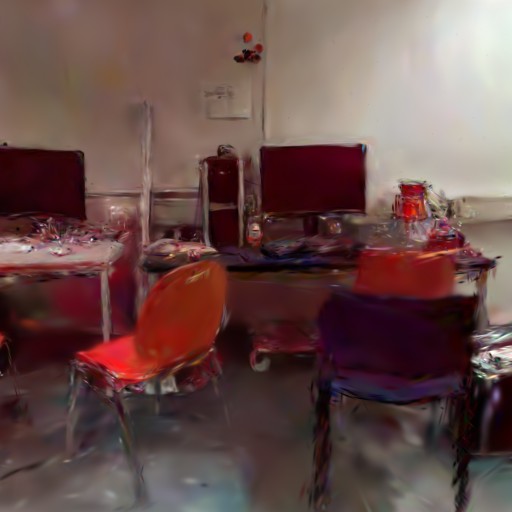}
    & \includegraphics[width=0.151\textwidth]{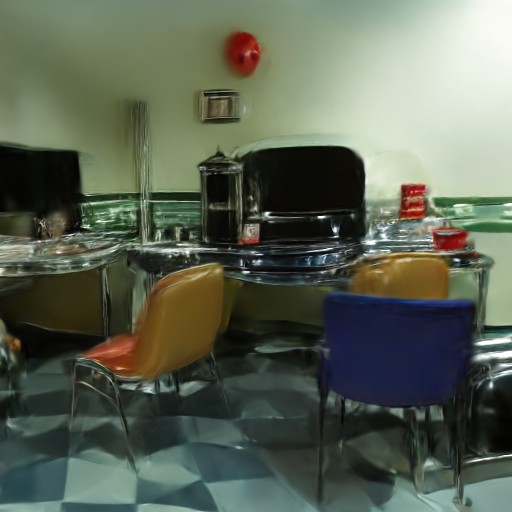}
    \\

\end{tabular}

    \vspace{-5pt}
    \caption{\textbf{Further Qualitative Comparison from Novel Views} Our method's ability to change the scene's shape allow its stylizations to be more aesthetically pleasing and exhibit more adherence to the style prompt.}
    \vspace{-5pt}
    \label{fig:qualitative-comparison-supp}    
\end{figure*}

\begin{figure*}
    \vspace{-5pt}
    \centering
    \small
    
    \setlength\tabcolsep{0pt}
\renewcommand{\arraystretch}{0}
\begin{tabular}{ccccccc}
    \centering

     & Original & GaussCtrl~\cite{wu2024gaussctrl} & Instruct-GS2GS~\cite{vachha2024instruct} & Instruct-N2N~\cite{haque2023instruct} & DGE~\cite{chen2024dge} & Ours \\

    \raisebox{1.0\height}{\parbox[t]{12mm}{\rotatebox[origin=c]{90}{\makecell{a Fauvism \\ painting of a \\ conference room}}}}
    & \includegraphics[width=0.142\textwidth]{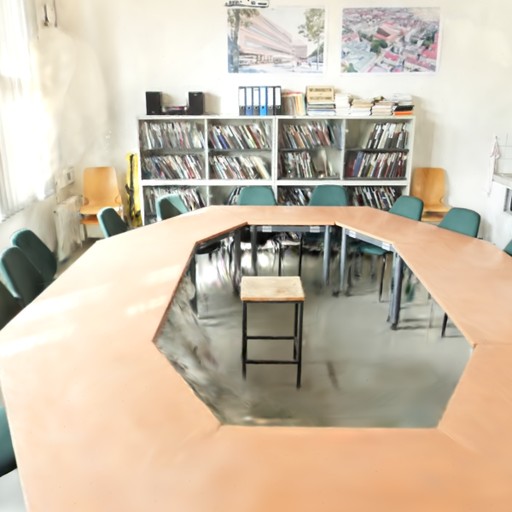}
    & \includegraphics[width=0.142\textwidth]{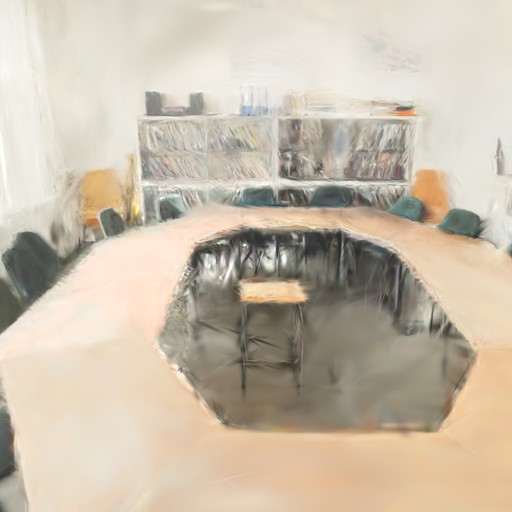}
    & \includegraphics[width=0.142\textwidth]{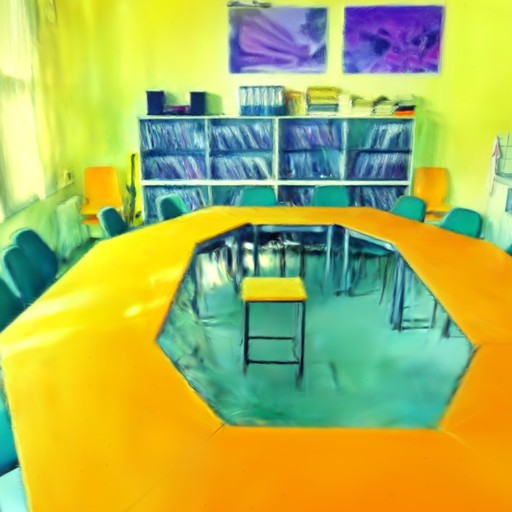}
    & \includegraphics[width=0.142\textwidth]{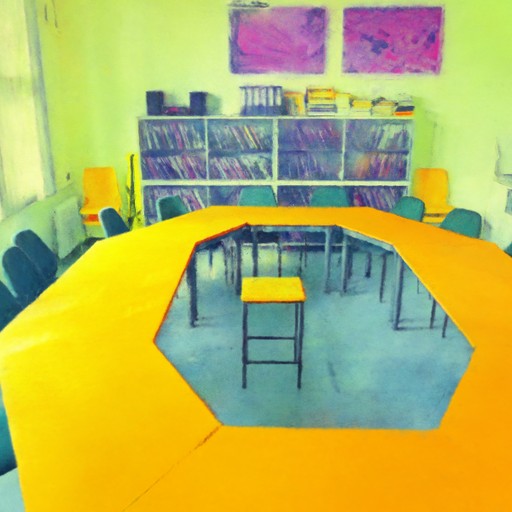}
    & \includegraphics[width=0.142\textwidth]{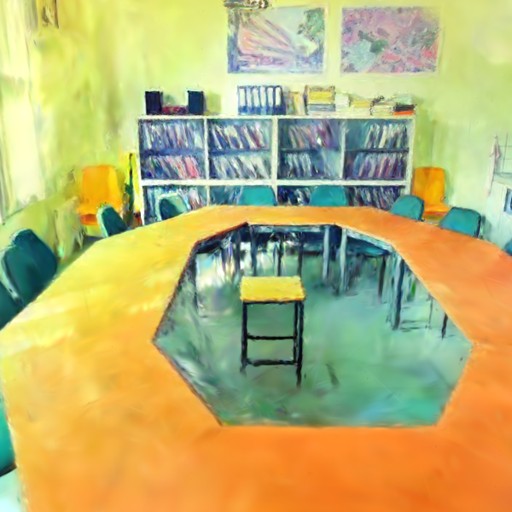}
    & \includegraphics[width=0.142\textwidth]{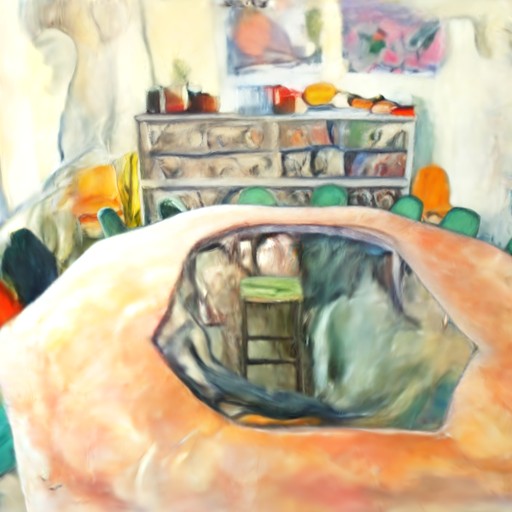}
    \\

    % \raisebox{1.0\height}{\parbox[t]{12mm}{\rotatebox[origin=c]{90}{\makecell{a Hiroshige \\ Utagawa \\ painting of a \\ kitchen}}}}
    % & \includegraphics[width=0.142\textwidth]{figures/comparisons_figure/raw_jpg_small/scannet_h_kitchen_7313597891086098452_original.jpg}
    % & \includegraphics[width=0.142\textwidth]{figures/comparisons_figure/raw_jpg_small/scannet_h_kitchen_7313597891086098452_gaussctrl.jpg}
    % & \includegraphics[width=0.142\textwidth]{figures/comparisons_figure/raw_jpg_small/scannet_h_kitchen_7313597891086098452_instruct_gs2gs.jpg}
    % & \includegraphics[width=0.142\textwidth]{figures/comparisons_figure/raw_jpg_small/scannet_h_kitchen_7313597891086098452_instruct_nerf2nerf.jpg}
    % & \includegraphics[width=0.142\textwidth]{figures/comparisons_figure/raw_jpg_small/scannet_h_kitchen_7313597891086098452_DGE.jpg}
    % & \includegraphics[width=0.142\textwidth]{figures/comparisons_figure/raw_jpg_small/scannet_h_kitchen_7313597891086098452_ours.jpg}
    % \\

    \raisebox{1.0\height}{\parbox[t]{12mm}{\rotatebox[origin=c]{90}{\makecell{Bastion game \\ style picture flat \\ of a bathroom}}}}
    & \includegraphics[width=0.142\textwidth]{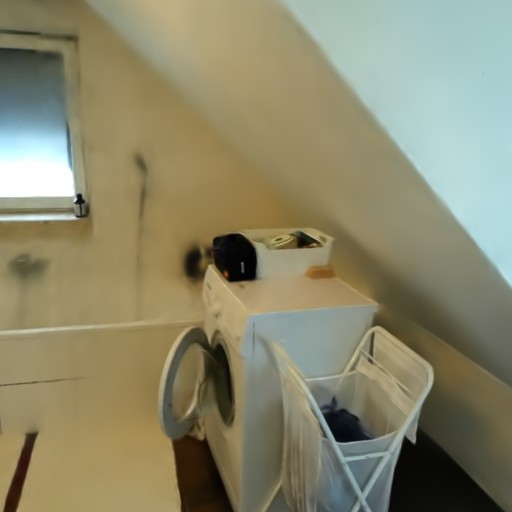}
    & \includegraphics[width=0.142\textwidth]{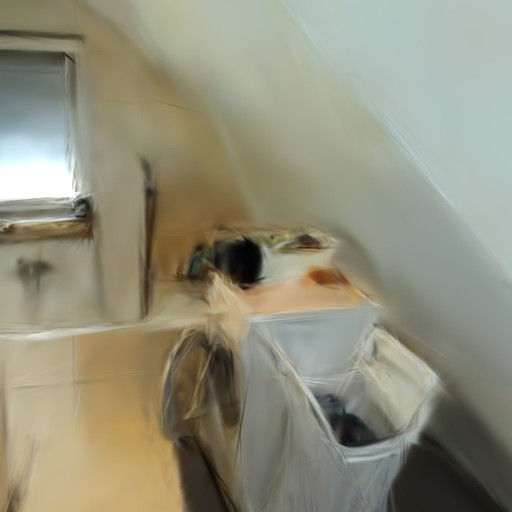}
    & \includegraphics[width=0.142\textwidth]{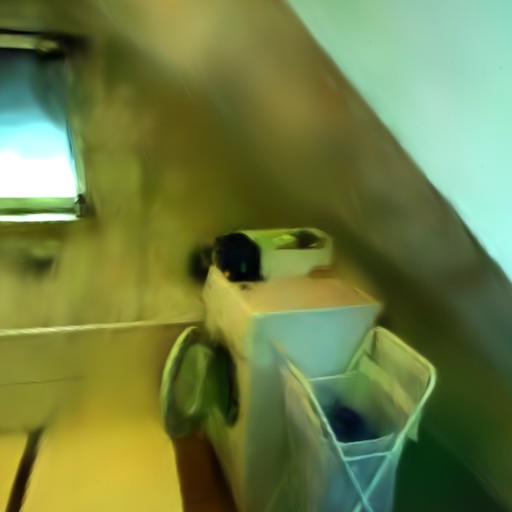}
    & \includegraphics[width=0.142\textwidth]{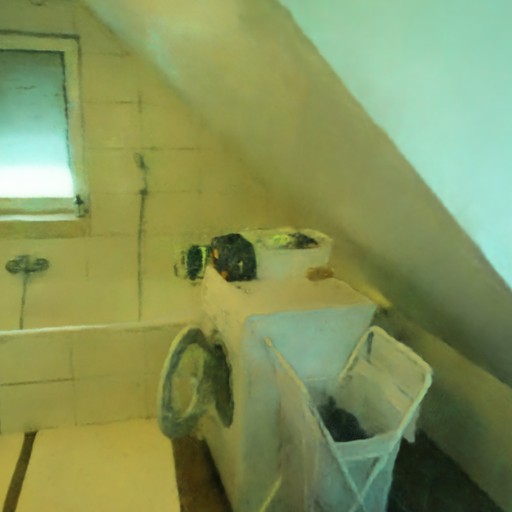}
    & \includegraphics[width=0.142\textwidth]{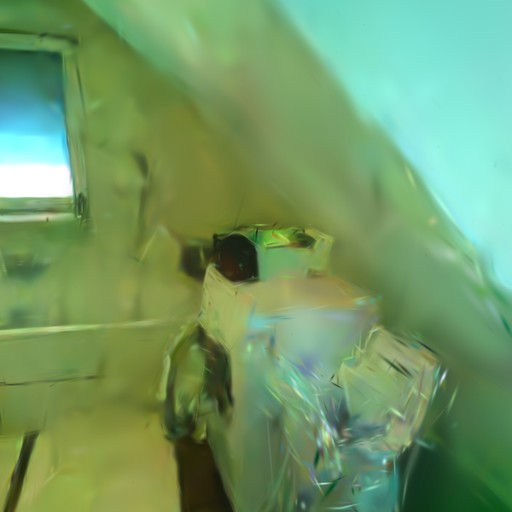}
    & \includegraphics[width=0.142\textwidth]{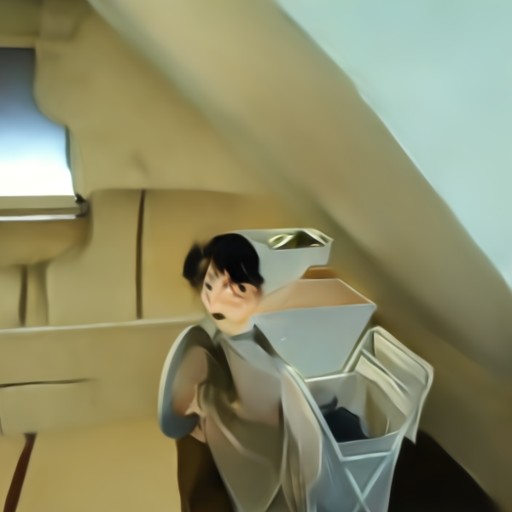}
    \\

    \raisebox{1.0\height}{\parbox[t]{12mm}{\rotatebox[origin=c]{90}{\makecell{a photo of a \\ robot dinosaur \\ on the road side}}}}
    & \includegraphics[width=0.142\textwidth]{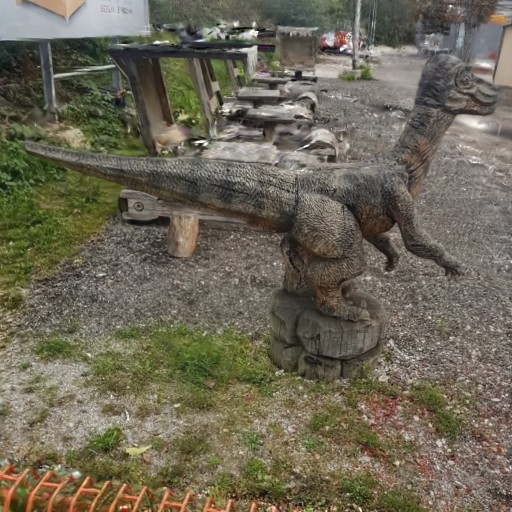}
    & \includegraphics[width=0.142\textwidth]{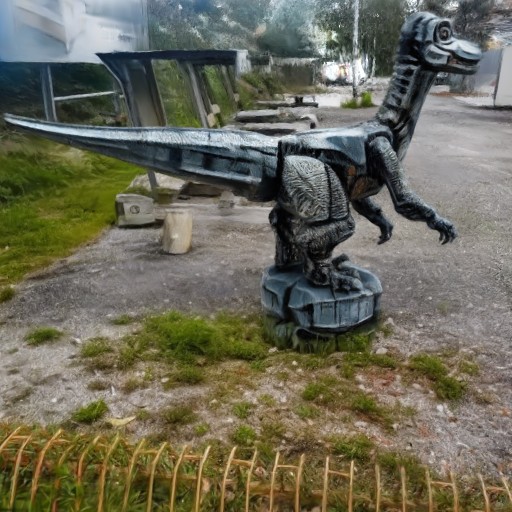}
    & \includegraphics[width=0.142\textwidth]{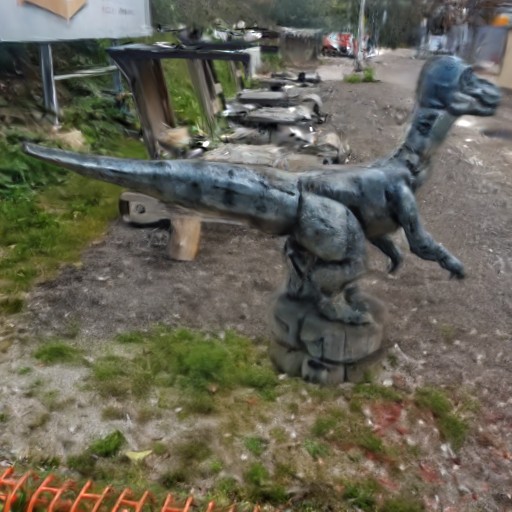}
    & \includegraphics[width=0.142\textwidth]{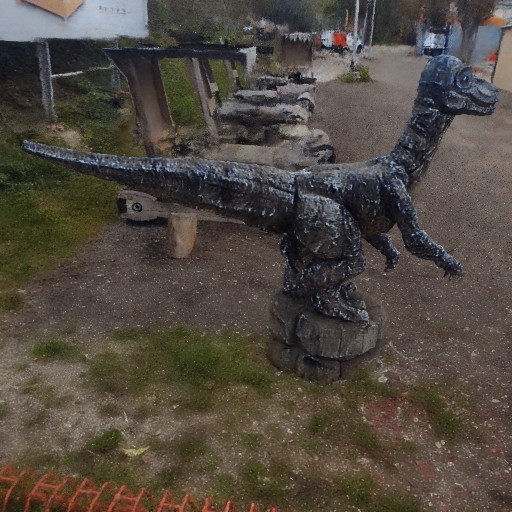}
    & \includegraphics[width=0.142\textwidth]{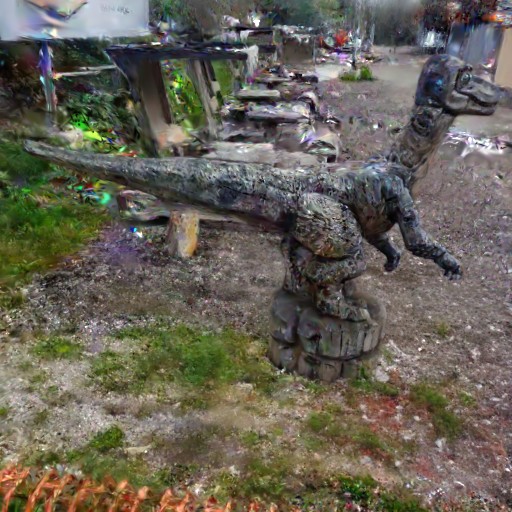}
    & \includegraphics[width=0.142\textwidth]{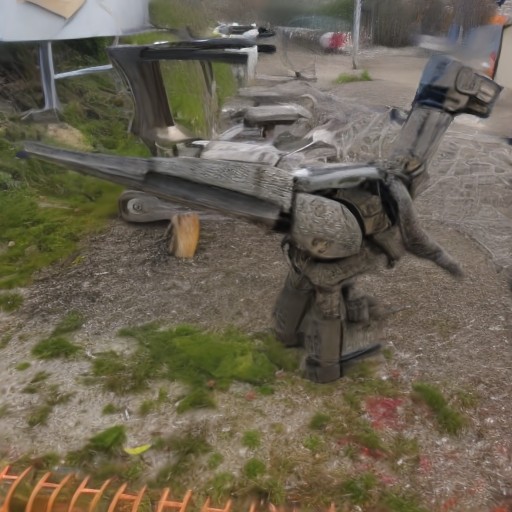}
    \\

    \raisebox{1.0\height}{\parbox[t]{12mm}{\rotatebox[origin=c]{90}{\makecell{a photo of a \\ giraffe in \\ a zoo}}}}
    & \includegraphics[width=0.142\textwidth]{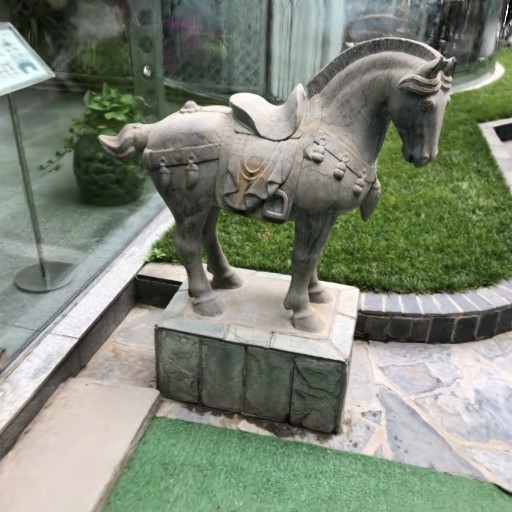}
    & \includegraphics[width=0.142\textwidth]{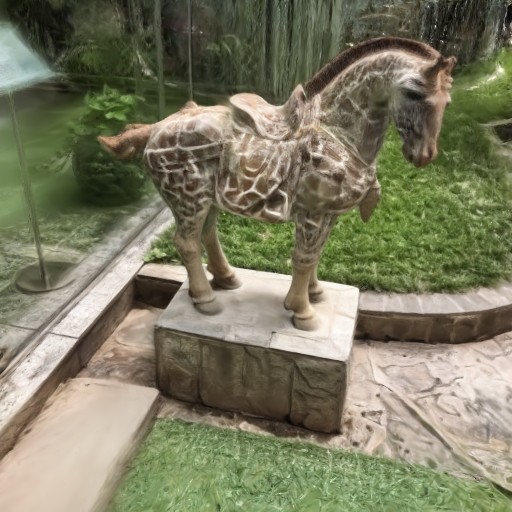}
    & \includegraphics[width=0.142\textwidth]{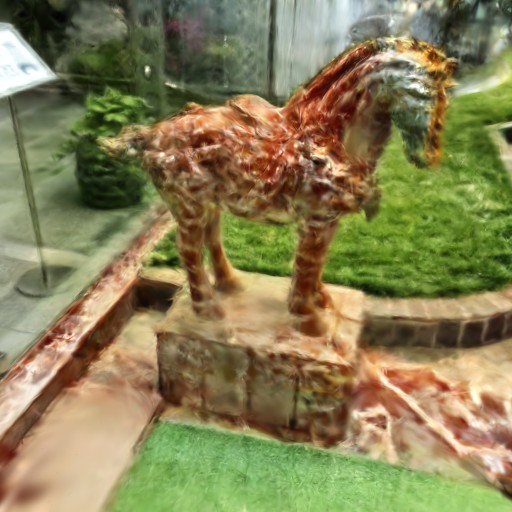}
    & \includegraphics[width=0.142\textwidth]{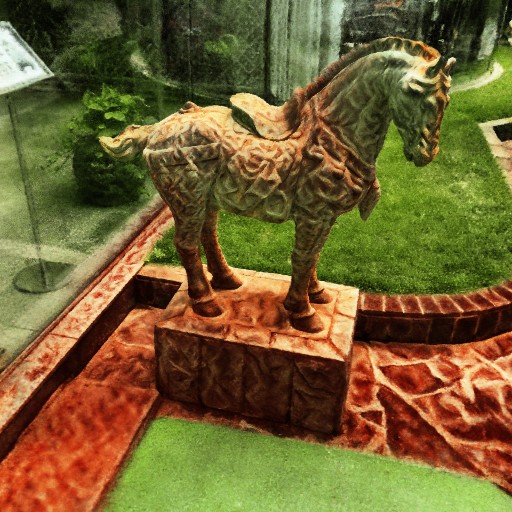}
    & \includegraphics[width=0.142\textwidth]{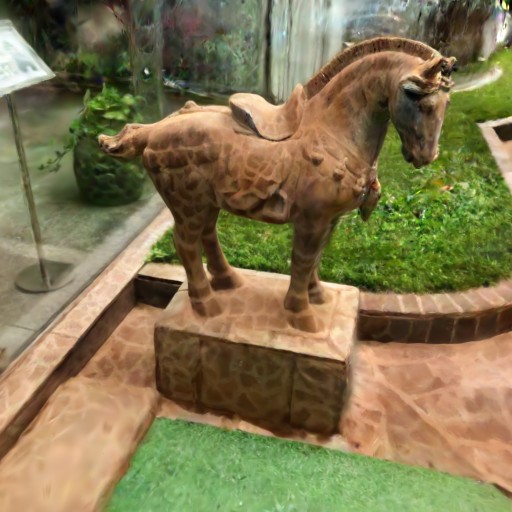}
    & \includegraphics[width=0.142\textwidth]{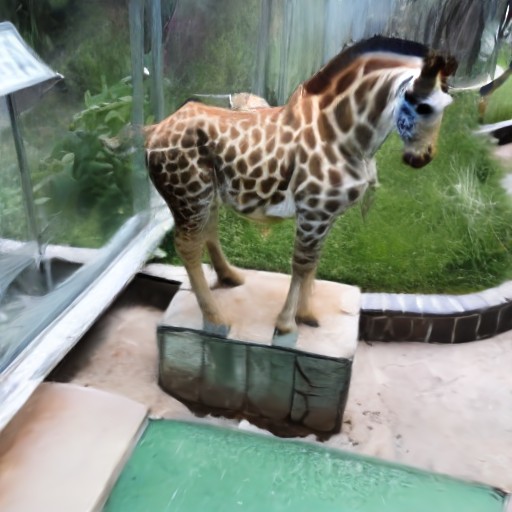}
    \\

    \raisebox{1.0\height}{\parbox[t]{12mm}{\rotatebox[origin=c]{90}{\makecell{Horse statue \\ made from very \\ large lego bricks}}}}
    & \includegraphics[width=0.142\textwidth]{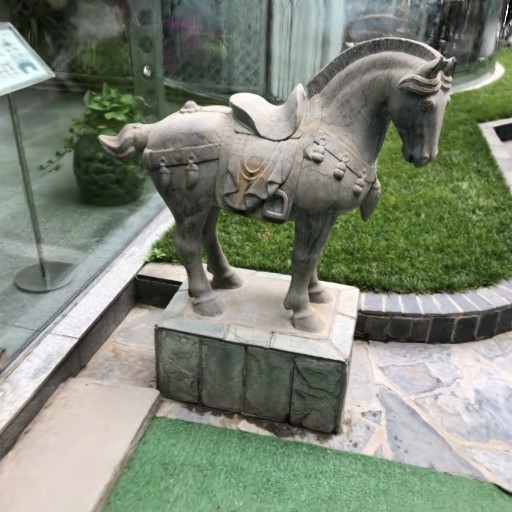}
    & \includegraphics[width=0.142\textwidth]{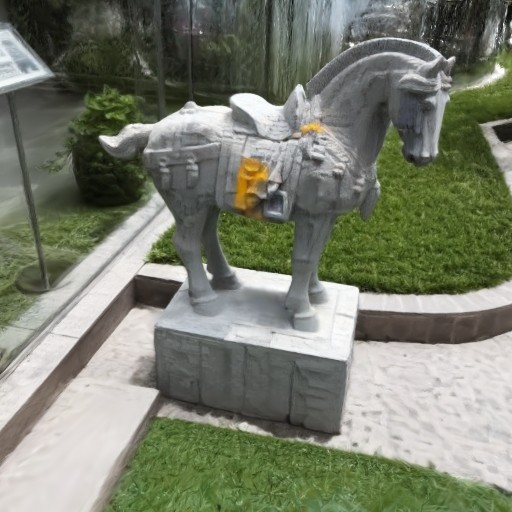}
    & \includegraphics[width=0.142\textwidth]{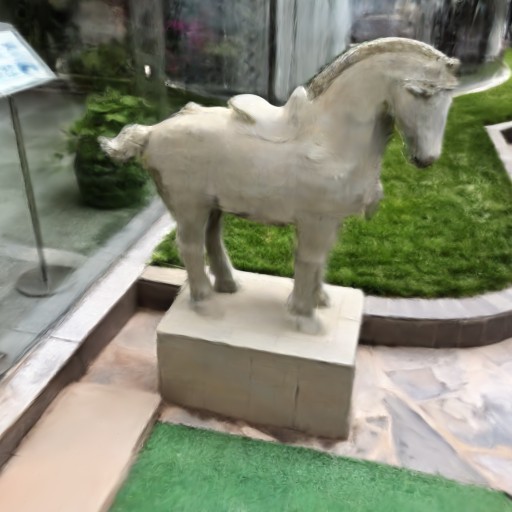}
    & \includegraphics[width=0.142\textwidth]{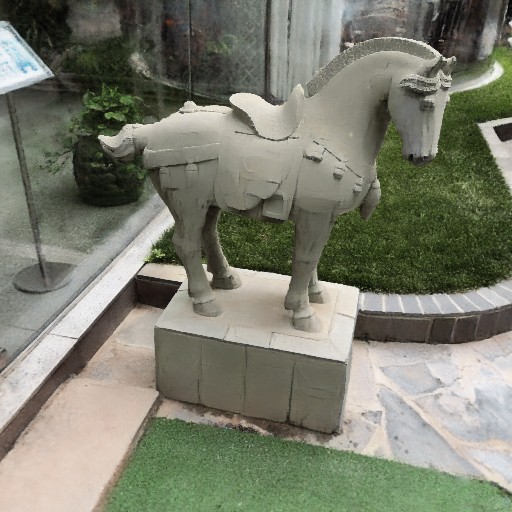}
    & \includegraphics[width=0.142\textwidth]{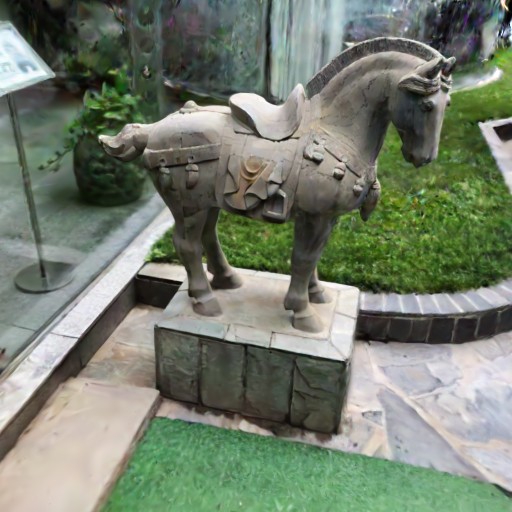}
    & \includegraphics[width=0.142\textwidth]{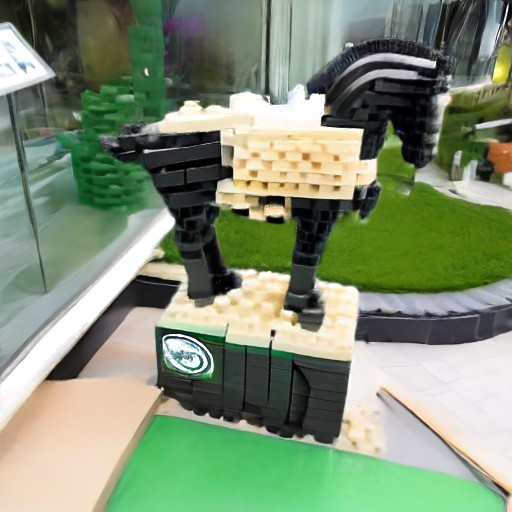}
    \\

    \raisebox{1.0\height}{\parbox[t]{12mm}{\rotatebox[origin=c]{90}{\makecell{tron legacy \\ nightrider neon  \\ futuristic night}}}}
    & \includegraphics[width=0.142\textwidth]{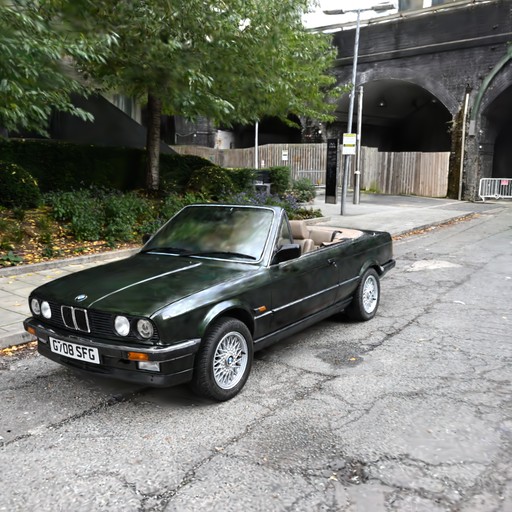}
    & \includegraphics[width=0.142\textwidth]{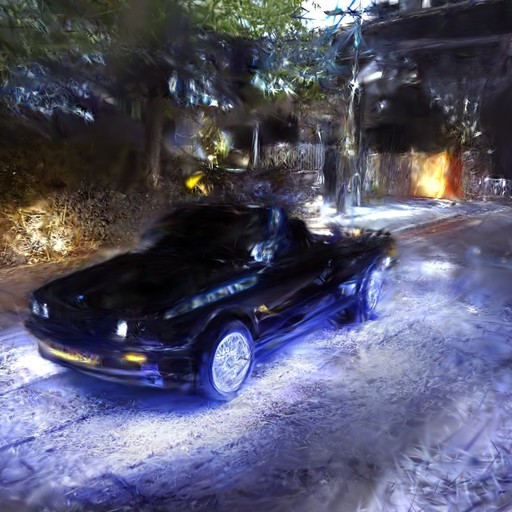}
    & \includegraphics[width=0.142\textwidth]{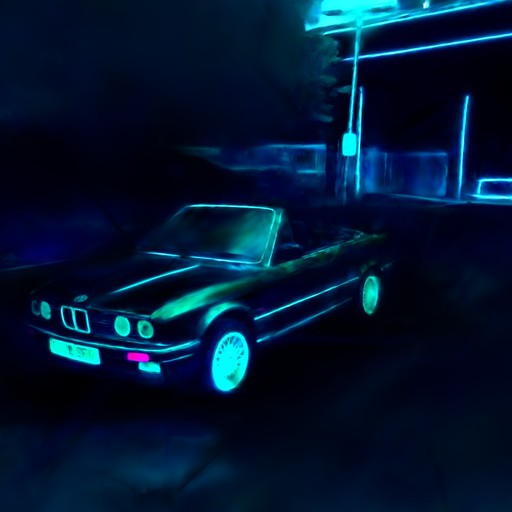}
    & \includegraphics[width=0.142\textwidth]{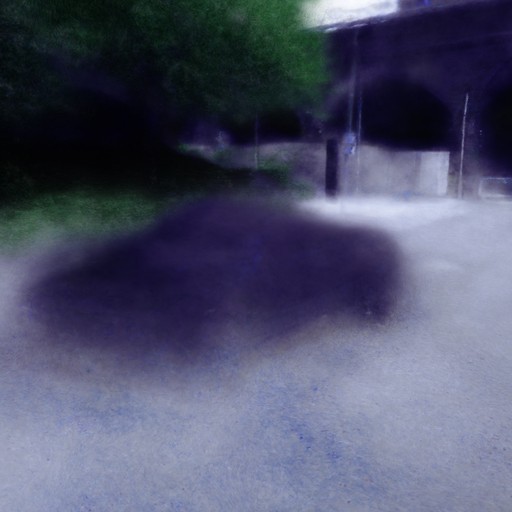}
    & \includegraphics[width=0.142\textwidth]{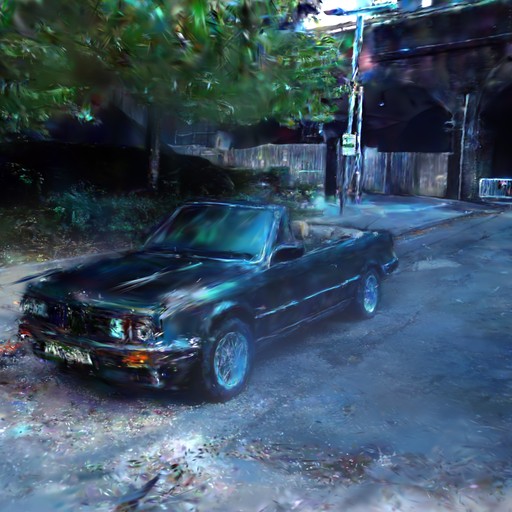}
    & \includegraphics[width=0.142\textwidth]{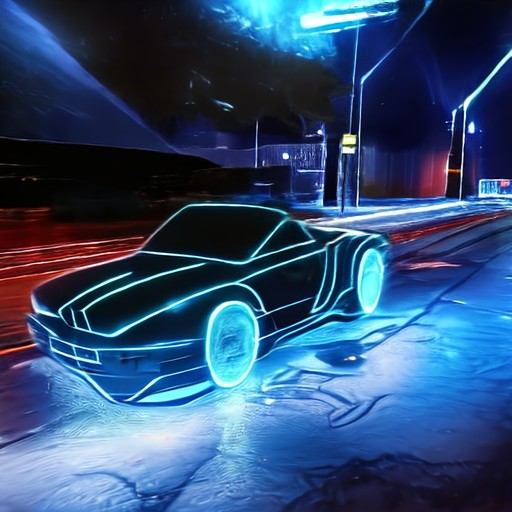}
    \\

    \raisebox{1.1\height}{\parbox[t]{4mm}{\rotatebox[origin=c]{90}{\makecell{a photo of a \\ Batmobile}}}}
    & \includegraphics[width=0.142\textwidth]{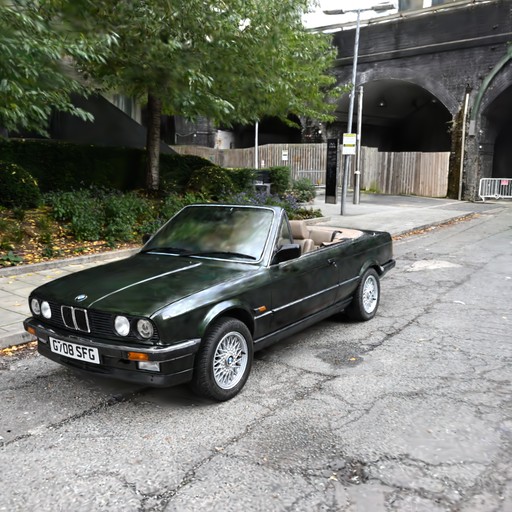}
    & \includegraphics[width=0.142\textwidth]{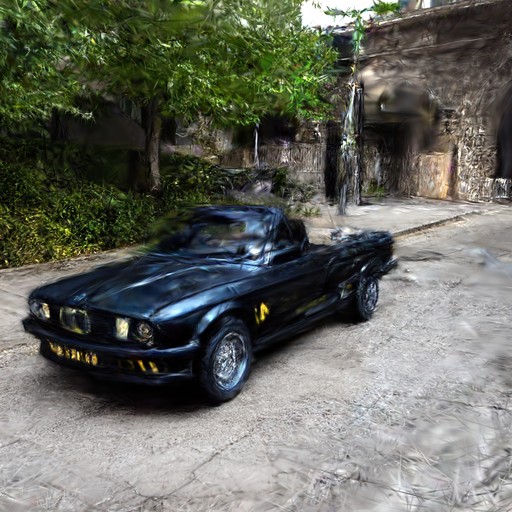}
    & \includegraphics[width=0.142\textwidth]{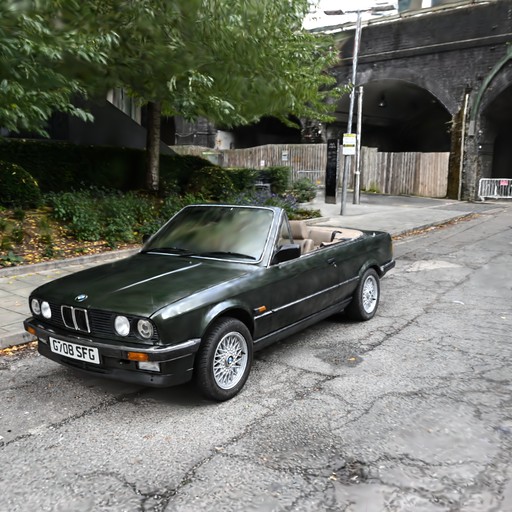}
    & \includegraphics[width=0.142\textwidth]{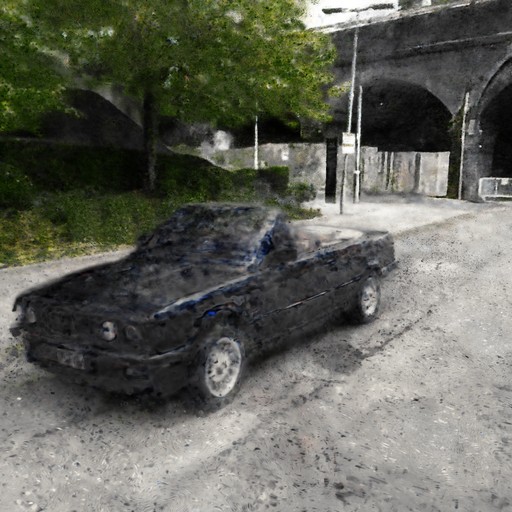}
    & \includegraphics[width=0.142\textwidth]{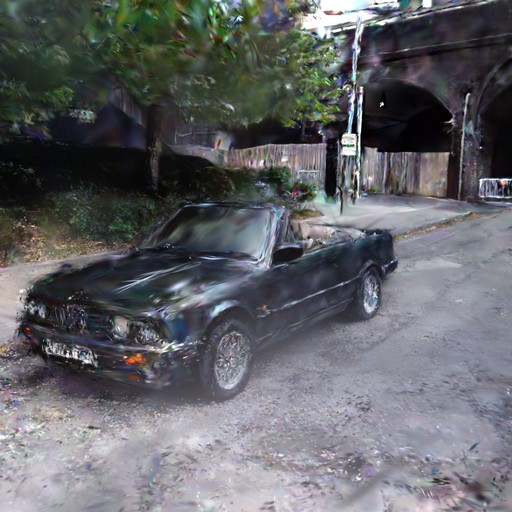}
    & \includegraphics[width=0.142\textwidth]{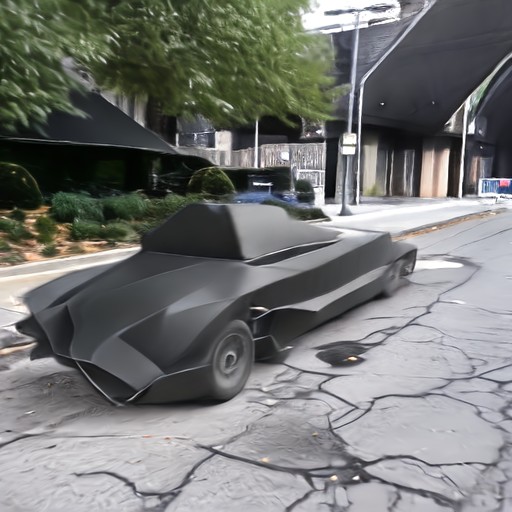}
    \\

    \raisebox{1.0\height}{\parbox[t]{4mm}{\rotatebox[origin=c]{90}{\makecell{a screenshot of \\ a Minecraft car}}}}
    & \includegraphics[width=0.142\textwidth]{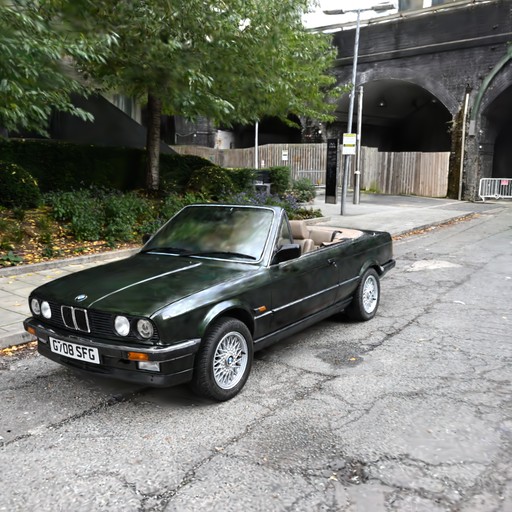}
    & \includegraphics[width=0.142\textwidth]{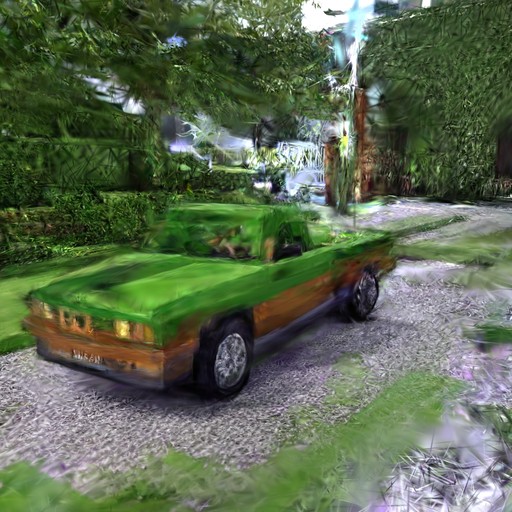}
    & \includegraphics[width=0.142\textwidth]{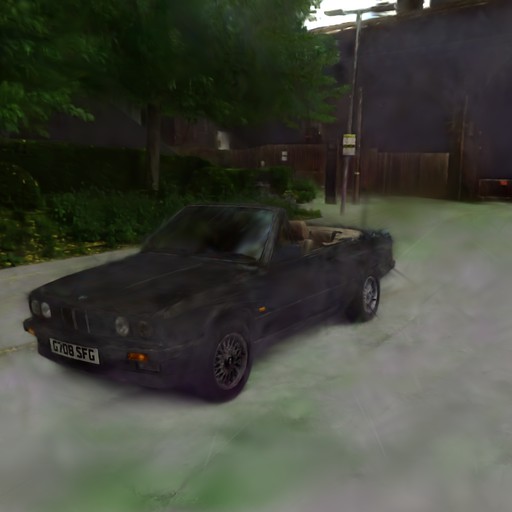}
    & \includegraphics[width=0.142\textwidth]{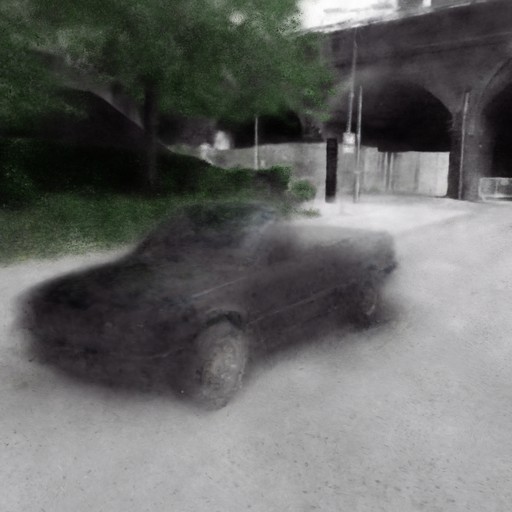}
    & \includegraphics[width=0.142\textwidth]{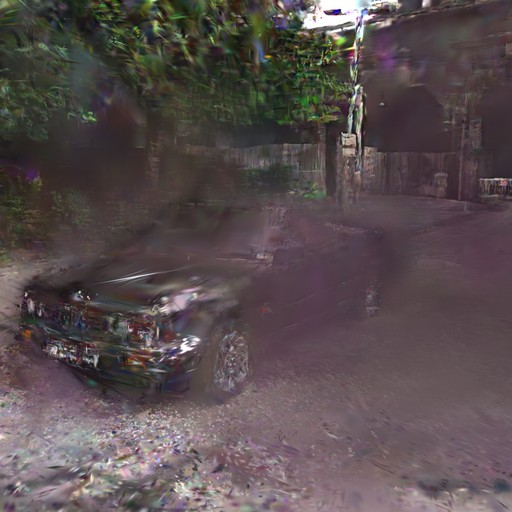}
    & \includegraphics[width=0.142\textwidth]{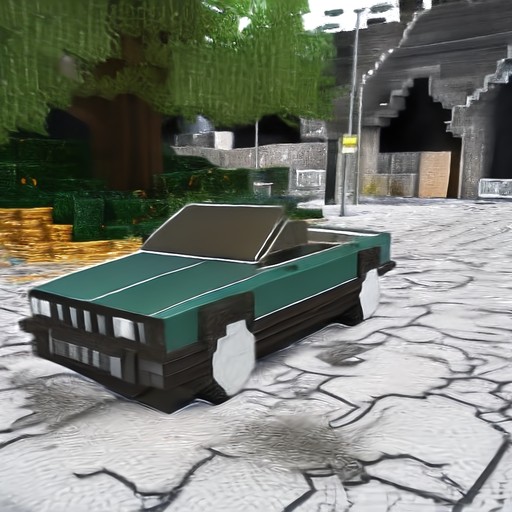}
    \\

\end{tabular}

    \vspace{-5pt}
    \caption{\textbf{Further Qualitative Comparison from Novel Views} Our method's ability to change the scene's shape allow its stylizations to be more aesthetically pleasing and exhibit more adherence to the style prompt.}
    \vspace{-5pt}
    \label{fig:qualitative-comparison-supp-2}    
\end{figure*}

We show additional qualitative results in Figure~\ref{fig:qualitative-comparison-supp} and Figure~\ref{fig:qualitative-comparison-supp-2}.

% \begin{figure*}
%     \vspace{-5pt}
%     \centering
%     \small
    
%     \include{figures/comparisons_figure/comparisons_transposed_supp_3}
    
%     \vspace{-5pt}
%     \caption{\todo{Caption}}
%     \vspace{-5pt}
%     \label{fig:qualitative-comparison-supp-3}    
% \end{figure*}

%%%%%%%%%%%%%%%%%%%%%%%%%%%%%%%%%%%%

%%%%%%%%%%%%%%%%%%%%%%%%%%%%%%%%%%%%
\section{Qualitative results of ablations}

We give further qualitative results for our ablations in fig.~\ref{fig:ablations-qualitative-comparison}, showing the impact of our cross-attention contributions.

\begin{figure*}
    \vspace{-5pt}
    \centering
    \small
    
    \setlength\tabcolsep{0pt}
\renewcommand{\arraystretch}{0}
\begin{tabular}{ccccc}
    \centering

     & Original & \makecell{Warp + RGB Inpaint \\ + Depth ControlNet} & \makecell{Warp + RGB Inpaint \\ + Depth ControlNet \\ + DAv2} & Ours \\

    \raisebox{1.2\height}{\parbox[t]{12mm}{\rotatebox[origin=c]{90}{\makecell{a polar bear in \\ the winter \\ forest}}}}
    & \includegraphics[width=0.16\textwidth]{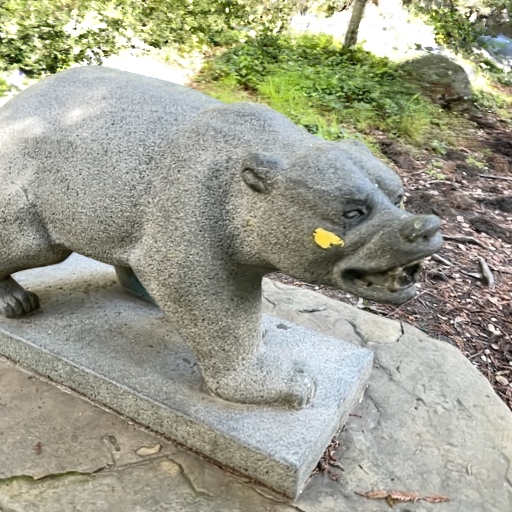}
    & \includegraphics[width=0.16\textwidth]{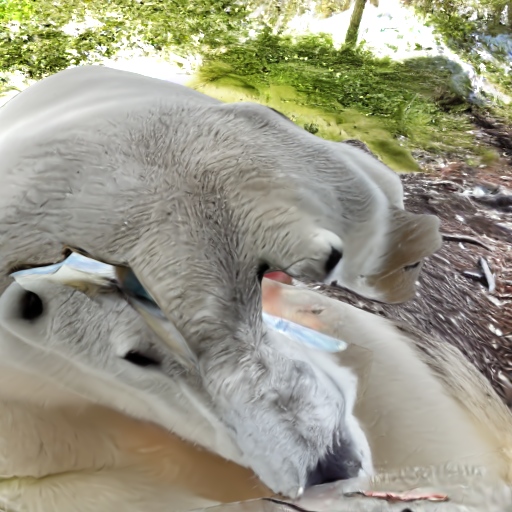}
    & \includegraphics[width=0.16\textwidth]{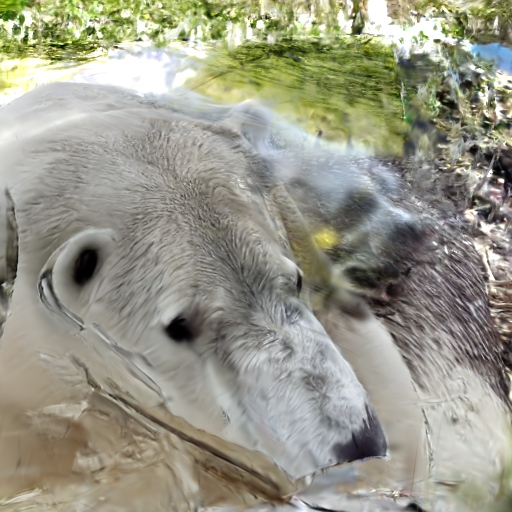}
    & \includegraphics[width=0.16\textwidth]{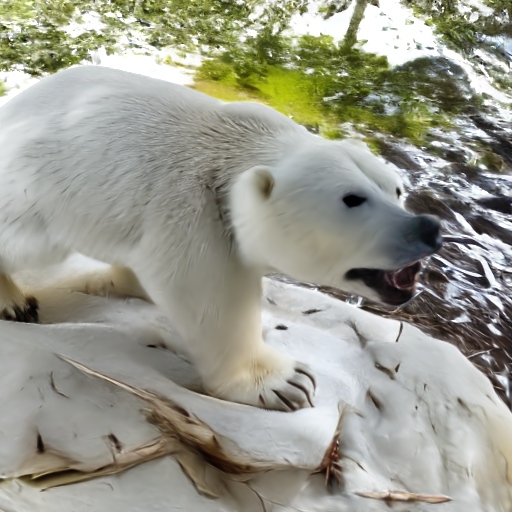}
    \\

    \raisebox{1.2\height}{\parbox[t]{12mm}{\rotatebox[origin=c]{90}{\makecell{ \\ a photo of a \\ human skeleton}}}}
    & \includegraphics[width=0.16\textwidth]{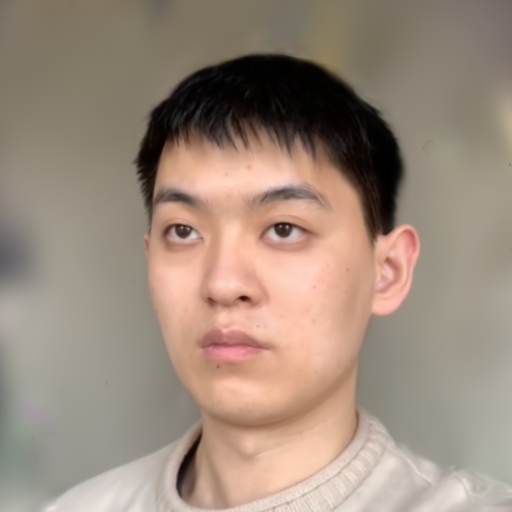}
    & \includegraphics[width=0.16\textwidth]{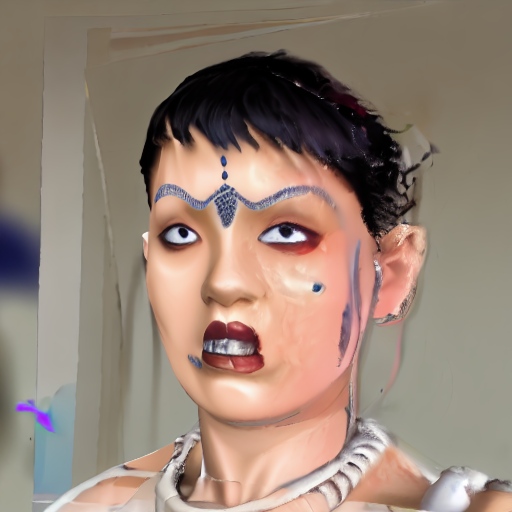}
    & \includegraphics[width=0.16\textwidth]{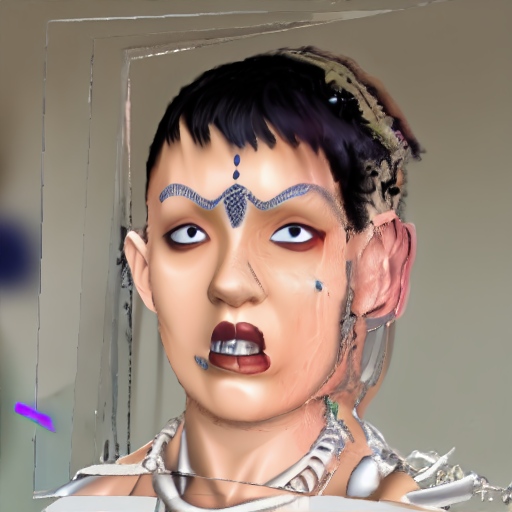}
    & \includegraphics[width=0.16\textwidth]{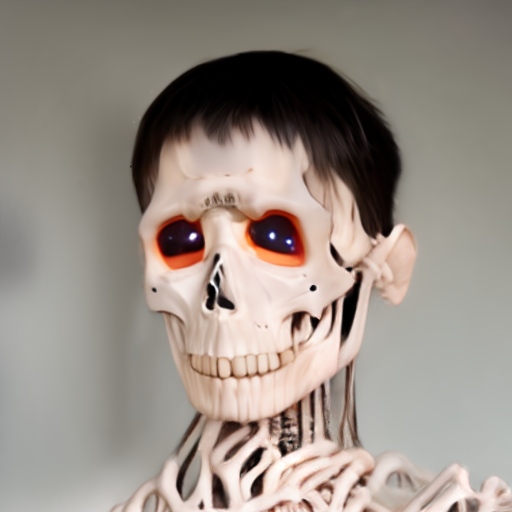}
    \\

    \raisebox{1.3\height}{\parbox[t]{12mm}{\rotatebox[origin=c]{90}{\makecell{Portrait of a \\ 19th-century \\ aristocrat}}}}
    & \includegraphics[width=0.16\textwidth]{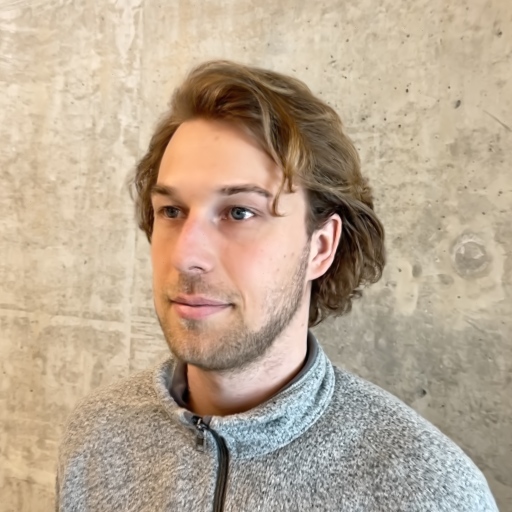}
    & \includegraphics[width=0.16\textwidth]{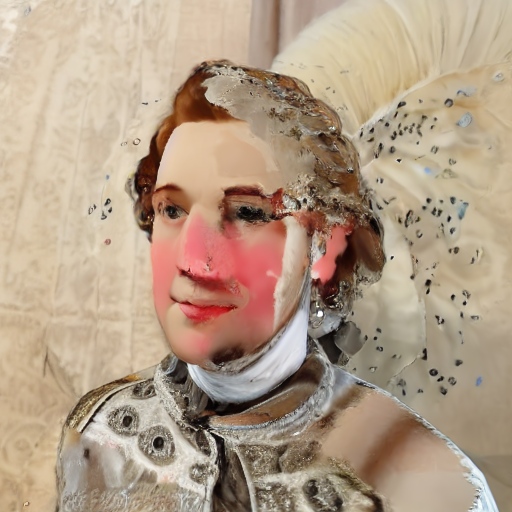}
    & \includegraphics[width=0.16\textwidth]{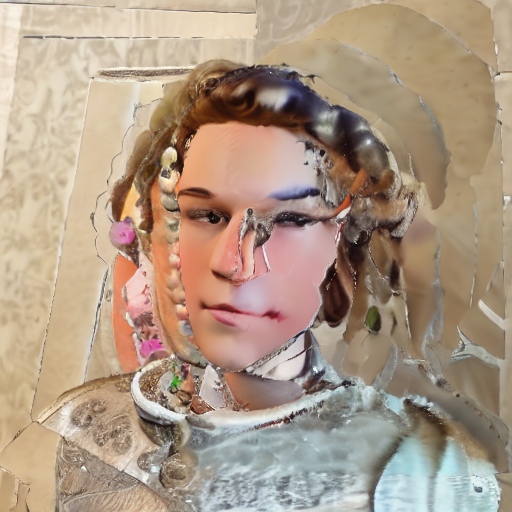}
    & \includegraphics[width=0.16\textwidth]{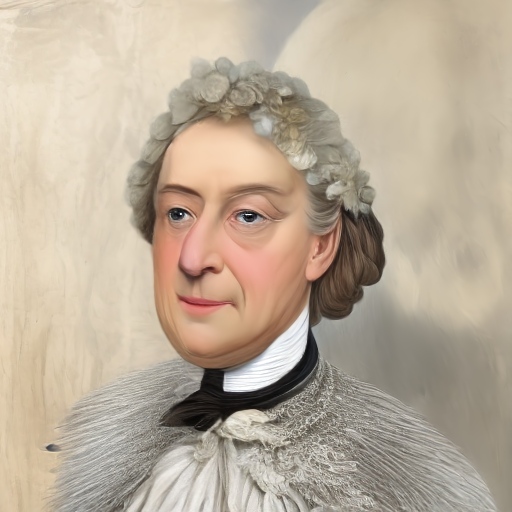}
    \\

    \raisebox{1.2\height}{\parbox[t]{12mm}{\rotatebox[origin=c]{90}{\makecell{a photo of a \\ fountain in the \\ desert}}}}
    & \includegraphics[width=0.16\textwidth]{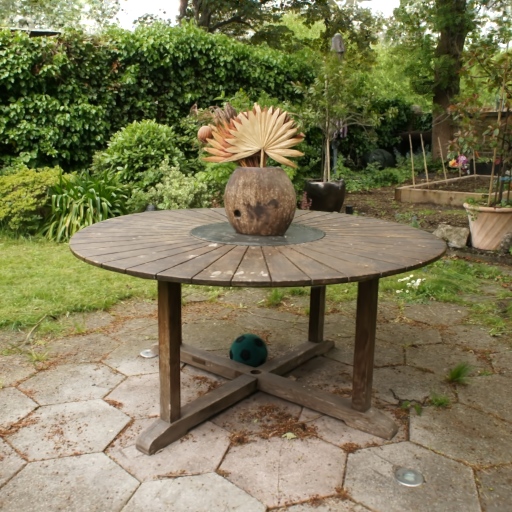}
    & \includegraphics[width=0.16\textwidth]{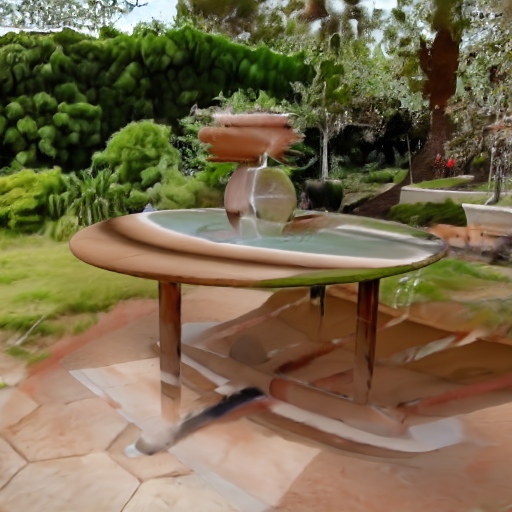}
    & \includegraphics[width=0.16\textwidth]{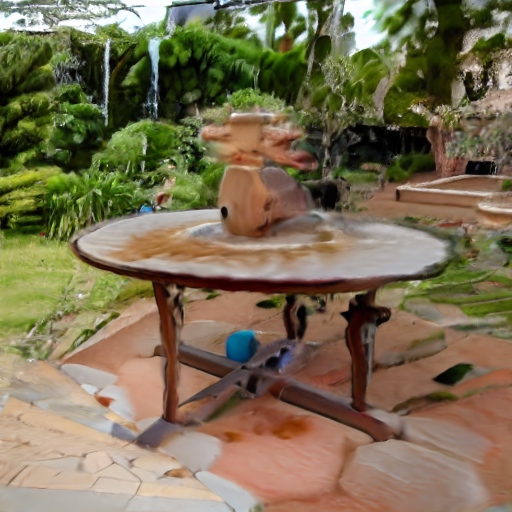}
    & \includegraphics[width=0.16\textwidth]{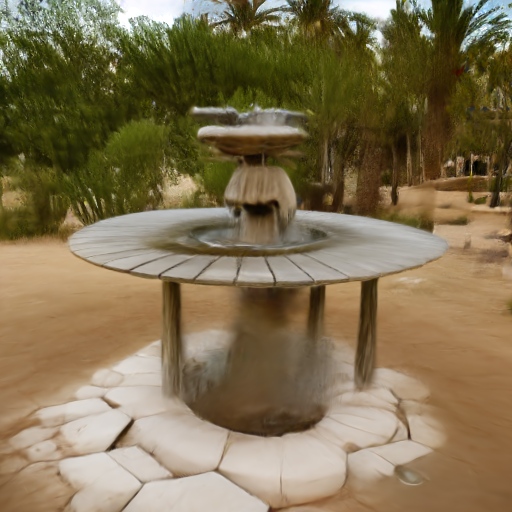}
    \\

    \raisebox{1.1\height}{\parbox[t]{12mm}{\rotatebox[origin=c]{90}{\makecell{futuristic night \\ city Cyberpunk \\ style robotics lab}}}}
    & \includegraphics[width=0.16\textwidth]{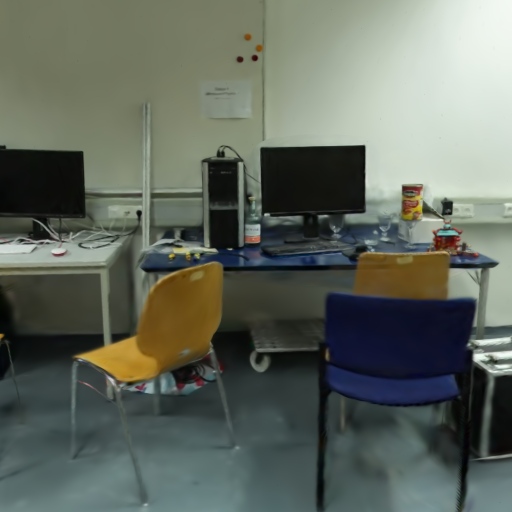}
    & \includegraphics[width=0.16\textwidth]{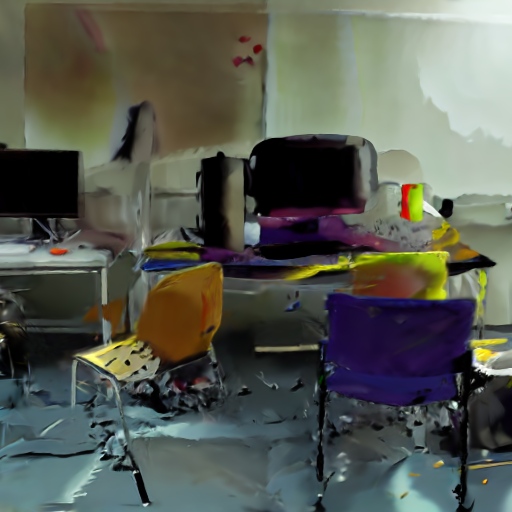}
    & \includegraphics[width=0.16\textwidth]{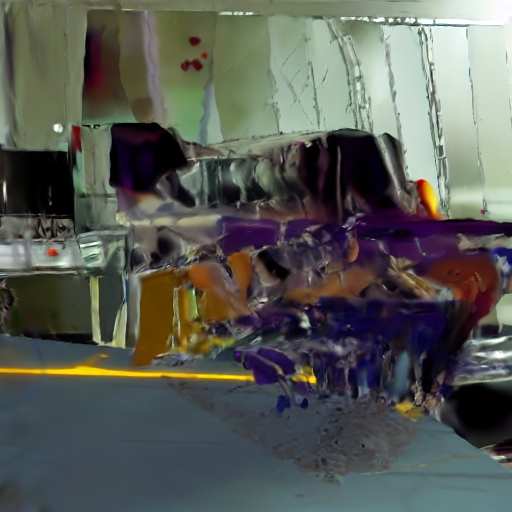}
    & \includegraphics[width=0.16\textwidth]{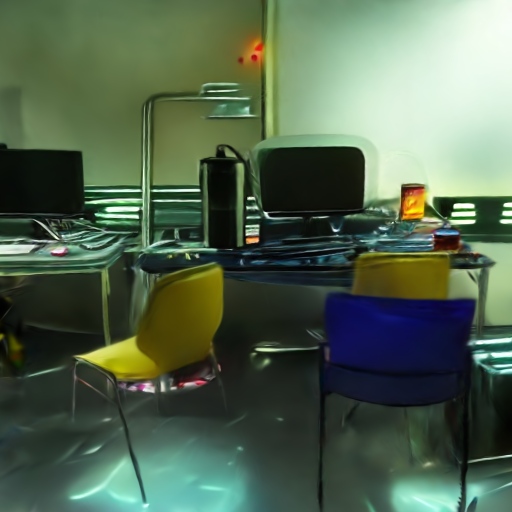}
    \\

    \raisebox{1.1\height}{\parbox[t]{12mm}{\rotatebox[origin=c]{90}{\makecell{a Fauvism \\ painting of a \\ conference room}}}}
    & \includegraphics[width=0.16\textwidth]{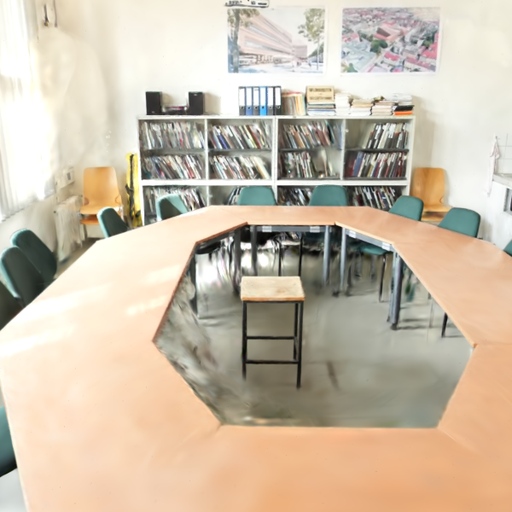}
    & \includegraphics[width=0.16\textwidth]{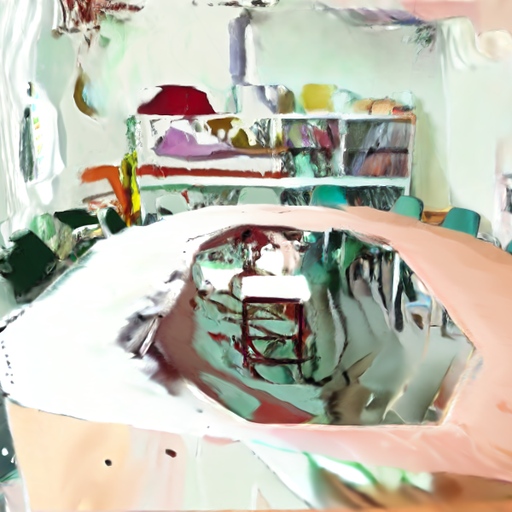}
    & \includegraphics[width=0.16\textwidth]{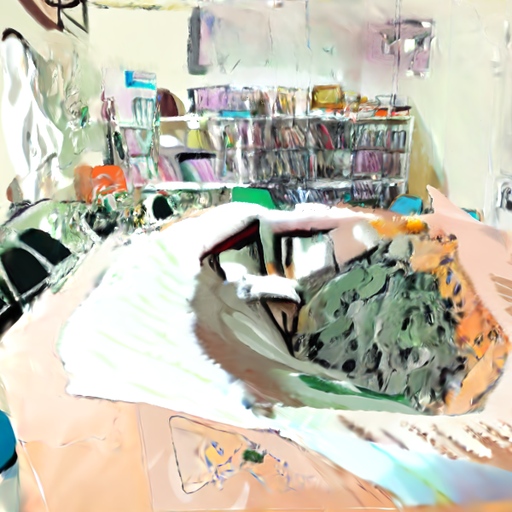}
    & \includegraphics[width=0.16\textwidth]{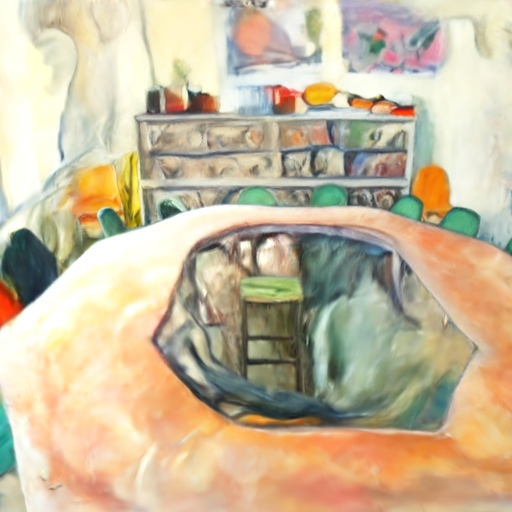}
    \\

    \raisebox{1.05\height}{\parbox[t]{12mm}{\rotatebox[origin=c]{90}{\makecell{Horse statue \\ made from stacked \\ wooden blocks}}}}
    & \includegraphics[width=0.16\textwidth]{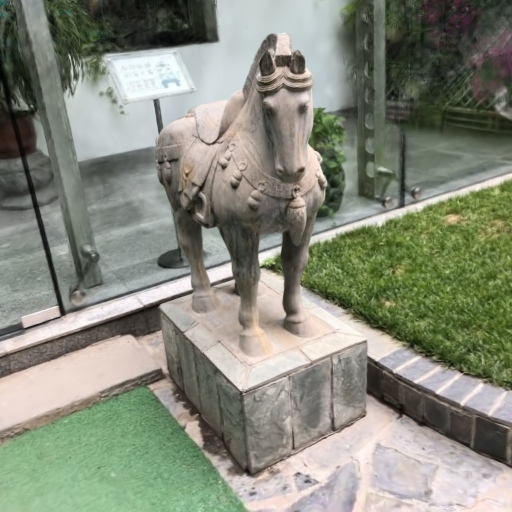}
    & \includegraphics[width=0.16\textwidth]{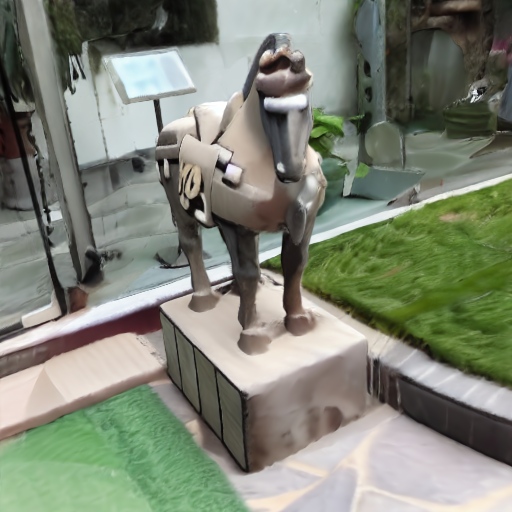}
    & \includegraphics[width=0.16\textwidth]{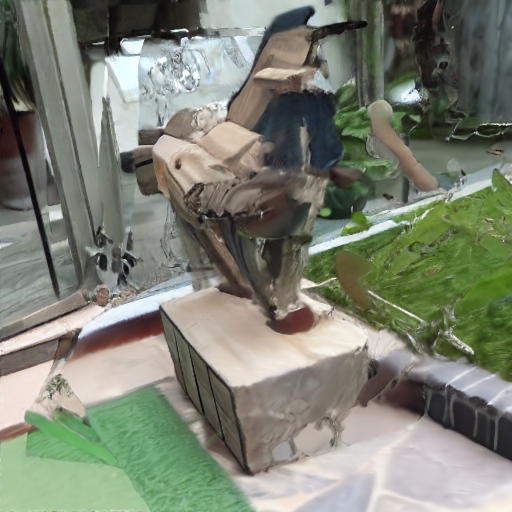}
    & \includegraphics[width=0.16\textwidth]{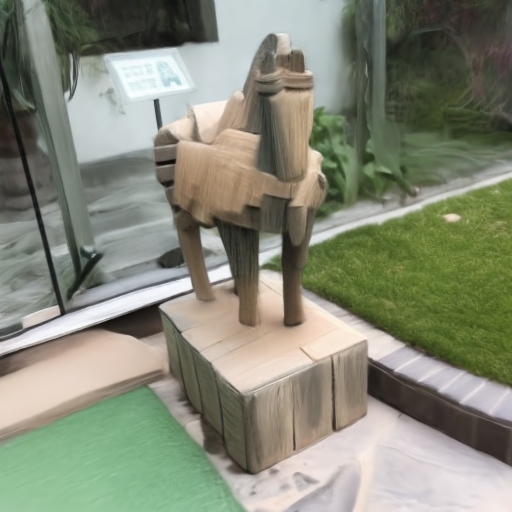}
    \\

\end{tabular}

    \vspace{-5pt}
    \caption{\textbf{Qualitative Comparison from Novel Views for Ablations}}
    \vspace{-5pt}
    \label{fig:ablations-qualitative-comparison}    
\end{figure*}

\section{Runtime}
We use an Nvidia A100 40GB to run our pipeline. On this hardware our method takes around 15 minutes to run on a typical scene, consisting of about 10 minutes for stylization and 5 minutes for training the stylized splat. These timings were measured with FP16 precision for inference, which we find produces no discernible decrease in quality.

% \begin{figure*}
%     \vspace{-5pt}
%     \centering
%     \small
    
%     \include{figures/abl_comparisons_figure/comparisons_transposed_2}
    
%     \vspace{-5pt}
%     \caption{\textbf{Qualitative Comparison from Novel Views for Ablations}}
%     \vspace{-5pt}
%     \label{fig:ablations-qualitative-comparison-2}    
% \end{figure*}

%%%%%%%%%%%%%%%%%%%%%%%%%%%%%%%%%%%%

% \clearpage
% {
%     \small
%     \bibliographystyle{ieeenat_fullname}
%     \bibliography{main}
% }


\end{document}